\documentclass[lettersize,journal]{IEEEtran}
\usepackage{amsmath,amsfonts}
\usepackage{algorithmic}
\usepackage{array}
\usepackage{textcomp}
\usepackage{stfloats}
\usepackage{url}
\usepackage{verbatim}
\usepackage{graphicx}
\def\BibTeX{{\rm B\kern-.05em{\sc i\kern-.025em b}\kern-.08em
    T\kern-.1667em\lower.7ex\hbox{E}\kern-.125emX}}
\usepackage{balance}

\usepackage{xspace}
\usepackage{graphicx}
\usepackage{amsmath}
\usepackage{amssymb}
\usepackage[font=small,labelfont=bf,tableposition=top]{caption}
\usepackage[font=scriptsize,skip=1pt,subrefformat=parens]{subcaption}
\usepackage{multirow}
\usepackage{enumitem}
\usepackage{arydshln} 
\usepackage[accsupp]{axessibility}
\usepackage[dvipsnames, svgnames, x11names, table,xcdraw]{xcolor}
\usepackage{colortbl,booktabs}
\usepackage{ragged2e}
\usepackage{makecell}
\usepackage{tabularx}
\usepackage{pifont}       
\usepackage{bbding}
\usepackage{fontawesome}
\usepackage{arydshln}

\newcommand{\bx}{\mathbf{x}}
\newcommand{\bX}{\mathbf{X}}

\newcommand{\bmu}{{\boldsymbol{\mu}}}
\newcommand{\bI}{\mathbf{I}}
\newcommand{\bSigma}{{\boldsymbol{\Sigma}}}
\newcommand{\bepsilon}{{\boldsymbol{\epsilon}}}

\newcommand{\re}[1]{\textcolor{Firebrick3}{#1}}

\newcommand{\bl}[1]{\textcolor{DeepSkyBlue3}{#1}}
\newcommand{\mg}[1]{\textcolor{DarkOliveGreen3}{#1}}

\newcommand{\gre}[1]{\textcolor{green}{#1}}
\definecolor{citecolor}{HTML}{0071BC}
\definecolor{linkcolor}{HTML}{ED1C24}
\definecolor{mydarkgreen}{rgb}{0.02,0.6,0.02}

\usepackage[pagebackref,colorlinks,citecolor=mydarkgreen,linkcolor=linkcolor]{hyperref}
\usepackage[capitalize,noabbrev]{cleveref}

\definecolor{orcidlogocol}{HTML}{A6CE39}

\usepackage{scalerel}
\usepackage{tikz}
\usetikzlibrary{svg.path}
\tikzset{
  orcidlogo/.pic={
    \fill[orcidlogocol] svg{M256,128c0,70.7-57.3,128-128,128C57.3,256,0,198.7,0,128C0,57.3,57.3,0,128,0C198.7,0,256,57.3,256,128z};
    \fill[white] svg{M86.3,186.2H70.9V79.1h15.4v48.4V186.2z}
                 svg{M108.9,79.1h41.6c39.6,0,57,28.3,57,53.6c0,27.5-21.5,53.6-56.8,53.6h-41.8V79.1z M124.3,172.4h24.5c34.9,0,42.9-26.5,42.9-39.7c0-21.5-13.7-39.7-43.7-39.7h-23.7V172.4z}
                 svg{M88.7,56.8c0,5.5-4.5,10.1-10.1,10.1c-5.6,0-10.1-4.6-10.1-10.1c0-5.6,4.5-10.1,10.1-10.1C84.2,46.7,88.7,51.3,88.7,56.8z};
  }
}
\newcommand\orcidicon[1]{\href{https://orcid.org/#1}{\mbox{\scalerel*{
\begin{tikzpicture}[yscale=-1,transform shape]
\pic{orcidlogo};
\end{tikzpicture}
}{|}}}}

\def\jing{\textcolor{black}}
\def\yushi{\textcolor{black}}

\begin{document}

\title{Temporal Feature Matters: A Framework for Diffusion Model Quantization}

\author{Yushi Huang$^{\orcidicon{0009-0002-7898-8402}}$,
        Ruihao~Gong$^{\orcidicon{0000-0002-6024-7086}}$,
        Xianglong Liu$^{\orcidicon{0000-0002-7618-3275}}$,~\IEEEmembership{Member,~IEEE,}
        Jing~Liu$^{\orcidicon{0000-0002-6745-3050}}$,
        Yuhang~Li$^{\orcidicon{0000-0002-6444-7253}}$,\\
        Jiwen~Lu$^{\orcidicon{ 0000-0002-6121-5529}}$,~\IEEEmembership{Fellow,~IEEE,} and Dacheng~Tao$^{\orcidicon{0000-0001-7225-5449}}$,~\IEEEmembership{Fellow,~IEEE}%

\thanks{This work was supported by the National Natural Science Foundation of China (No. 62476018), the Beijing Municipal Science and Technology Project (No. Z231100010323002), and the Postdoctoral Fellowship Program of CPSF (No. BX20250487) (Corresponding author: Xianglong Liu).}
\thanks{Yushi Huang is with the Department of Electrical and Computer Engineering, the Hong Kong University of Science and Technology, Hong Kong 999077 (e-mail: yhuangin@connect.ust.hk).}
\thanks{Ruihao Gong and Xianglong Liu are with the State Key Laboratory of Complex \& Critical Software Environment, Beihang University, Beijing 100191, China (e-mail: gongruihao@buaa.edu.cn; xlliu@buaa.edu.cn).}
\thanks{Jing Liu is with the Department of Information Technology, Monash University, Melbourne, VIC 3800, Australia (e-mail: liujing\_95@outlook.com).}
\thanks{Yuhang Li is with the Department of Electrical Engineering, Yale University, New Haven, CT 06511, USA (e-mail: yuhang.li@yale.edu).}
\thanks{Jiwen Lu is with the Department of Automation, Tsinghua University, Beijing 100084, China (e-mail: lujiwen@tsinghua.edu.cn).}
\thanks{Dacheng Tao is with the College of Computing and Data Science, Nanyang Technological University, Singapore 639798 (e-mail: dacheng.tao@ntu.edu.sg).}
\thanks{Our code is available at \url{https://github.com/ModelTC/TFMQ-DM}.}
}%

\markboth{Journal of \LaTeX\ Class Files,~Vol.~14, No.~8, August~2021}%
{Shell \MakeLowercase{\textit{et al.}}: Bare Advanced Demo of IEEEtran.cls for IEEE Computer Society Journals}

\maketitle

\begin{abstract}
\justifying
    Diffusion models, widely used for image generation, face significant challenges related to their broad applicability due to prolonged inference times and high memory demands. Efficient Post-Training Quantization (PTQ) is crucial to address these issues. However, unlike traditional models, diffusion models critically rely on the timestep for the multi-round denoising. Typically, each timestep is encoded into a hypersensitive temporal feature by several modules. Despite this, existing PTQ methods do not optimize these modules individually. Instead, they employ unsuitable reconstruction objectives and complex calibration methods, leading to significant disturbances in the temporal feature and denoising trajectory, as well as reduced compression efficiency. To address these challenges, we introduce a novel quantization framework that includes three strategies: 1)~\textbf{TIB-based Maintenance}: Based on our innovative Temporal Information Block~(TIB) definition, Temporal Information-aware Reconstruction~(TIAR) and Finite Set Calibration~(FSC) are developed to efficiently align original temporal features. 2)~\textbf{Cache-based Maintenance}: Instead of indirect and complex optimization for the related modules, pre-computing and caching quantized counterparts of temporal features are developed to minimize errors. 3)~\textbf{Disturbance-aware Selection}: Employ temporal feature errors to guide a fine-grained selection between the two maintenance strategies for further disturbance reduction. This framework preserves most of the temporal information and ensures high-quality end-to-end generation. Extensive testing on various datasets, diffusion models, and hardware confirms our superior performance and acceleration.
\end{abstract}


\section{Introduction}
\label{sec:intro}
\IEEEPARstart{G}{enerative} modeling is pivotal in machine learning, particularly in applications like image~\cite{kang2023scaling, songddim, ho2020denoising, rombach2022ldm, ho2022classifierfree, podell2023sdxlimprovinglatentdiffusion}, voice~\cite{shen2018natural, ren2022fastspeech}, and text synthesis~\cite{brown2020language, zhang2022opt}. Diffusion models have demonstrated remarkable proficiency in generating high-quality samples across diverse domains. In comparison to generative adversarial networks (GANs)~\cite{goodfellow2020generative} and variational autoencoders (VAEs)~\cite{kingma2022autoencoding}, diffusion models successfully sidestep issues such as model collapse and posterior collapse, offering a more stable training regimen. However, the substantial computational cost~\cite{ma2024deepcache, yang2023diffusion} poses a critical bottleneck hampering the widespread adoption of diffusion models. Specifically, this cost mainly stems from two primary factors. First, these models typically require hundreds of denoising steps to generate images, rendering the procedure considerably slower than that of GANs. Prior efforts~\cite{songddim, lu2022dpmsolver, kong2021fastdpm, liu2022pseudo} have tackled this challenge by seeking shorter and more efficient sampling trajectories, thereby reducing the number of necessary denoising steps. Second, the substantial parameters and complex architecture of diffusion models demand considerable time and memory resources for mobile device inference~\cite{chen2023speed, zhao2023mobilediffusion}, particularly for foundational models pre-trained on large-scale datasets, \emph{e.g.}, Stable Diffusion~\cite{rombach2022ldm}, SD-XL~\cite{podell2023sdxlimprovinglatentdiffusion}, and SD-XL-turbo~\cite{sauer2023adversarial}. Our work aims to address the latter challenge, focusing on the compression of diffusion models to enhance their efficiency and applicability.

Quantization, a technique for mapping high-precision floating-point numbers to lower-precision counterparts, stands as the most prevalent method for model compression~\cite{nagel2020adaround, gong2019dsq, nagel2021white, esser2020lsq, bhalgat2020lsq}. Among different quantization paradigms, post-training quantization (PTQ)~\cite{nagel2020adaround, wei2022qdrop, hubara2020improvingadaquant} incurs lower overhead and is more user-friendly without the need for retraining or fine-tuning with a huge amount of training data. While PTQ on conventional models has undergone extensive study~\cite{nagel2020adaround, li2021brecq, wei2022qdrop, gong2019dsq}, its adaptation to diffusion models has shown huge performance degradation, especially under low-bit settings. For instance, Q-Diffusion~\cite{li2023qdiffusion} exhibits $\mathbf{6.81}$  Fréchet Inception Distance~(FID)~\cite{heusel2018gans} increasing on CelebA-HQ $256\times256$~\cite{karras2019stylebased} under $4$-bit weight and $8$-bit activation quantization. We believe the reason they fail to achieve better results is that they all overlook the sampling data independence and uniqueness of hypersensitive temporal features, which are generated from timestep $t$ through a few modules, used to control the denoising trajectory in diffusion models. As a result, the temporal feature disturbance stemming from quantization significantly impacts model performance in the existing studies.

To tackle temporal feature disturbance, we first find that the modules generating temporal features are independent of the sampling data and thus treat the whole module as the Temporal Information Block~(TIB). 
However, existing methods~\cite{he2023ptqd,li2023qdiffusion,shang2022ptq4dm,wang2023towards,so2023temporal} do not separately optimize this block during the quantization process, causing temporal features to overfit to limited calibration data. Additionally, since the maximum timestep for denoising is a finite positive integer, the temporal feature and the activations during its generation form a finite set. Therefore, the optimal approach is to optimize each element in this set individually. Inspired by these observations and analyses, we propose two temporal feature maintenance techniques: 1)~\textbf{TIB-based Maintenance}: Based on TIB, a novel quantization reconstruction approach, Temporal Information-aware Reconstruction~(TIAR), is devised. It aims to reduce temporal feature error as the optimization objective while isolating the network's components related to sampled data during weight adjustment. Besides, we also introduce a calibration strategy, Finite Set Calibration~(FSC), for the finite set of the temporal feature and activation during its generation. 2)~\textbf{Cache-based Maintenance}: Further leveraging the independent and finite nature of the temporal feature, we directly pre-compute these features, and then optimize and cache their quantized versions to reduce the disturbance. In detail, this novel design only requires reloading cached features, avoiding online generation during inference, which enhances latency in specific cases. Moreover, we employ a 3)~\textbf{Disturbance-aware Selection}, taking advantage of both maintenance approaches to further eliminate temporal feature error. In addition, only one generation process before deployment can determine the selection results for every single temporal feature, respectively. This highly ensures the efficiency of our selection scenario. Finally, evaluating on multiple datasets, diverse tasks, advanced models, and different hardware, our novel framework for reducing temporal feature disturbance achieved nearly lossless model compression with high inference speed. 

In summary, our contributions are as follows:
\begin{itemize}[leftmargin=*]
    \item {We discover that existing quantization methods suffer from \jing{hypersensitive temporal feature disturbances and mismatched problem}, which disrupt the denoising trajectory of diffusion models and significantly affect the quality of generated images.}
    \item {We reveal that the disturbance comes from \jing{two sources}: an inappropriate reconstruction target and a lack of awareness of finite activations. Both inducements ignore the special characteristics of \jing{modules related to temporal information}.}
    \item {We propose an advanced quantization framework, consisting of 1)~TIB-based Maintenance: Temporal Information-aware Reconstruction~(TIAR) and Finite Set Calibration~(FSC). Both are based on a Temporal Information Block specially defined for diffusion models. 2)~Cache-based Maintenance: Directly optimize and save quantized temporal feature, and then reload for inference. 3)~Disturbance-aware Selection: Fine-grainedly select our maintenance strategies offline employing temporal feature error.}
    \item {Extensive experiments demonstrate our superior performance. For example, our framework reduces the FID score by $\mathbf{5.61}$ under the W$4$A$8$ configuration for SD-XL~\cite{podell2023sdxlimprovinglatentdiffusion}. Additionally, we deploy the quantized model on different hardware to demonstrate the inference acceleration, \emph{e.g.}, $\mathbf{2.20\times}$ and $\mathbf{5.76\times}$ speedup on Intel$^\circledR$ Xeon$^\circledR$ Gold $6448Y$ Processor and NVIDIA $H800$ GPU for SD-XL, respectively.}
\end{itemize}

This paper extends our conference version~\cite{Huang_2024_CVPR} in several aspects. 1)~We compare and fine-grainedly analyze the sensitivity and quantization-induced disturbance between temporal and other non-temporal features. 2)~We propose Cache-based Maintenance parallel to TIB-based Maintenance with high efficiency. It can achieve comparable performance and even lower inference costs in some cases. 3)~We propose Disturbance-aware Selection, which fine-grainedly harnesses both maintenance strategies to further reduce disturbance in temporal features. 4)~We apply our framework to advanced Text-to-Image~(T2I) models, \emph{e.g.}, SD-XL~\cite{podell2023sdxlimprovinglatentdiffusion}, SD-XL-turbo~\cite{sauer2023adversarial}, and FLUX.1-Schnell~\cite{flux2024}. Moreover, we also apply our methods to the video generation model--OpenSora~\cite{opensora}. 5)~We deploy our quantized model employing CUTLASS~\cite{kerr2017cutlass} with our customized kernel on NVIDIA $H800$ GPU and OpenVino~\cite{openvino} on Intel$^\circledR$ Xeon$^\circledR$ Gold $6448Y$ Processor to show significant acceleration. Additionally, we also test the efficiency of the quantized model on NVIDIA Jetson Nano Orin and iPhone $15$ Pro Max. 6)~We provide comprehensive experiments and more ablative studies to investigate the effectiveness of our methods.

\section{Related Work}
\label{sec:relatedwork}

\subsection{Efficient Diffusion Models}
Diffusion models have demonstrated the ability to generate high-quality images. However, their extensive iterative process, coupled with the computational cost of denoising via networks, has impeded wide-ranging applications. To tackle this challenge, prior research has explored efficient methods to expedite the denoising process~\cite{watson2022learning, huang2025harmonicaharmonizingtraininginference,wnag2024ptsbenchcomprehensiveposttrainingsparsity}. These methods fall into two categories: those requiring re-training and advanced samplers tailored for pre-trained models. The former category includes techniques like diffusion process learning~\cite{chung2022come, lyu2022accelerating, zheng2022truncated, franzese2022much}, noise scale adaptation~\cite{nichol2021improvedDDPM, kingma2021variational}, knowledge distillation~\cite{salimans2022progressivedistillation, luhman2021kdingdm, kim2023bk}, and sample trajectory refinement~\cite{watson2022learning, lam2021bilateral, nguyen2024swiftbrush}. While these methods can speed up sampling, re-training a diffusion model proves resource-intensive and time-consuming. In contrast, the latter category involves designing efficient samplers for pre-trained diffusion models, eliminating the need for re-training. Key methods in this category encompass analytical trajectory estimation~\cite{bao2022analytic, bao2022estimating}, implicit samplers~\cite{kong2021fastdpm, songddim, zhang2022gddim}, and differential equation (DE) solvers such as customized stochastic differential equations (SDE)~\cite{song2020score, jolicoeur2021gotta, kim2022denoisingMCMC} and ordinary differential equations (ODE)~\cite{lu2022dpmsolver, liu2022pseudo, zhang2022fast}. While these approaches can reduce sampling iterations, diffusion models' extensive parameters, and computational complexity restrict their deployment in real-world settings. In this paper, our work focuses on diminishing the time and memory overhead of the single-step denoising process using low-bit quantization in a post-training manner, a method orthogonal to previous speedup techniques.

Besides the above-mentioned studies, there are also some works, \emph{e.g.}, SnapFusion~\cite{li2023snapfusiontexttoimagediffusionmodel} and BitsFusion~\cite{sui2024bitsfusion199bitsweight} aim to extremely compress the diffusion model with a mix of multiple techniques for deployment on edge devices. However, these methods also require extensive training resources.

\subsection{Model Quantization}
Quantization is a predominant technique for minimizing storage and computational costs. It can be categorized into quantization-aware training (QAT)~\cite{gong2019dsq, zhang2023root, zhuang2018towards, huang2025qvgenpushinglimitquantized} and post-training quantization (PTQ)~\cite{li2021brecq, nagel2020adaround, wei2022qdrop, gong2024llmcbenchmarkinglargelanguage}. QAT requires intensive model training with substantial data and computational demands. Correspondingly, PTQ compresses models without re-training, making it a preferred method due to its minimal data requirements and easy deployment on real hardware. In PTQ, given a high-precision value $x$, we  map it into discrete one $\hat{x}$ using uniform quantization expressed as:
\begin{equation}
    \hat{x} = \Phi ( \lfloor \frac{x}{s} \rceil + z, 0, 2^{b}-1),
    \label{eq:uniform_quant}
\end{equation}
where $s$ is the quantization step size, $z$ is the zero offset, and $b$ is the target bit-width. The clamp function $\Phi(\cdot)$ clips the rounded value $\left\lfloor \frac{x}{s} \right\rceil + z$ within the range of $[0, 2^{b}-1]$. However, naive quantization may lead to accuracy degradation, especially for low-bit quantization. Recent studies~\cite{li2021brecq, frantar2022optimal, liu2023pd, li2022efficient} have explored innovative strategies based on reconstruction to preserve model performance after low-bit quantization. For instance,  AdaRound~\cite{nagel2020adaround} extends PTQ to $4$-bit on traditional vision models using a novel rounding mechanism with layer-wise reconstruction~\cite{hubara2020improvingadaquant, wang2020towards}. BRECQ~\cite{li2021brecq} balances cross-layer dependency and generalization error by leveraging neural network blocks and employing block-wise reconstruction. Additionally, QDrop~\cite{wei2022qdrop} considers the impact of activation quantization during reconstruction.

In contrast, the iterative denoising process in diffusion models presents new challenges for PTQ in comparison to traditional models. PTQ4DM~\cite{shang2022ptq4dm} represents the initial attempt to quantize diffusion models to $8$-bit, albeit with limited experiments and lower resolutions. Conversely, Q-Diffusion~\cite{li2023qdiffusion} achieves enhanced performance and is evaluated on a broader dataset range. Moreover, PTQD~\cite{he2023ptqd} eliminates quantization noise through correlated and residual noise correction. Notably, traditional single-timestep PTQ calibration methods are unsuitable for diffusion models due to significant activation distribution changes with each timestep~\cite{shang2022ptq4dm, li2023qdiffusion, wang2023towards, so2023temporal}. ADP-DM~\cite{wang2023towards} proposes group-wise quantization across timesteps for diffusion models, and TDQ~\cite{so2023temporal} introduces distinct quantization parameters for different timesteps. However, all of the above works overlook the specificity of hypersensitive temporal features. To address temporal feature disturbance in the aforementioned works, our study delves into the inducements of the phenomenon and introduces a novel PTQ framework for diffusion models, significantly enhancing quantized diffusion model performance.

\section{Preliminaries}\label{pre}
\noindent\textbf{Diffusion models.} Diffusion models~\cite{songddim, ho2020denoising} iteratively add Gaussian noise with a variance schedule $\beta_1, \ldots, \beta_T \in (0, 1)$ to data $\bx_0 \sim q(\bx)$ for $T$ times as sampling process, resulting in a sequence of noisy samples $\bx_1, \ldots, \bx_T$. In DDPMs~\cite{ho2020denoising}, the sampling process is a Markov chain, taking the form:
\begin{align}
        q(\bx_{1:T} | \bx_0) & = \prod_{t=1}^T q(\bx_t | \bx_{t-1} ), \\
        q(\bx_t|\bx_{t-1}) & = \mathcal{N}(\bx_t;\sqrt{\alpha_t}\bx_{t-1}, \beta_t \bI),
        \label{eq:sampling}
\end{align}
where $\alpha_t=1-\beta_t$. Conversely, the denoising process removes noise from a sample of Gaussian noise $\bx_T \sim \mathcal{N}(\mathbf{0}, \mathbf{I})$ to gradually generate high-fidelity images. Nevertheless, due to the unavailability of the true reverse conditional distribution $q(\bx_{t-1} | \bx_t)$, diffusion models approximate it via variational inference by learning a Gaussian distribution $p_\theta(\bx_{t-1} | \bx_t)=\mathcal{N}(\bx_{t-1}; \bmu_\theta(\bx_t, t), \bSigma_\theta(\bx_t, t))$, the $\bmu_\theta$ can be derived by reparameterization trick as follows:
\begin{equation}
    \bmu_\theta(\bx_t, t)  = \frac{1}{\sqrt{\alpha_t}}\left( \bx_t - \frac{\beta_t}{\sqrt{1-\bar\alpha_t}} \bepsilon_\theta(\bx_t, t) \right), 
    \label{eq:ddpm_reverse_mean}
\end{equation}
where $\bar\alpha_t = \prod_{i=1}^t \alpha_i$ and $\bepsilon_\theta(\cdot)$ is a noise estimation model. The variance $\bSigma_\theta(\bx_t, t)$ can be either learned~\cite{nichol2021improvedDDPM} or fixed to a constant schedule~\cite{ho2020denoising} $\sigma_t$. When employing the latter method, $\bx_{t-1}$ can be expressed as:
\begin{equation}
\bx_{t-1} = \frac{1}{\sqrt{\alpha_t}}\left( \bx_t - \frac{\beta_t}{\sqrt{1-\bar\alpha_t}} {\bepsilon}_\theta(\bx_t, t) \right) + \sigma_t \mathbf{z},
\label{eq:denoising}
\end{equation}
where $\mathbf{z} \sim \mathcal{N}(\mathbf{0}, \mathbf{I})$.

\begin{figure}[!ht]
   \centering
    \setlength{\abovecaptionskip}{0.2cm}
     \includegraphics[width=0.48\textwidth]{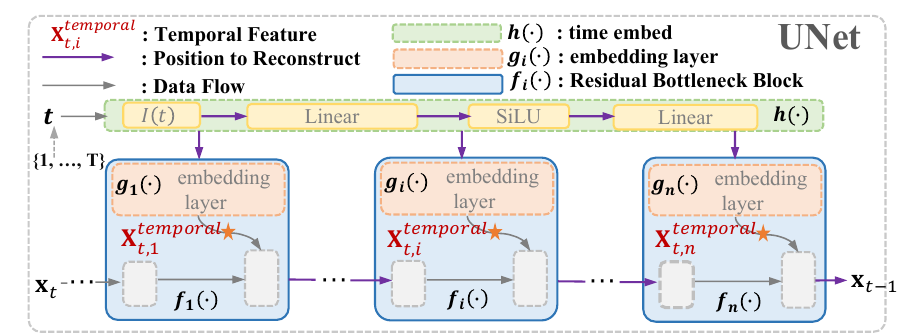}
     \caption{UNet architecture with a single denoising process at $t$. We omit the Transformer Blocks, some convolutions, and the sampler in the figure.}
    \label{fig:unet}
\end{figure}

\vspace{1em}
\noindent\textbf{Reconstruction on diffusion models.}\label{recon} Depicted in Fig.~\ref{fig:unet}, UNet architecture~\cite{ronneberger2015unet}, the predominant model employed as $\bepsilon_\theta(\cdot)$ in Eq.~(\ref{eq:denoising}) to predict Gaussian noise, can be divided into blocks that incorporate residual connections (such as Residual Bottleneck Blocks or Transformer Blocks~\cite{peebles2023scalable}) and the remaining layers. Numerous PTQ techniques for diffusion models are grounded in layer/block-wise reconstruction~\cite{shang2022ptq4dm, he2023ptqd, li2023qdiffusion, so2023temporal} to \jing{learn} optimal quantization parameters. For example, in the Residual Bottleneck Block, this approach typically minimizes the loss function as follows:
\begin{equation}
    \mathcal{L}_i=\| f_i(\cdot) - \widehat{f_i}(\cdot)\|_{F}^2,
    \label{lossfunction}
\end{equation}
where $\|\cdot\|^2_{F}$ denotes the Frobenius norm. The function $f_i(\cdot)$ represents the \textit{$i$-th} Residual Bottleneck Block, and $\widehat{f_i}(\cdot)$ denotes its quantized counterpart. Furthermore, in the following sections, we use $n$ to denote the total number of Residual Bottleneck Blocks in a single diffusion model. Unless otherwise stated, we adopt the aforementioned method as our baseline in Sec.~\ref{sec:tfmq}.

\vspace{1em}
\noindent\textbf{Temporal feature in diffusion models. } Also shown in Fig.~\ref{fig:unet}, timestep $t$ is encoded with \verb|time embed|\footnote{\href{https://github.com/CompVis/stable-diffusion/blob/main/ldm/modules/diffusionmodules/openaimodel.py\#L507}{PyTorch \texttt{time embed} implementation in diffusion models.}} and then passes through the \verb|embedding layer|\footnote{\href{https://github.com/CompVis/stable-diffusion/blob/main/ldm/modules/diffusionmodules/openaimodel.py\#L218}{PyTorch \texttt{embedding layer} implementation in diffusion models.}} in each Residual Bottleneck Block, resulting in a series of unique activations. In this paper, we denote these activations as temporal features. Notably, temporal features are independent of $\bx_t$ and unrelated to other temporal features from different timesteps. For enhanced clarity in our notation: the \verb|time embed| function is denoted as $h(\cdot)$, the \verb|embedding layer| within the \textit{$i$-th} Residual Bottleneck Block as $g_i(\cdot)$, and the resultant \textit{$i$-th} temporal feature at timestep $t$ as $\bX_{t,i}^{temporal}$. As illustrated in Fig.~\ref{fig:unet}, the relationship among these components is expressed as
\begin{equation}
\bX_{t,i}^{temporal} = g_i(h(t)).
\label{emb_rel}
\end{equation}

Additionally, we have found that temporal features play a crucial role in the context of the diffusion model as they hold unique and substantial physical implications. These features contain temporal information that indicates the current image’s position along the denoising trajectory. Within the architecture of the UNet, each timestep is converted into these temporal features, which then guide the denoising process by applying them to the image features generated in successive iterations.

\section{Temporal Feature Matters}\label{sec:tfmq}
We organize this section as follows. Firstly, we observe the disturbance induced by the previous method of the highly sensitive temporal feature in Sec.~\ref{sec:sensitivity_disturbance}. Further findings in Sec.~\ref{sec:impact} confirm the drastic impact. Concurrently, we analyze the inducements in Sec.~\ref{sec:inducement}. Finally, we propose our quantization framework in Sec.~\ref{sec:quantization_framework}.

\subsection{Disturbance Observations}\label{sec:sensitivity_disturbance}
Based on Sec.~\ref{pre}, we investigate how temporal features exhibit salient sensitivity to disturbance, and then we identify the phenomenon of temporal feature disturbance induced by quantization.

\begin{figure}[!ht]
   \centering
    \setlength{\abovecaptionskip}{0.2cm}
     \includegraphics[width=0.48\textwidth]{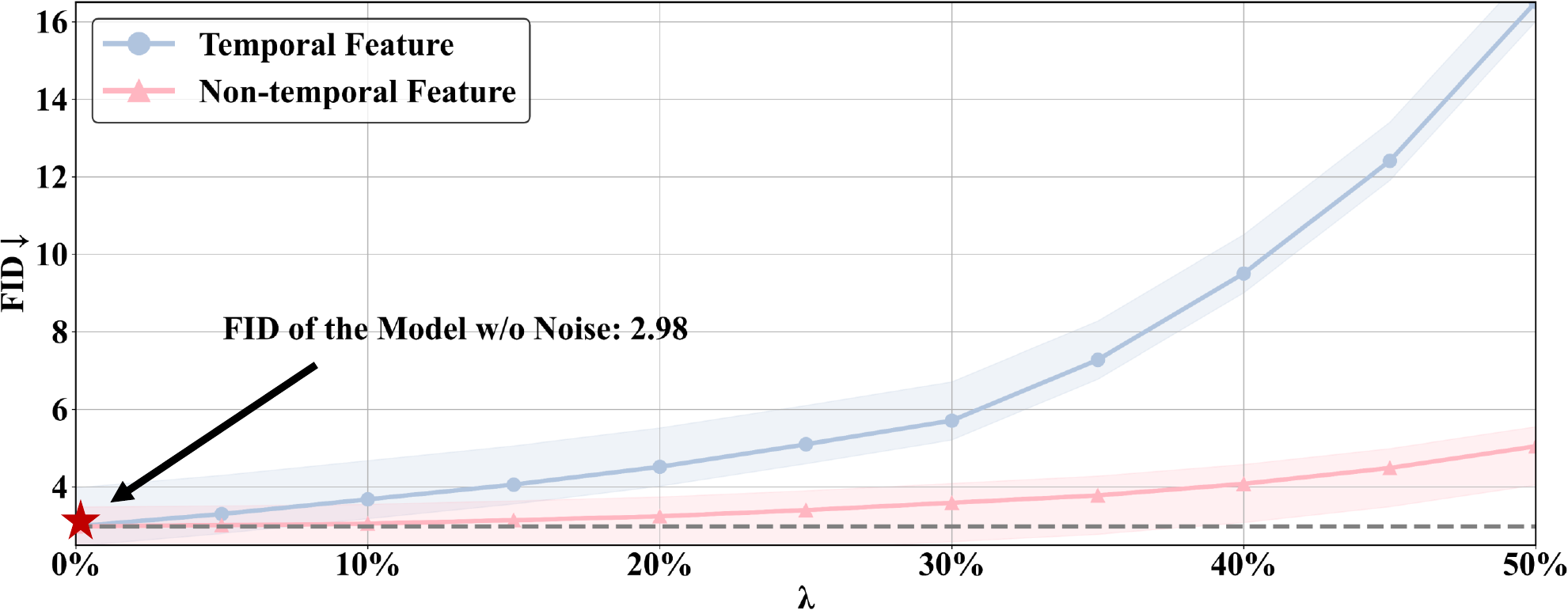}
     \caption{FID~\cite{heusel2018gans} on LSUN-Bedrooms $256\times256$~\cite{yu2016lsun} for LDM-4, with Gaussian noise applied to its temporal/non-temporal features.}
    \label{fig:sensitivity}
\end{figure}

\vspace{1em}
\noindent\textbf{Temporal feature sensitivity.} Beyond the special physical significance of temporal feature on image generation, we find these features are more sensitive than others~(\emph{i.e.}, other activations in the model can endure greater disturbances with minimal impact on performance). To validate this, we first introduce non-temporal features $\bX_{t,i}^{non-temporal}$ to represent activations except temporal features for every timestep $t$. Then, we randomly select $n$ non-temporal features~\footnote{Hence, here $i=1, \ldots, n$, the same for the temporal feature.} at $t$, and apply varying levels of random Gaussian noise to both (\emph{i.e.}, temporal and non-temporal features). The noise can be formulated as follows:
\begin{equation}
    \mathbf{\Delta}_\lambda= \lambda\mathbf{\Delta},
    \label{eq:noise}
\end{equation}
where $\lambda$ represents the noise level for the model. $\mathbf{\Delta}\sim\mathcal{N}(\mu, \sigma^2)$, where $\mu$ and $\sigma^2$ denote mean and variance for the corresponding temporal/non-temporal feature, respectively. As shown in Fig.~\ref{fig:sensitivity}, the FID deteriorates sharply with increasing $\lambda$ when disturbances are applied to the temporal features, in contrast to a gradual increase observed for random non-temporal features.

\vspace{1em}
\noindent\textbf{Temporal feature disturbance.} Having established the sensitivity of the temporal feature to disturbance, we thoroughly analyze its variations before and after the quantization of \verb|embedding layers| and \verb|time embed| in the Stable Diffusion model. Prior to this analysis, we introduce the temporal/non-temporal feature error at $t$ as defined by:
\begin{equation}
    E_t^{*}=\frac{\sum_{i=1}^n\text{cos}(\bX_{t,i}^{*}, \widehat{\bX_{t,i}^{*}})}{n},
    \label{eq:temporal_feature_error}
\end{equation}
where $*$ can be ``$temporal$" or ``$non-temporal$", $\text{cos}(\cdot)$ denotes cosine similarity, and $\widehat{\bX_{t,i}^*}$ signifies the temporal/non-temporal feature corresponding to $\bX_{t,i}^*$ in the quantized model. Notably, sample-independent temporal features do not have cumulative error from the iterative quantized denoising process. Conversely, to eliminate the cumulative error amplifying the quantization error, we employ the full-precision denoising from $T$ to $t+1$ to get the input $\bx_t$, and then fetch $\{\widehat{\bX_{t,i}^{non-temporal}}\}_{i=1,\ldots n}$ from the quantized denoising at $t$. 
\begin{figure}[!ht]
   \centering
    \setlength{\abovecaptionskip}{0.2cm}
     \includegraphics[width=0.48\textwidth]{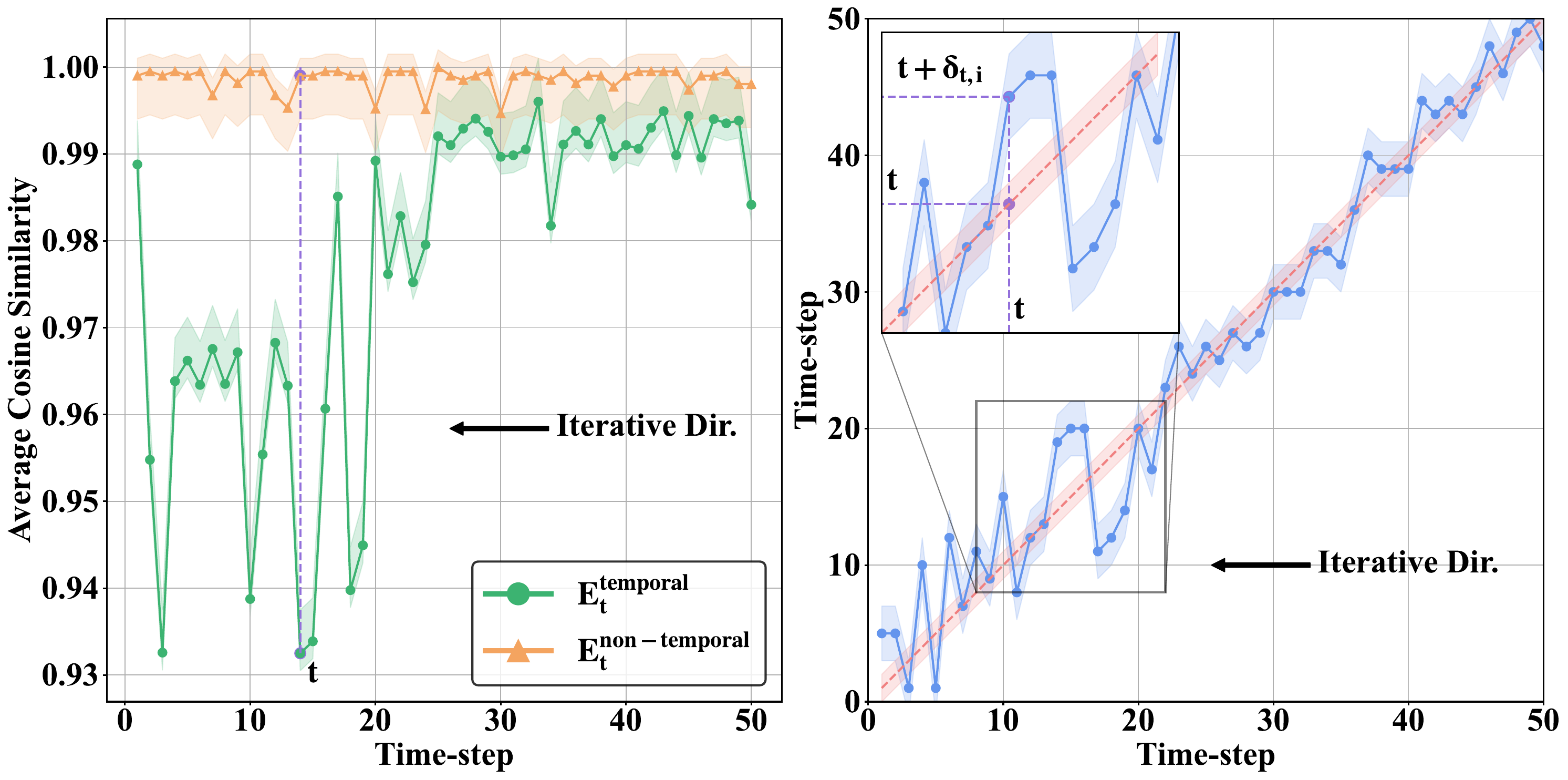}
     \caption{(Left)~Temporal feature disturbance. The green inflection points highlight the significant phenomenon of temporal feature disturbance. (Right)~Temporal information mismatch~($i=11$). The coordinates of the inflection points on the blue curve can be denoted as $(t, t+\delta_{t, i})$. It indicates $\widehat{\bX_{t,i}^{temporal}}$ exhibits the highest similarity with $\bX_{t+\delta_{t, i}, i}^{temporal}$. To be noted, for any point on the polylines in the above figures, assuming its x-axis is $t_0\in [1,\ldots, T]$, we pursue $T-t_0+1$ denoising processes to obtain its y-axis from the same random noise. Specifically, the model is set to full precision from $T$ to $t_0+1$ and only quantized precision at $t_0$.}
    \label{fig:disturbance_mismatch}
\end{figure}
As illustrated in Fig.~\ref{fig:disturbance_mismatch}~(Left). Quantization induces notable errors in temporal features, far exceeding those in non-temporal ones. We refer to this phenomenon, characterized by substantial temporal feature errors within diffusion models, as temporal feature disturbance.

\subsection{Disturbance Impacts}\label{sec:impact}
In addition to the intense disturbances caused by previous quantization approaches to highly sensitive temporal features, we further explore their subsequent impacts.

\vspace{1em}
\noindent\textbf{Temporal information mismatch.} Obviously, temporal feature disturbance alters the original embedded temporal information. Specifically, $\bX_{t, i}^{temporal}$ is intended to correspond to timestep $t$. However, due to significant errors, the quantized model's $\widehat{\bX_{t, i}^{temporal}}$ is no longer accurately associated with $t$, resulting in what we term as temporal information mismatch: 
\begin{equation}
    t \leftarrow \bX_{t,i}^{temporal},\quad t \nleftarrow \widehat{\bX_{t,i}^{temporal}}.
    \label{misma}
\end{equation}
Furthermore, as depicted in Fig.~\ref{fig:disturbance_mismatch}~(Right), we even observe a pronounced temporal information mismatch in Stable Diffusion. Specifically, the temporal feature generated by the quantized model at timestep $t$ exhibits a divergence from that of the full-precision model at the corresponding timestep. Instead, it tends to align more closely with the temporal feature corresponding to $t + \delta_t$, importing wrong temporal information from $t+\delta_t$.

\begin{figure}[!ht]
\centering
\setlength{\abovecaptionskip}{0.2cm}
\renewcommand{\arraystretch}{0.5}
\begin{tabular*}{\linewidth}{@{\extracolsep{\fill}}c}
\toprule
\footnotesize{\shortstack{\textit{``A man in the forest riding a horse.''}}} \\
\midrule
\includegraphics[width=0.488\textwidth]{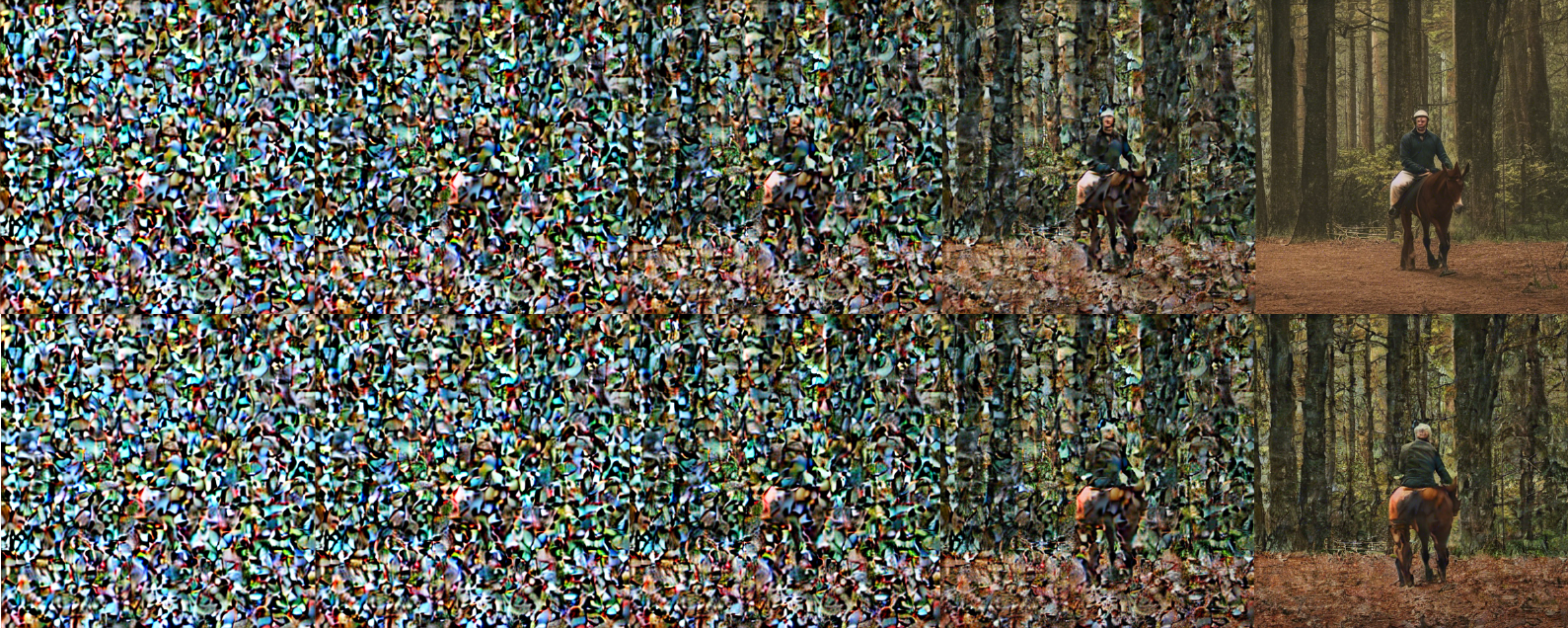} \\
\bottomrule
\end{tabular*}
\caption{\label{denoising_compare} Denoising process of full-precision~(Upper) and W$4$A$8$ quantized~(Lower) Stable Diffusion $(T = 50)$ under the same experiment settings, $\bx_0$. It is noteworthy that, in the quantized model employed here, to showcase the impact of temporal features, only the layers generating temporal features are quantized, and the components unrelated to the generation of temporal features are maintained in full precision.
}
\end{figure}

\vspace{1em}
\noindent\textbf{Trajectory deviation.} Temporal information mismatch delivers wrong temporal information, therefore, causing a deviation in the corresponding temporal position of the image within the denoising trajectory, ultimately leading to:
\begin{equation}
    \bx_t \nRightarrow \bx_{t-1},
    \label{devi}
\end{equation}
where we apply disrupted temporal features to the model. Evidently, the deviation in the denoising trajectory intensifies as the number of denoising iterations increases, resulting in the final generated image failing to align accurately with $\bx_0$. This evolution is illustrated in Fig.~\ref{denoising_compare}, where we maintain UNet in full precision, \jing{except for} \verb|embedding layers| and \verb|time embed|. We sort out a detailed relationship between temporal feature errors and final generation quality in Sec.~\ref{app:relation}.

\begin{figure*}[!ht]
    \centering
    \setlength{\abovecaptionskip}{0.2cm}
    \includegraphics[width=0.85\textwidth]{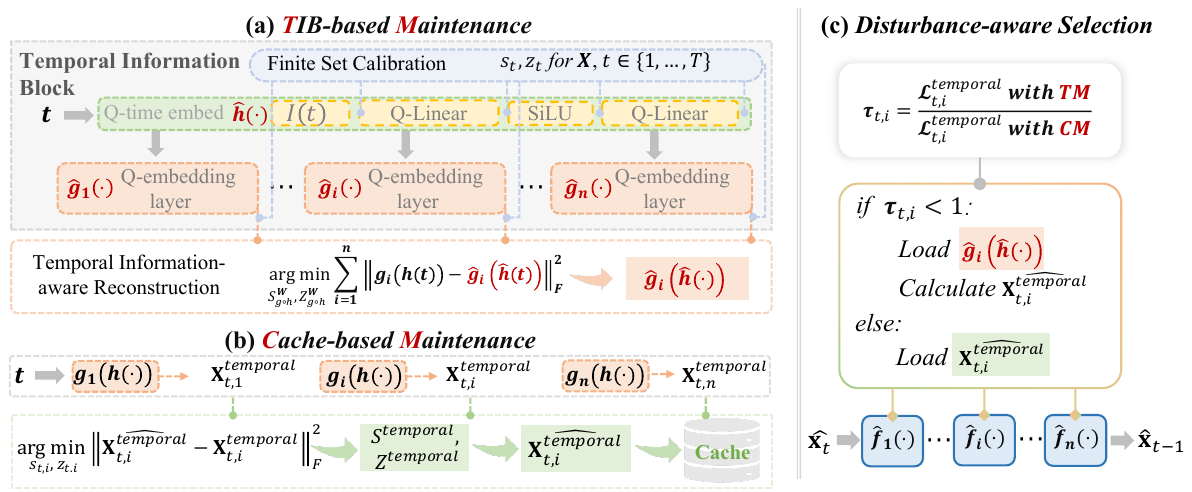}
     \caption{Overview of the proposed framework. (a)~\textbf{TIB-based Maintenance}: Based on a Temporal Information Block, we enable Temporal Information-aware Reconstruction and Finite Set Calibration for weights and activations quantization, respectively. (b)~\textbf{Cache-based Maintenance}:  We directly quantize pre-fetched temporal features and cache their quantized version. During inference, we can reuse them and drop $\{g_i\}_{i=1,\ldots,n}\cup\{h\}$. (c)~\textbf{Disturbance-aware Selection}: We select the appropriate maintenance for every single temporal feature utilizing the ratio of temporal feature error (\emph{i.e.}, Eq.~(\ref{eq:select})) between the two maintenance methods. This framework achieves the maintenance of temporal features and yields promising results.}
    \label{fig:framework}
\end{figure*}

\subsection{Inducement Analyses}\label{sec:inducement}
In this section, we explore the two inducements of temporal feature disturbance. For the purpose of clarity, in the subsequent sections, ``reconstruction" specifically points to slight weight adjustment for minimal quantization error, while ``calibration" specifically refers to activation calibration.

\vspace{1em}
\noindent\textbf{Inappropriate reconstruction target.} Previous PTQ methods~\cite{he2023ptqd, li2023qdiffusion, so2023temporal} have achieved remarkable progress on diffusion models. However, these existing methods overlook the temporal feature's independence and its distinctive physical significance. In their reconstruction processes, there was a lack of optimization for the \verb|embedding layers|. Instead, a Residual Bottleneck Block of coarser granularity was selected as the reconstruction target as depicted in Fig.~\ref{fig:unet}. This method involves two potential factors causing temporal feature disturbance: 1)~Optimize the objective as expressed in Eq.~(\ref{lossfunction}) to decrease the reconstruction loss of the Residual Bottleneck Block, as opposed to directly alleviating temporal feature disturbance. 2)~During backpropagation of the reconstruction process, \verb|embedding layers| independent from $\bx_t$ are affected by $\bx_t$, resulting in an overfitting scenario on limited calibration data.

To further substantiate our analyses, we compare the above-mentioned reconstruction method, \emph{e.g.}, BRECQ~\cite{li2021brecq} with the approach where the parameters of the \verb|embedding layers| were frozen during the reconstruction of the Residual Bottleneck Block and initialized solely through Min-Max~\cite{nagel2021white}. As shown in Tab.~\ref{tab:recon_comp}, the Freeze strategy exhibits superior results, which verify that \verb|embedding layers| served as their own optimization objective and maintaining their independence of $\bx_t$ can significantly mitigate temporal feature disturbance, particularly at low-bit.
\begin{table}[!ht]\setlength{\tabcolsep}{10pt}
  \centering
  \caption{FID, sFID, and SQNR on LSUN-Bedrooms $256\times256$~\cite{yu2016lsun} for LDM-4. ``Prev'' represents BRECQ. Freeze denotes our trial. We conduct the same experiments with more baselines as ``Prev'' in Sec.~\ref{app:induce-vali}, which further validate our analysis.}
  \resizebox{0.9\linewidth}{!}{
  \begin{tabular}{lclll}
    \toprule
    \textbf{Methods} & \textbf{\#Bits (W/A)} & FID$\downarrow$ & sFID$\downarrow$ & SQNR$\uparrow$\\
    \midrule
    Full Prec. & 32/32 & 2.98 & 7.09 & -\\
    \midrule
    Prev & 8/8 & 7.51 &  12.54 & -0.12\\
    \rowcolor[gray]{0.92}Freeze & 8/8 & \textbf{5.76\textsubscript{\bl{-1.75}}} & \textbf{8.42\textsubscript{\bl{-4.12}}} & \textbf{4.07\textsubscript{\mg{+4.19}}}\\
    \midrule
    Prev & 4/8  & 9.36 & 22.73 & -1.61\\
    \rowcolor[gray]{0.92}Freeze & 4/8 & \textbf{7.08\textsubscript{\bl{-2.28}}} & \textbf{16.82\textsubscript{\bl{-5.91}}} & \textbf{3.42\textsubscript{\mg{+5.03}}}\\
    \bottomrule
\end{tabular}
}
    \label{tab:recon_comp}
\end{table}

\vspace{1em}
\noindent\textbf{Unaware of finite activations within $h(\cdot)$ and $g_i(\cdot)$.} We observe that, given $T$ as a finite positive integer, the set of all possible activation values for \verb|embedding layers| and \verb|time embed| is finite and strictly dependent on timesteps. Within this set, activations corresponding to the same layer display notable range variations across different timesteps~\footnote{The details of range variations can be found in Sec. B of \cite{Huang_2024_CVPR}}. Previous methods~\cite{so2023temporal, wang2023towards} mainly focus on finding the optimal calibration method for $\widehat\bx_t$-related network components. Moreover, akin to the first inducement, their calibration is directly towards the Residual Bottleneck Block, which proves suboptimal~\footnote{The evidence can be found in Sec. C of \cite{Huang_2024_CVPR}}. However, based on the finite activations, we can employ calibration methods, especially for these time-related activations, to better adapt to their range variations. %

\subsection{Quantization Framework}\label{sec:quantization_framework}
To mitigate temporal feature disturbance, we propose two different temporal feature maintenance strategies categorized into TIB-based and cache-based ones in Sec.~\ref{sec:tib-based} and Sec.~\ref{sec:cache-based}, respectively. Finally, we incorporate both scenarios together with our selection strategy to further solve the problem in Sec.~\ref{sec:select}. Our methods are outlined in Fig.~\ref{fig:framework}.
\subsubsection{TIB-based Maintenance}\label{sec:tib-based}
Referring to the two inducements aforementioned, we define a novel Temporal Information Block~(TIB) to maintain the temporal features. Built on the block, Temporal Information-aware Reconstruction and Finite Set Calibration are proposed to solve the two inducements analyzed above.

\vspace{1em}
\noindent\textbf{Temporal information block~(TIB).} Based on the inducements, it is crucial to meticulously separate the reconstruction and calibration process for each \verb|embedding layer| and Residual Bottleneck Block to enhance quantized model performance. Considering the unique structure of the UNet, we consolidate all \verb|embedding layers| and \verb|time embed| into a unified Temporal Information Block~(TIB), which can be denoted as $\{g_i\}_{i=1,\ldots, n}\cup\{h\}$~(see Fig. \ref{fig:framework}~(a)).

\vspace{1em}
\noindent\textbf{Temporal information-aware reconstruction.} Based on the Temporal Information Block~(TIB), we introduce the Temporal Information-aware Reconstruction~(TIAR) to address the first inducement. The optimization goal for the block during the reconstruction phase is captured by the following objective:
\begin{equation}
     \mathop{\arg\,\min}\limits_{\mathcal{S}^{\mathbf{W}}_{g\circ h}, \mathcal{Z}^{\mathbf{W}}_{g\circ h}}\sum_{i=1}^{n} \underbrace{\| g_i(h(t)) -  \widehat{g}_i(\widehat{h}(t)) \|_{F}^2}_{\mathcal{L}^{temporal}_{t,i}},
    \label{time_embedding_block_lossfunction}
\end{equation}
where $\widehat{h}(\cdot)$ and $\widehat{g}_i(\cdot)$ represent the quantized counterparts of $h(\cdot)$ and $g_i(\cdot)$, respectively. Moreover, $\mathcal{S}^{\mathbf{W}}_{g\circ h}$ and $\mathcal{Z}^{\mathbf{W}}_{g\circ h}$ denote all the quantization scales and zero offsets for every weight of linear operators~\footnote{Linear layers and convolutions.} in $\{g_i\}_{i=1,\ldots,n}\cup\{h\}$, respectively. This reconstruction strategy focuses on adjusting weights to pursue a minimal disturbance for temporal features.

\vspace{1em}
\noindent\textbf{Finite set calibration.}  To address the challenge posed by the wide span of activations within a finite set for the second inducement, we propose Finite Set Calibration~(FSC) for activation quantization. This strategy employs $T$ sets of quantization parameters for activation, such as $\{(s_T, z_T), \ldots, (s_1, z_1)\}$ for an arbitrary activation $\bX$ within \texttt{embedding layers} and \texttt{time embed}. At timestep $t$, the quantization function for the $\bX$ can be expressed as:
\begin{equation}
    \hat{\bX} =\Phi(\left\lfloor{\frac{\bX}{s_{t}}}\right\rceil+z_{t}, 0,2^{b}-1),
    \label{fsc}
\end{equation}
where $s_t$ and $z_t$ are the corresponding scale and zero offset. Equipped with FSC, we can obtain $\mathcal{S}^{\mathbf{X}}_{g\circ h}$ and $\mathcal{Z}^{\mathbf{X}}_{g\circ h}$, which refer to all the quantization scales and zero offsets for the activations~\footnote{We consider the outputs and inputs of every linear operator like \cite{shang2022ptq4dm, he2023ptqd}.} in $\{g_i\}_{i=1,\ldots,n}\cup\{h\}$, respectively. To be noted, the calibration target for these activations is also aligned with the output of the TIB. Besides that, we find that Min-max~\cite{nagel2021white} can achieve satisfactory results with high efficiency~(more evidence in Sec.~\ref{sec:cali-tib}) for range estimation. In Sec.~\ref{sec:efficiency-tib}, we perform a detailed analysis to demonstrate that our method incurs a negligible extra cost for inference.

\subsubsection{Cache-based Maintenance}\label{sec:cache-based}
Further considering sample independence and finitude of the temporal feature, the feature associated with each $t$ and $i$ remains constant and can therefore be pre-computed offline. This allows us to directly optimize the quantization parameters for these pre-computed full-precision features and cache the quantized counterparts with the parameters to address the issues. The objective for all $t$ and $i$ can be formulated as follows:
\begin{equation}
    \mathop{\arg\,\min}\limits_{s_{t,i}, z_{t,i}} \underbrace{\|\widehat{\bX_{t,i}^{temporal}}-\bX_{t,i}^{temporal}\|^2_F}_{\mathcal{L}^{temporal}_{t,i}},
    \label{eq:cache}
\end{equation}
where $s_{t,i}$ and $z_{t,i}$ denote the quantization scale and zero offset for the pre-gained $\bX_{t,i}^{temporal}$ (\emph{i.e.}, the output of $g_i\circ h$ at $t$). Finally, the obtained quantization parameters can be defined as $\mathcal{S}^{temporal}=\{s_{t, i}\}_{t=1,\ldots, T \& i=1,\ldots, n}$ and $\mathcal{Z}^{temporal}=\{z_{t, i}\}_{t=1,\ldots, T \& i=1,\ldots, n}$. Specifically, in this strategy, we employ LSQ~\cite{esser2020lsq} to optimize every objective~\footnote{From Eq.~(\ref{eq:cache}), it is obvious that we have to optimize $n\times T$ objectives across different indexes of temporal features.} separately, which can be done in just a few minutes. During inference, this strategy enables us to get the temporal feature by reloading the cached items. In addition, we find that this approach can help improve latency in cases. More detailed analysis can be found in Sec.~\ref{sec:efficiency-cache}.

For clarity, we demonstrate the difference between the two proposed maintenance strategies. In TIB-based Maintenance, we optimize quantization parameters of weights and activations for $\{g_i\}_{i=1,\ldots,n}\cup\{h\}$. In contrast, we only adjust the quantization parameters of the final output activations of $\{g_i\}_{i=1,\ldots,n}\cup\{h\}$ in Cache-based Maintenance, which implies that $\mathcal{S}^{temporal}\in\mathcal{S}^{\mathbf{X}}_{g\circ h}$ and $\mathcal{Z}^{temporal}\in\mathcal{Z}^{\mathbf{X}}_{g\circ h}$. Thus, it is evident that Eq.~(\ref{time_embedding_block_lossfunction}) and Eq.~(\ref{eq:cache}) have the same objective (\emph{i.e.}, minimize $\mathcal{L}_{t, i}^{temoral}$ at corresponding $t$ and $i$), but optimize different sets of parameters in different ways.

\subsubsection{Disturbance-aware Selection}\label{sec:select}
\begin{figure}[!ht]
    \centering
    \setlength{\abovecaptionskip}{0.2cm}
     \includegraphics[width=0.48\textwidth]{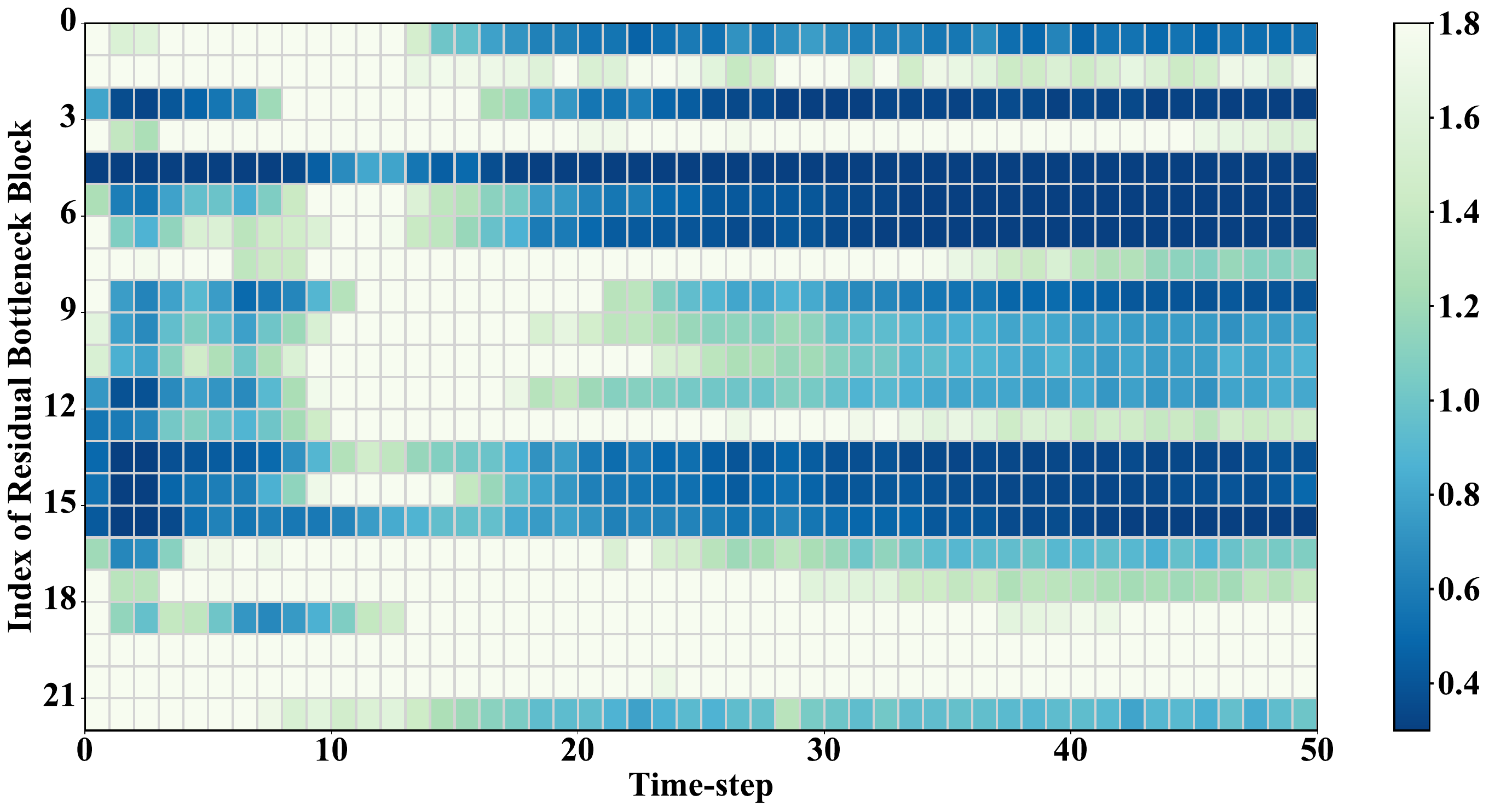}
     \caption{Comparison of the temporal feature disturbance between TIB-based and Cache-based Maintenance with W$4$A$8$ quantized Stable Diffusion~($T=50$).}
    \label{fig:select-vis}
\end{figure}
\begin{figure}[!ht]
\centering
\setlength{\abovecaptionskip}{0.2cm}
\renewcommand{\arraystretch}{0.5}
\begin{tabular*}{\linewidth}{@{\extracolsep{\fill}}c}
\toprule
\footnotesize{\shortstack{\textit{``A small, neat, simple kitchen with lots of cupboards and a small work table} \\\textit{ in the middle of the room.''}}} \\
\midrule
\includegraphics[width=0.488\textwidth]{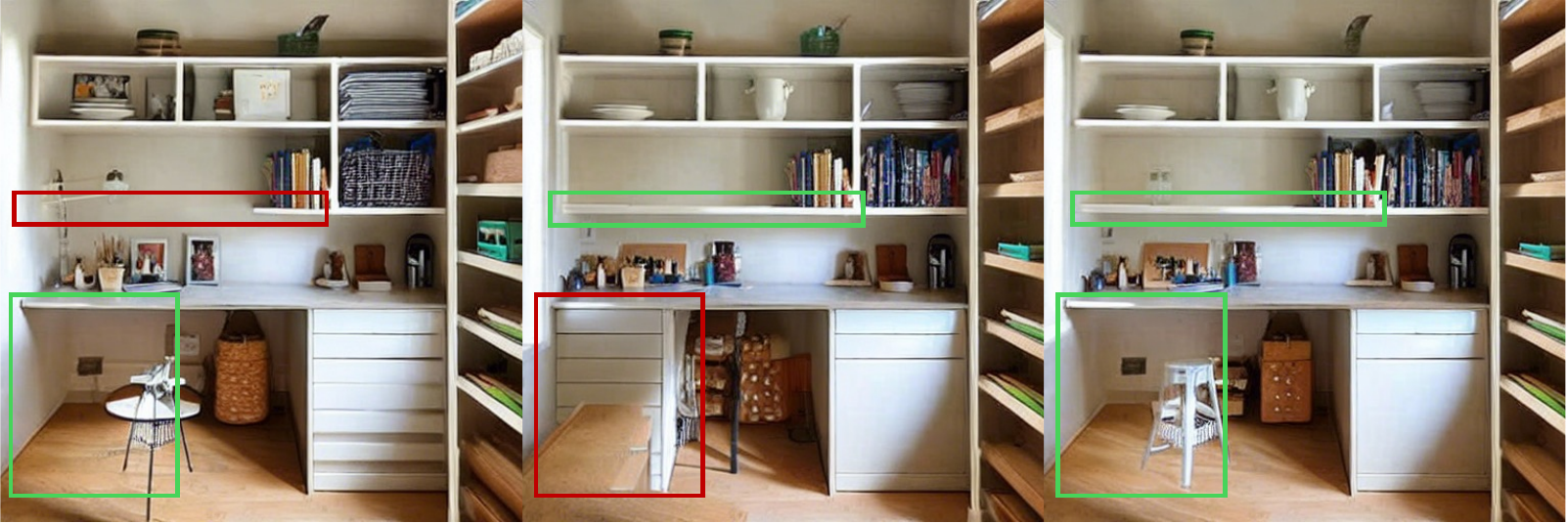}\\
\midrule
\footnotesize{13.36 \hspace{2.2cm} 13.31 \hspace{2.2cm} 13.29}\\
\bottomrule
\end{tabular*}
\caption{\label{fig:select-comp} Samples from W$4$A$8$ quantized Stable Diffusion on MS-COCO~\cite{lin2015microsoft} caption with TIB-based Maintenance~(Left), Cache-based Maintenance~(Middle), and Disturbance-aware Selection~(Right), where \gre{green rectangles} present finer points than \re{red rectangles}. Numbers in the last row denote FID on the MS-COCO caption of the corresponding methods.
}
\end{figure}

We visualize the comparison of the temporal feature disturbance between the two maintenance approaches in Fig.~\ref{fig:select-vis}, where the grid value of $(t, i)$ represents the ratio of temporal feature error\footnote{Here we employ $\mathcal{L}_{t,i}^{temporal}$ in Eq.~(\ref{eq:cache}), instead of that in Eq.~(\ref{eq:temporal_feature_error}) as temporal feature error, since over $82\%$ of cosine similarity with our two maintenance approaches are almost exactly equal to $1$, which shows the remarkable effect of our methods.} between the two approaches as follows:
\begin{equation}
    \tau_{t,i}=\frac{\text{$\mathcal{L}_{t,i}^{temporal}$ with TIB-based Maintenance}}{\text{$\mathcal{L}_{t,i}^{temporal}$ with Cache-based Maintenance}}.
    \label{eq:select}
\end{equation}
We believe the discrepancy in temporal feature errors across different $t$ and $i$ between our two methods results from their different sets of optimization parameters and optimization methods. Based on the above observation, we propose Disturbance-aware Selection to better mitigate the temporal feature disturbance. Specifically, when $\tau_{t,i} < 1$, we employ TIB-based Maintenance to obtain $\widehat{\bX_{t,i}^{temporal}}$. Conversely, we opt for Cache-based Maintenance. The pipeline of our selection is shown in Fig.~\ref{fig:framework} (c). For efficiency, our selection procedure can be done offline (\emph{i.e.}, before deployment) with only one image generation pass to calculate all the $\tau_{t,i}$. To validate its performance, Fig.~\ref{fig:select-comp} demonstrates that our straightforward but effective selection strategy successfully leverages the strengths of both maintenance approaches.

\begin{table*}[!ht]\setlength{\tabcolsep}{5pt}
   \centering
  \caption{Quantization results for unconditional image generation with LDM-4 on LSUN-Bedrooms $256\times 256$, FFHQ $256\times 256$ and CelebA-HQ $256 \times 256$, LDM-8 on LSUN-Churches $256 \times 256$. ``*'' represents our implementation according to open-source codes and ``\dag'' means directly rerunning open-source codes. The subscript numbers represent the improvements from our framework compared with the previous methods. \bl{Blue}/\mg{Green} color is to represent the negative/positive number. ``TM" is TIB-based Maintenance, ``CM" is Cache-based Maintenance, and ``DS" denotes our Disturbance-aware Selection.} 
  \resizebox{\linewidth}{!}{
  \begin{tabular}{lcllllllllllll}
    \toprule
    \multicolumn{1}{c}{\multirow{2}{*}{\textbf{Methods}}} & \multicolumn{1}{c}{\multirow{2}{*}{\textbf{\#Bits (W/A)}}} & \multicolumn{3}{c}{\textbf{LSUN-Bedrooms $256 \times 256$}} & \multicolumn{3}{c}{\textbf{LSUN-Churches $256 \times 256$}} & \multicolumn{3}{c}{\textbf{CelebA-HQ $256 \times 256$}} & \multicolumn{3}{c}{\textbf{FFHQ $256 \times 256$}}\\ \cmidrule(r){3-5} \cmidrule(r){6-8} \cmidrule(r){9-11} \cmidrule(r){12-14}
    & & FID$\downarrow$ & sFID$\downarrow$ & SQNR$\uparrow$ & FID$\downarrow$ & sFID$\downarrow$ & SQNR$\uparrow$ & FID$\downarrow$ & sFID$\downarrow$ & SQNR$\uparrow$ & FID$\downarrow$ & sFID$\downarrow$ & SQNR$\uparrow$\\
    \midrule
    Full Prec. & 32/32 & 2.98 &  7.09 & - & 4.12 & 10.89 & - & 8.74 & 10.16 & - & 9.36 & 8.67 & -\\
    \midrule
     PTQ4DM*~\cite{shang2022ptq4dm} & 4/32 & 4.83 &  7.94 & 4.47 & 4.92 & 13.94 & 0.62 & 13.67 & 14.72 & 4.18 & 11.74 & 12.18 & -2.87 \\
    Q-Diffusion\dag~\cite{li2023qdiffusion} & 4/32 & 4.20 &  7.66 & 5.11 & 4.55 & 11.90 & 1.22 & 11.09 & 12.00 & 5.67 & 11.60 & 10.30 & -2.04 \\
    PTQD*~\cite{he2023ptqd} & 4/32 & 4.42 &  7.88 & 4.89 & 4.67 & 13.68 & 1.15 & 11.06 & 12.21 & 5.52 & 12.01 & 11.12 & -2.36 \\
    \rowcolor[gray]{0.92}TM & 4/32 & 3.60\textsubscript{~\bl{-0.60}} &  7.61\textsubscript{~\bl{-0.05}} & 8.45\textsubscript{~\mg{+3.34}} & 4.07\textsubscript{~\bl{-0.48}} & 
    11.41\textsubscript{~\bl{-0.49}} & 2.21\textsubscript{~\mg{+0.99}} & 8.74\textsubscript{~\bl{-2.32}} & 10.18\textsubscript{~\bl{-1.82}} & 6.93\textsubscript{~\mg{+1.26}}& 9.89\textsubscript{~\bl{-1.71}} & 9.06\textsubscript{~\bl{-1.24}} & -1.98\textsubscript{~\mg{+0.06}}\\
    
    \rowcolor[gray]{0.92}CM & 4/32 & 3.58\textsubscript{~\bl{-0.62}} &  7.42\textsubscript{~\bl{-0.24}} & 7.72\textsubscript{~\mg{+2.61}}& 4.10\textsubscript{~\bl{-0.45}} & 11.23\textsubscript{~\bl{-0.62}} & 2.18\textsubscript{~\mg{+0.96}}& 8.73\textsubscript{~\bl{-2.33}} & 10.15\textsubscript{~\bl{-1.85}} & 7.08\textsubscript{~\mg{+1.41}}& 9.91\textsubscript{~\bl{-1.69}} & 9.03\textsubscript{~\bl{-1.27}} & -1.89\textsubscript{~\mg{+0.15}}\\
    \rowcolor[gray]{0.92}DS & 4/32 & \textbf{3.41\textsubscript{~\bl{-0.79}}} &  \textbf{7.40\textsubscript{~\bl{-0.26}}} & \textbf{9.01\textsubscript{~\mg{+3.90}}}& \textbf{4.03\textsubscript{~\bl{-0.52}}} & \textbf{10.88\textsubscript{~\bl{-1.02}}} & \textbf{2.41\textsubscript{~\mg{+1.19}}}& \textbf{8.72\textsubscript{~\bl{-2.34}}} & \textbf{10.12\textsubscript{~\bl{-1.88}}} & \textbf{7.15\textsubscript{~\mg{+1.48}}}& \textbf{9.76\textsubscript{~\bl{-1.84}}} & \textbf{8.86\textsubscript{~\bl{-1.44}}}& \textbf{-1.87\textsubscript{~\mg{+0.17}}}\\
    \midrule
    
    PTQ4DM*~\cite{shang2022ptq4dm} & 8/8 & 4.75 &  9.59 & 5.16 & 4.80 & 13.48 & 1.20& 14.42 & 15.06 & 5.31 & 10.73 & 11.65& -1.80\\
    Q-Diffusion\dag~\cite{li2023qdiffusion} & 8/8 & 4.51 &  8.17 & 5.21& 4.41 & 12.23 & 1.62 & 12.85 & 14.16 & 6.01& 10.87 & 10.01& -1.72\\
    PTQD~\cite{he2023ptqd} & 8/8 & 3.75 &  9.89 & 6.60* & 4.89* & 14.89* & 1.51* & 12.76* & 13.54* & 5.87* & 10.69* & 10.97*& -1.75* \\
    \rowcolor[gray]{0.92}TM & 8/8 & 3.14\textsubscript{~\bl{-0.61}} &  7.26\textsubscript{~\bl{-0.91}} & 9.12\textsubscript{~\mg{+2.52}}& 4.01\textsubscript{~\bl{-0.40}} & 10.98\textsubscript{~\bl{-1.25}} & 2.53\textsubscript{~\mg{+0.91}}& \textbf{8.71\textsubscript{~\bl{-4.05}}} & 10.20\textsubscript{~\bl{-3.34}} & 7.21\textsubscript{~\mg{+1.20}}& 9.46\textsubscript{~\bl{-1.23}} & 8.73\textsubscript{~\bl{-1.28}}& -1.68\textsubscript{~\mg{+0.04}}\\
    \rowcolor[gray]{0.92}CM & 8/8 & 3.11\textsubscript{~\bl{-0.64}} &  7.12\textsubscript{~\bl{-1.05}} & 9.43\textsubscript{~\mg{+2.83}}& 4.11\textsubscript{~\bl{-0.30}} & 10.82\textsubscript{~\bl{-1.41}} & 2.47\textsubscript{~\mg{+0.85}}& \textbf{8.71\textsubscript{~\bl{-4.05}}} & 10.18\textsubscript{~\bl{-3.36}} & 7.23\textsubscript{~\mg{+1.22}}& 9.41\textsubscript{~\bl{-1.28}} & 8.69\textsubscript{~\bl{-1.32}}& -1.66\textsubscript{~\mg{+0.06}}\\
    \rowcolor[gray]{0.92}DS & 8/8 & \textbf{3.08\textsubscript{~\bl{-0.67}}} &  \textbf{7.10\textsubscript{~\bl{-1.07}}} & \textbf{9.70\textsubscript{~\mg{+3.10}}}& \textbf{3.97\textsubscript{~\bl{-0.44}}} & \textbf{10.78\textsubscript{~\bl{-1.45}}} & \textbf{2.60\textsubscript{~\mg{+0.98}}}& \textbf{8.71\textsubscript{~\bl{-4.05}}} & \textbf{10.16\textsubscript{~\bl{-3.38}}} & \textbf{7.34\textsubscript{~\mg{+1.33}}}& \textbf{9.35\textsubscript{~\bl{-1.34}}} & \textbf{8.63\textsubscript{~\bl{-1.38}}}& \textbf{-1.63\textsubscript{~\mg{+0.09}}}\\
    \midrule
    
    PTQ4DM~\cite{shang2022ptq4dm} & 4/8 & 20.72 &  54.30 & 1.42* & 4.97* & 14.87* & -0.34* & 17.08* & 17.48* & 1.02* & 11.83* & 12.91*& -3.44*\\
    Q-Diffusion\dag~\cite{li2023qdiffusion} & 4/8 & 6.40 &  17.93 & 3.89 & 4.66 & 13.94 & 0.46 & 15.55 & 16.86 & 2.12 & 11.45 & 11.15& -2.94 \\
    PTQD~\cite{he2023ptqd} & 4/8 & 5.94 &  15.16 & 4.42* & 5.10* & 13.23* & -0.25* & 15.47* & 17.38* & 3.31* & 11.42* & 11.43*& -3.18* \\
    \rowcolor[gray]{0.92}TM & 4/8 & 3.68\textsubscript{~\bl{-2.26}} &  7.65\textsubscript{~\bl{-7.51}} & 8.02\textsubscript{~\mg{+3.60}}& 4.14\textsubscript{~\bl{-0.52}} & 11.46\textsubscript{~\bl{-1.77}} & 1.97\textsubscript{~\mg{+1.51}}& 8.76\textsubscript{~\bl{-6.71}} & 10.26\textsubscript{~\bl{-6.60}} & 6.78\textsubscript{~\mg{+3.47}}& 9.97\textsubscript{~\bl{-1.45}} & 9.14\textsubscript{~\bl{-2.01}}& -2.70\textsubscript{~\mg{+0.24}}\\
    \rowcolor[gray]{0.92}CM & 4/8 & 3.91\textsubscript{~\bl{-2.03}} &  8.61\textsubscript{~\bl{-6.55}} & 
    7.30\textsubscript{~\mg{+2.88}}& 4.46\textsubscript{~\bl{-0.20}} & 11.39\textsubscript{~\bl{-1.84}} & 1.88\textsubscript{~\mg{+1.42}}& 8.76\textsubscript{~\bl{-6.71}} & 10.18\textsubscript{~\bl{-6.68}} & 6.71\textsubscript{~\mg{+3.40}}& 10.03\textsubscript{~\bl{-1.39}} & 9.11\textsubscript{~\bl{-2.04}}&
    -2.71\textsubscript{~\mg{+0.23}}\\
    \rowcolor[gray]{0.92}DS & 4/8 & \textbf{3.61\textsubscript{~\bl{-2.33}}} &  \textbf{7.49\textsubscript{~\bl{-7.67}}} & \textbf{8.51\textsubscript{~\mg{+4.09}}}& \textbf{4.13\textsubscript{~\bl{-0.53}}} & \textbf{10.95\textsubscript{~\bl{-2.28}}} & \textbf{2.13\textsubscript{~\mg{+1.67}}}& \textbf{8.73\textsubscript{~\bl{-6.74}}} & \textbf{10.07\textsubscript{~\bl{-6.79}}} & \textbf{6.80\textsubscript{~\mg{+3.49}}}& \textbf{9.81\textsubscript{~\bl{-1.61}}} & \textbf{9.10\textsubscript{~\bl{-2.05}}}& \textbf{-2.68\textsubscript{~\mg{+0.26}}}\\
    \bottomrule
\end{tabular}
}
    \label{tab:sota_ldm}
\end{table*}

\section{Experiments}  
We organize the experiments as follows: In Sec.~\ref{sec:imple}, we first demonstrate implementation details. Then, we exhibit the outstanding performance gained by our framework in Sec.~\ref{sec:perform}. Further, we also deploy our quantized models on various hardware to investigate their inference efficiency in Sec.~\ref{sec:deploy}. Last, we conduct comprehensive ablative studies for the proposed framework in Sec.~\ref{sec:ablation}. \yushi{It is worth noting that we also discuss and conduct experiments for video generation in Sec.~\ref{app:video}. A guideline on when to use which proposed strategies and sub-$4$-bit quantization results are included in Sec.~\ref{app:guide} and Sec.~\ref{app:sub-4}, respectively.}

\subsection{Implementation Details}\label{sec:imple}
\vspace{1em}
\noindent\textbf{Models and datasets.} In this section, we conduct image generation experiments to evaluate the proposed quantization framework on various diffusion models: pixel-space diffusion model DDPM~\cite{ho2020denoising} for unconditional image generation, latent-space diffusion model LDM~\cite{rombach2022ldm} for unconditional image generation and class-conditional image generation. We also apply our work to Stable Diffusion-v1-4~\cite{rombach2022ldm}, SD-XL-base-1.0~\cite{podell2023sdxlimprovinglatentdiffusion}, and SD-XL-turbo~\cite{sauer2023adversarial} for text-guided image generation. In our experiments, We use seven standard benchmarks: CIFAR-10 $32 \times 32$~\cite{krizhevsky2009learning}, LSUN-Bedrooms $256 \times 256$~\cite{yu2016lsun}, LSUN-Churches $256 \times 256$~\cite{yu2016lsun}, CelebA-HQ $256 \times 256$~\cite{karras2018progressive}, ImageNet $256 \times 256$~\cite{deng2009imagenet}, FFHQ $256 \times 256$~\cite{karras2019stylebased}, and MS-COCO~\cite{lin2015microsoft}.

\vspace{1em}
\noindent\textbf{Quantization settings.} We use channel-wise quantization for weights and tensor-wise quantization for
activations, as it is a common practice. In our experimental setup, we employ BRECQ~\cite{li2021brecq} and AdaRound~\cite{nagel2020adaround}. Drawing from empirical insights derived from conventional model quantization practices~\cite{10.1007/978-3-319-46493-0_32, NIPS2015_ae0eb3ee}, we maintain the input and output layers of the model in full precision. Mirroring the details outlined in Q-Diffusion~\cite{li2023qdiffusion}~\footnote{This can be found in the appendix of \cite{li2023qdiffusion}. For SD-XL and SD-XL-turbo, we collect $8$ and $1024$ COCO prompts, respectively.}, we generate calibration data through full-precision diffusion models. Moreover, for weight quantization, we reconstruct quantized weights for $20K$ iterations with a mini-batch size of $32$ for DDPM and LDM, $8$ for Stable Diffusion and SD-XL-turbo, and $4$ for SD-XL. For activation quantization, we utilize EMA~\cite{jacob2018quantization} to estimate the ranges of activations, also equipping the paradigm from~\cite{so2023temporal, he2024efficientdm, wang2024quest}. This stage applies a mini-batch size of $16$ for all models. To be noted, we follow the block-wise reconstruction in \cite{li2023qdiffusion, he2023ptqd}. For Cache-based Maintenance, we employ LSQ~\cite{bhalgat2020lsq} for $10K$ iterations to quantize pre-fetched features in $8$-bit for all configurations, including weight-only quantization.

\vspace{1em}
\noindent\textbf{Evaluation metrics.} For each experiment, we evaluate the performance of diffusion models with Fréchet Inception Distance (FID)~\cite{heusel2018gans}, and Signal-to-quantization-noise Ratio~(SQNR)~\cite{pandey2023softmax}, which measures the quantization distortion in detail. In the case of LDM and text-guided generation experiments, we also include sFID~\cite{salimans2016improved}, which better captures spatial relationships than FID. For ImageNet and CIFAR-10 experiments, we additionally provide Inception Score~(IS)~\cite{salimans2016improved} as a reference metric. Further, in the context of text-guided experiments, we extend our evaluation to include the compatibility of image-caption pairs, employing the CLIP score~\cite{hessel2022clipscore}. The ViT-B/32~\cite{dosovitskiy2020image} is used as the backbone when computing the CLIP score. To ensure consistency in the reported outcomes,  all results are derived from our implementation or from other papers, where experiments are conducted under conditions consistent with ours. More specifically, in the evaluation process of each experiment, we sample $50K$ images from DDPM or LDM, $30K$ images from Stable Diffusion or SD-XL-turbo, and $10K$ images from SD-XL. All experiments are conducted utilizing one $H800$ GPU and implemented with
the PyTorch framework~\cite{paszke2019pytorch} without special claims.

\subsection{Performance Comparison}\label{sec:perform}

\begin{table*}[!ht]
  \centering
  \begin{minipage}[b]{0.455\linewidth}\setlength{\tabcolsep}{8pt}
  \caption{Quantization results for unconditional image generation with DDIM on CIFAR-10 $32 \times 32$.} 
  \resizebox{\linewidth}{!}{
  \begin{tabular}{lclll}
    \toprule
    \multicolumn{1}{c}{\multirow{2}{*}{\textbf{Methods}}} & \multicolumn{1}{c}{\multirow{2}{*}{\textbf{\#Bits (W/A)}}} & \multicolumn{3}{c}{\textbf{CIFAR-10 $32 \times 32$}} \\ \cmidrule(r){3-5}
    & & IS$\uparrow$ & FID$\downarrow$ & SQNR$\uparrow$ \\
    \midrule
    Full Prec. & 32/32 & 9.04 & 4.23 & -\\
    \midrule
     PTQ4DM*~\cite{shang2022ptq4dm} & 4/32 & 9.02 & 5.65 & 3.68\\
    Q-Diffusion\dag~\cite{li2023qdiffusion} & 4/32 & 8.78 & 5.08 & 4.12\\
    \rowcolor[gray]{0.92}TM & 4/32 & 9.14\textsubscript{~\mg{+0.12}} & 4.73\textsubscript{~\bl{-0.35}} & 5.03\textsubscript{~\mg{+0.91}}\\
     \rowcolor[gray]{0.92}CM & 4/32 & 9.16\textsubscript{~\mg{+0.14}} & 4.68\textsubscript{~\bl{-0.40}} & 5.21\textsubscript{~\mg{+1.09}}\\
      \rowcolor[gray]{0.92}DS & 4/32 & \textbf{9.21\textsubscript{~\mg{+0.19}}} & \textbf{4.49\textsubscript{~\bl{-0.59}}} & \textbf{5.29\textsubscript{~\mg{+1.17}}}\\
    \midrule
    
    PTQ4DM~\cite{shang2022ptq4dm} & 8/8 & 9.02 & 19.59 & 4.12*\\
    Q-Diffusion\dag~\cite{li2023qdiffusion} & 8/8 & 8.89 & 4.78 & 4.78\\
    TDQ~\cite{so2023temporal} & 8/8 & 8.85 & 5.99 & --\\
    \rowcolor[gray]{0.92}TM & 8/8 & 9.07\textsubscript{~\mg{+0.05}} & 4.24\textsubscript{~\bl{-0.54}} & 5.67\textsubscript{~\mg{+0.89}}\\
    \rowcolor[gray]{0.92}CM & 8/8 & 9.05\textsubscript{~\mg{+0.03}} & 4.25\textsubscript{~\bl{-0.53}} & 5.95\textsubscript{~\mg{+1.17}}\\
    \rowcolor[gray]{0.92}DS & 8/8 & \textbf{9.08\textsubscript{~\mg{+0.06}}} & \textbf{4.18\textsubscript{~\bl{-0.61}}} & \textbf{5.98\textsubscript{~\mg{+1.20}}}\\
    \midrule
    
    PTQ4DM*~\cite{shang2022ptq4dm} & 4/8 & 8.93 & 5.14 & 1.96\\
    Q-Diffusion\dag~\cite{li2023qdiffusion} & 4/8 & 9.12 & 4.98 & 2.07\\
    \rowcolor[gray]{0.92}TM & 4/8 & 9.13\textsubscript{~\mg{+0.01}} & 4.78\textsubscript{~\bl{-0.20}} & 3.43\textsubscript{~\mg{+1.36}}\\
    \rowcolor[gray]{0.92}CM & 4/8 & 9.13\textsubscript{~\mg{+0.01}} & 4.59\textsubscript{~\bl{-0.39}} & 3.22\textsubscript{~\mg{+1.15}}\\
    \rowcolor[gray]{0.92}DS & 4/8 &\textbf{9.15\textsubscript{~\mg{+0.03}}} & \textbf{4.46\textsubscript{~\bl{-0.52}}} & \textbf{3.51\textsubscript{~\mg{+1.44}}}\\
    \bottomrule
\end{tabular}
}
    \label{tab:sota_ddim}
    \end{minipage}\hfill
    \begin{minipage}[b]{0.495\linewidth}\setlength{\tabcolsep}{5pt}
    \caption{Quantization results for class-conditional image generation with LDM-4 on ImageNet $256 \times 256$.} 
  \resizebox{\linewidth}{!}{
  \begin{tabular}{lcllll}
    \toprule
    \multicolumn{1}{c}{\multirow{2}{*}{\textbf{Methods}}} & \multicolumn{1}{c}{\multirow{2}{*}{\textbf{\#Bits (W/A)}}} & \multicolumn{4}{c}{\textbf{ImageNet $256 \times 256$}} \\ \cmidrule(r){3-6}
    & & IS$\uparrow$ & FID$\downarrow$ & sFID$\downarrow$ & SQNR$\uparrow$\\
    \midrule
    Full Prec. & 32/32 & 235.64 & 10.91 & 7.67 & -\\
    \midrule
    Q-Diffusion*~\cite{li2023qdiffusion} & 4/32 & 213.56 & 11.87 & 8.76 & 6.16\\
    PTQD\dag~\cite{he2023ptqd} & 4/32 & 201.78 & 11.65 & 9.06 & 5.97\\
    \rowcolor[gray]{0.92}TM & 4/32 & 223.81\textsubscript{~\mg{+10.25}} & 10.50\textsubscript{~\bl{-1.15}} & 7.98\textsubscript{~\bl{-0.78}} & \textbf{8.64\textsubscript{~\mg{+2.48}}} \\
    \rowcolor[gray]{0.92}CM & 4/32 & 228.46\textsubscript{~\mg{+13.90}} & 10.42\textsubscript{~\bl{-1.23}} & 8.01\textsubscript{~\bl{-0.75}} & 8.21\textsubscript{~\mg{+2.05}} \\
    \rowcolor[gray]{0.92}DS & 4/32 & \textbf{234.72\textsubscript{~\mg{+20.16}}} & \textbf{10.38\textsubscript{~\bl{-1.27}}} & \textbf{7.81\textsubscript{~\bl{-0.95}}} & 8.58\textsubscript{~\mg{+2.42}} \\
    \midrule
    
    PTQ4DM~\cite{shang2022ptq4dm} & 8/8 & 161.75 & 12.59 & - & -\\
    Q-Diffusion*~\cite{li2023qdiffusion} & 8/8 & 187.65 & 12.80 & 9.87 & 4.89\\
    PTQD~\cite{he2023ptqd} & 8/8 & 153.92 & 11.94 & 8.03 & 5.41\\
    \rowcolor[gray]{0.92}TM & 8/8 & 198.86\textsubscript{~\mg{+11.21}} & 10.79\textsubscript{~\bl{-1.15}} & 7.65\textsubscript{~\bl{-0.38}} & 8.90\textsubscript{~\mg{+4.49}}\\
    \rowcolor[gray]{0.92}CM & 8/8 & 204.41\textsubscript{~\mg{+16.76}} & 9.99\textsubscript{~\bl{-1.95}} & 7.54\textsubscript{~\bl{-0.49}} & 9.03\textsubscript{~\mg{+0.13}}\\
    \rowcolor[gray]{0.92}DS & 8/8 & \textbf{206.12\textsubscript{~\mg{+18.47}}} & \textbf{9.85\textsubscript{~\bl{-2.09}}} & \textbf{7.52\textsubscript{~\bl{-0.51}}} & \textbf{9.12\textsubscript{~\mg{+0.22}}}\\
    \midrule
    
    Q-Diffusion*~\cite{li2023qdiffusion} & 4/8 & 212.51 & 10.68 & 14.85 & 4.02\\
    PTQD~\cite{he2023ptqd} & 4/8 & 214.73 & 10.40 & 12.63 & 3.89\\
    \rowcolor[gray]{0.92}TM & 4/8 & 221.82\textsubscript{~\mg{+7.09}} & 10.29\textsubscript{~\bl{-0.11}} & 7.35\textsubscript{~\bl{-5.28}} &
    6.78\textsubscript{~\mg{+2.76}}\\
    \rowcolor[gray]{0.92}CM & 4/8 & 220.01\textsubscript{~\mg{+5.28}} & 9.73\textsubscript{~\bl{-0.67}} & 7.32\textsubscript{~\bl{-5.31}} &
    7.01\textsubscript{~\mg{+2.99}}\\
    \rowcolor[gray]{0.92}DS & 4/8 & \textbf{223.97\textsubscript{~\mg{+9.24}}} & \textbf{9.46\textsubscript{~\bl{-0.94}}} & \textbf{7.28\textsubscript{~\bl{-5.35}}} &
    \textbf{7.03\textsubscript{~\mg{+3.01}}}\\
    \bottomrule
\end{tabular}
}
    \label{tab:sota_class}
    \end{minipage}
\end{table*}
\begin{table*}[!ht]\setlength{\tabcolsep}{5pt}
  \centering
  \caption{Quantization results for text-guided image generation with Stable Diffusion-v1-4 and SD-XL on MS-COCO captions.} 
  \resizebox{\linewidth}{!}{
  \begin{tabular}[t!]{lclllllllllllll}
    \toprule
    \multicolumn{1}{c}{\multirow{2}{*}{\textbf{Methods}}} & \multicolumn{1}{c}{\multirow{2}{*}{\textbf{\#Bits (W/A)}}} & \multicolumn{4}{c}{\textbf{Stable Diffusion-v1-4}} &  \multicolumn{4}{c}{\textbf{SD-XL-base-1.0}} &  \multicolumn{4}{c}{\textbf{SD-XL-turbo}} \\ \cmidrule(r){3-6} \cmidrule(r){7-10} \cmidrule(r){11-14}
    & & FID$\downarrow$ & sFID$\downarrow$ & CLIP$\uparrow$ & SQNR$\uparrow$ & FID$\downarrow$ & sFID$\downarrow$ & CLIP$\uparrow$ & SQNR$\uparrow$ & FID$\downarrow$ & sFID$\downarrow$ & CLIP$\uparrow$ & SQNR$\uparrow$\\
    \midrule
    Full Prec. & 32/32 & 13.15 & 19.31 & 31.46 & - & 34.10 & 57.03 & 31.33 & - & 18.21 & 28.70 & 31.74 & -\\
    \midrule
    Q-Diffusion\dag~\cite{li2023qdiffusion} & 4/32 & 13.58 & 19.50 & 31.43 & -2.62 & 28.62 & 60.34  & 30.88 & 1.39 & 21.75 & 30.64 & 31.01 & 0.02 \\
    \rowcolor[gray]{0.92}TM & 4/32 & \textbf{13.21\textsubscript{~\bl{-0.37}}} & 19.03\textsubscript{~\bl{-0.47}} &
    31.44\textsubscript{~\mg{+0.01}} &
    5.61\textsubscript{~\mg{+8.23}} & 25.82\textsubscript{~\bl{-2.80}} & 57.21\textsubscript{~\bl{-3.13}} &
   \textbf{31.12\textsubscript{\mg{+0.24}}}&
    3.74\textsubscript{~\mg{+2.35}} & 18.25\textsubscript{~\bl{-3.50}} & 29.56\textsubscript{~\bl{-1.08}} &
   31.01\textsubscript{\mg{+0.00}}&
    1.87\textsubscript{~\mg{+1.85}}\\

    \rowcolor[gray]{0.92}CM & 4/32 & 13.22\textsubscript{~\bl{-0.36}} & \textbf{19.01\textsubscript{~\bl{-0.49}}} &
    31.44\textsubscript{~\mg{+0.01}} &
    5.60\textsubscript{~\mg{+8.22}} & 26.13\textsubscript{~\bl{-2.49}} & 56.71\textsubscript{~\bl{-3.63}} &
   31.11\textsubscript{\mg{+0.23}}&
    3.21\textsubscript{~\mg{+1.82}} & 18.56\textsubscript{~\bl{-3.19}} & 28.82\textsubscript{~\bl{-1.82}} &
   31.02\textsubscript{\mg{+0.01}}&
    1.98\textsubscript{~\mg{+1.96}}\\
    \rowcolor[gray]{0.92}DS & 4/32 & \textbf{13.21\textsubscript{~\bl{-0.37}}} & \textbf{19.01\textsubscript{~\bl{-0.49}}} &
    \textbf{31.41\textsubscript{~\mg{+0.04}}} &
    \textbf{5.64\textsubscript{~\mg{+8.26}}} & \textbf{25.63\textsubscript{~\bl{-2.99}}} & \textbf{56.14\textsubscript{~\bl{-4.20}}} &
   \textbf{31.12\textsubscript{\mg{+0.24}}}&
    \textbf{3.87\textsubscript{~\mg{+2.48}}} & \textbf{18.22\textsubscript{~\bl{-3.53}}} & \textbf{28.01\textsubscript{~\bl{-2.63}}} &
   \textbf{31.04\textsubscript{\mg{+0.03}}}&
    \textbf{2.37\textsubscript{~\mg{+2.35}}}\\
    \midrule
    
     Q-Diffusion\dag~\cite{li2023qdiffusion} & 8/8 & 13.31 & 20.54 & 31.34 & -2.60 & 28.21  & 60.06 & 31.15 & 2.89 & 21.65 & 30.71 & 31.07 & 1.99\\
    \rowcolor[gray]{0.92}TM & 8/8 & 13.09\textsubscript{~\bl{-0.22}} & 19.91\textsubscript{~\bl{-0.63}} &
    31.34\textsubscript{~\mg{+0.00}}&
    6.69\textsubscript{~\mg{+9.29}} & 25.27\textsubscript{~\bl{-2.94}} & 56.36\textsubscript{~\bl{-3.70}} &
   31.20\textsubscript{\mg{+0.05}}&
    5.29\textsubscript{~\mg{+2.40}} & 18.08\textsubscript{~\bl{-3.57}} & 30.00\textsubscript{~\bl{-0.71}} &
   31.15\textsubscript{\mg{+0.08}}&
    3.58\textsubscript{~\mg{+1.59}}\\
    
    \rowcolor[gray]{0.92}CM & 8/8 & 13.08\textsubscript{~\bl{-0.23}} & 19.93\textsubscript{~\bl{-0.61}} &
    31.32\textsubscript{~\mg{+0.02}}&
    6.62\textsubscript{~\mg{+9.22}} & 25.12\textsubscript{~\bl{-3.09}} & 56.20\textsubscript{~\bl{-3.86}} &
   31.22\textsubscript{\mg{+0.07}}&
    5.76\textsubscript{~\mg{+2.87}} & 18.17\textsubscript{~\bl{-3.66}} & 29.41\textsubscript{~\bl{-1.30}} &
   31.16\textsubscript{\mg{+0.09}}&
    3.73\textsubscript{~\mg{+1.74}}\\
    
    \rowcolor[gray]{0.92}DS & 8/8 & \textbf{13.06\textsubscript{~\bl{-0.25}}} & \textbf{19.88\textsubscript{~\bl{-0.68}}} &
    \textbf{31.30\textsubscript{~\mg{+0.04}}}&
    \textbf{6.70\textsubscript{~\mg{+9.30}}} & \textbf{25.04\textsubscript{~\bl{-3.17}}} & \textbf{56.12\textsubscript{~\bl{-3.94}}} &
   \textbf{31.24\textsubscript{\mg{+0.09}}}&
    \textbf{5.99\textsubscript{~\mg{+3.10}}} & \textbf{18.02\textsubscript{~\bl{-3.81}}} & \textbf{29.38\textsubscript{~\bl{-1.33}}} &
   \textbf{31.19\textsubscript{\mg{+0.12}}}&
    \textbf{3.97\textsubscript{~\mg{+1.88}}}\\
    \midrule
    
     Q-Diffusion\dag~\cite{li2023qdiffusion} & 4/8 & 14.49 & 20.43 & 31.21 & -2.59 & 31.18  & 60.22 & 31.10 & 0.88 & 23.71 & 30.67 & 30.98 & -0.26\\
    \rowcolor[gray]{0.92}TM & 4/8 & 13.36\textsubscript{~\bl{-1.13}} & 20.14\textsubscript{~\bl{-0.29}} &
   31.28\textsubscript{\mg{+0.07}}&
    5.59\textsubscript{~\mg{+8.15}} & 25.68\textsubscript{~\bl{-5.50}} & 56.95\textsubscript{~\bl{-3.27}} &
   \textbf{31.15\textsubscript{\mg{+0.05}}} & 3.12\textsubscript{\mg{+2.24}} & 18.24\textsubscript{~\bl{-5.47}} & 30.38\textsubscript{~\bl{-0.29}} &
   31.02\textsubscript{\mg{+0.04}} &
    1.64\textsubscript{~\mg{+1.90}}\\
    
    \rowcolor[gray]{0.92}CM & 4/8 & 13.31\textsubscript{~\bl{-1.18}} & 20.13\textsubscript{~\bl{-0.31}} &
   31.28\textsubscript{\mg{+0.07}}&
    5.56\textsubscript{~\mg{+8.12}} & 25.89\textsubscript{~\bl{-5.29}} & 56.54\textsubscript{~\bl{-3.68}} &
   31.13\textsubscript{\mg{+0.03}}& 3.28\textsubscript{\mg{+2.40}} & 18.39\textsubscript{~\bl{-5.32}} & 29.48\textsubscript{~\bl{-1.19}} &
   31.05\textsubscript{\mg{+0.07}} &
    1.78\textsubscript{~\mg{+2.04}}\\
    
    \rowcolor[gray]{0.92}DS & 4/8 & \textbf{13.29\textsubscript{~\bl{-1.20}}} & \textbf{20.09\textsubscript{~\bl{-0.35}}} &
   \textbf{31.30\textsubscript{\mg{+0.09}}}&
    \textbf{5.60\textsubscript{~\mg{+8.16}}} & \textbf{25.57\textsubscript{~\bl{-5.61}}} & \textbf{56.42\textsubscript{~\bl{-3.80}}} &
   \textbf{31.15\textsubscript{\mg{+0.05}}}&
    \textbf{3.49\textsubscript{~\mg{+2.61}}} & \textbf{18.21\textsubscript{~\bl{-5.50}}} & \textbf{29.11\textsubscript{~\bl{-1.56}}} &
   \textbf{31.06\textsubscript{\mg{+0.08}}}&
    \textbf{1.82\textsubscript{~\mg{+2.08}}}\\
    \bottomrule
\end{tabular}
}
    \label{tab:sota_text}
\end{table*}
\vspace{1em}
\noindent\textbf{Unconditional generation.} In the experiments conducted on the LDM with DDIM sampler~\cite{songddim}, we maintain the same experimental settings as presented in ~\cite{rombach2022ldm}, including the number of steps, variance schedule, and classifier-free guidance scale  (denoted by \texttt{eta} and \texttt{cfg} in the following, respectively). As shown in Tab.~\ref{tab:sota_ldm}, the FID performance differences relative to the full precision (FP) model are all within $0.7$ for all settings employing DS. Specifically, on the CelebA-HQ $256 \times 256$ dataset, DS exhibits an FID reduction of $\mathbf{6.74}$, an sFID reduction of $\mathbf{6.79}$, and an SQNR increase of $\mathbf{4.09}$ in the W$4$A$8$ setting compared to the current algorithms. It is noticeable that existing methods, whether in $4$-bit or $8$-bit, show significant performance degradation when compared to the FP model on face datasets like CelebA-HQ $256 \times 256$ and FFHQ $256 \times 256$, whereas our TM/CM/DS shows almost no performance degradation. Importantly, DS also achieves significant performance improvement on the LSUN-Bedrooms $256 \times 256$ in the W$4$A$8$ setting, with FID and sFID reductions of $\mathbf{2.33}$ and $\mathbf{7.67}$ compared to PTQD~\cite{he2023ptqd}, respectively. Regarding LDM-8 on LSUN-Churches $256 \times 256$, we attribute the moderate improvement, compared to other datasets. We believe that the use of the LDM-8 model with a downsampling factor of $8$ may be more quantization-friendly. Existing methods have already achieved satisfactory results on this dataset. Nonetheless, our method still approaches the performance of the FP model more closely than existing methods.  

Besides the experiments on LDM, we have also conducted experiments with DDPM on CIFAR-10 $32\times 32$. As shown in Tab.~\ref{tab:sota_ddim}, our methods still achieve comprehensive improvements in terms of IS, FID, and SQNR compared to the existing advanced methods. However, due to the lower resolution and simplicity of the images in this dataset, existing methods show minimal performance degradation, so the results we obtain may not be as pronounced.

\vspace{1em}
\noindent\textbf{Class-conditional generation.}
On the ImageNet $256 \times 256$ dataset, we employed a denoising process with a $20$-step DDIM sampler~\cite{songddim}, setting \texttt{eta} and \texttt{cfg} to $0.0$ and $3.0$, respectively. In Tab.~\ref{tab:sota_class}, compared to PTQD, our method achieves an FID reduction of $\mathbf{2.09}$ on W$8$A$8$, and a $\mathbf{5.35}$ sFID decrease on W$4$A$8$. Under all the conditions with DS, we observe an improvement of over $\mathbf{9}$ in IS. Particularly noteworthy is that, across various quantization settings, our method consistently achieved lower FID and higher SQNR compared to the FP model.

\begin{figure*}[!ht]
\centering
\setlength{\abovecaptionskip}{0.2cm}
\renewcommand{\arraystretch}{0.5}
\begin{tabularx}{\linewidth}{X@{\hspace{1.5pt}}X@{\hspace{1.5pt}}X}
\toprule
\footnotesize{\shortstack{\textit{``A chocolate cake on a table.''}}} & \footnotesize{\shortstack{\textit{``A man in the snow on a snow board.''}}} & \footnotesize{\shortstack{\textit{``An image of a bus picking up people on its route.''}}}\\
\midrule
\includegraphics[width=0.323\textwidth]{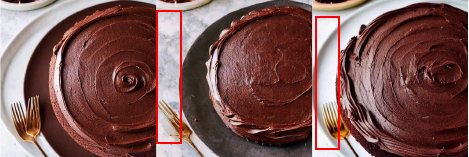} & \includegraphics[width=0.323\textwidth]{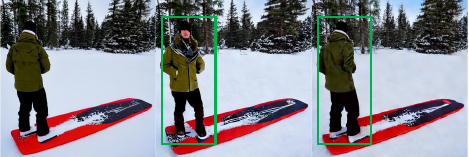}&
\includegraphics[width=0.323\textwidth]{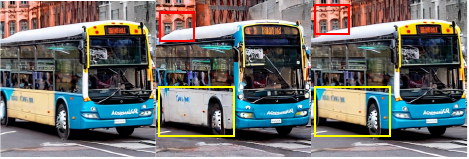}\\
\midrule
\footnotesize{\shortstack{\textit{``Elephant displayed in window of city building}\\\textit{ near roadway.
''}}} & \footnotesize{\shortstack{\textit{``A plate if filled to the rim with sweet pastries
}\\\textit{ and desserts.
''}}} & \footnotesize{\shortstack{\textit{``A cat laying next to a pair of shoes on a hard}\\\textit{ wood floor.
''}}}\\
\midrule
\includegraphics[width=0.323\textwidth]{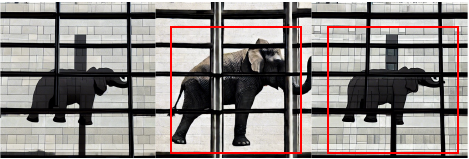} & \includegraphics[width=0.323\textwidth]{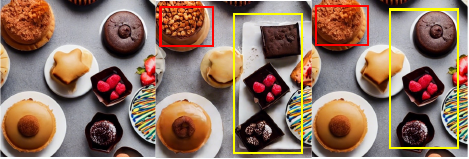} &\includegraphics[width=0.323\textwidth]{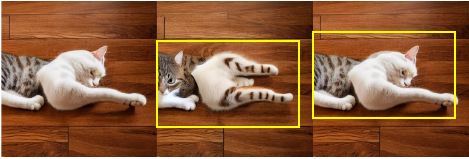}\\
\bottomrule
\end{tabularx}
\caption{\label{fig:qualitative-comp} The images below the corresponding MS-COCO captions are generated from FP (Left), Q-Diffusion (Middle), and our framework (Right) under W$4$A$8$ with Stable Diffusion-v1-4. Different key details are highlighted using rectangles.
}
\end{figure*}
\begin{figure*}[!ht]
    \centering
     \setlength{\abovecaptionskip}{0.2cm}
    \includegraphics[width=0.9\textwidth]{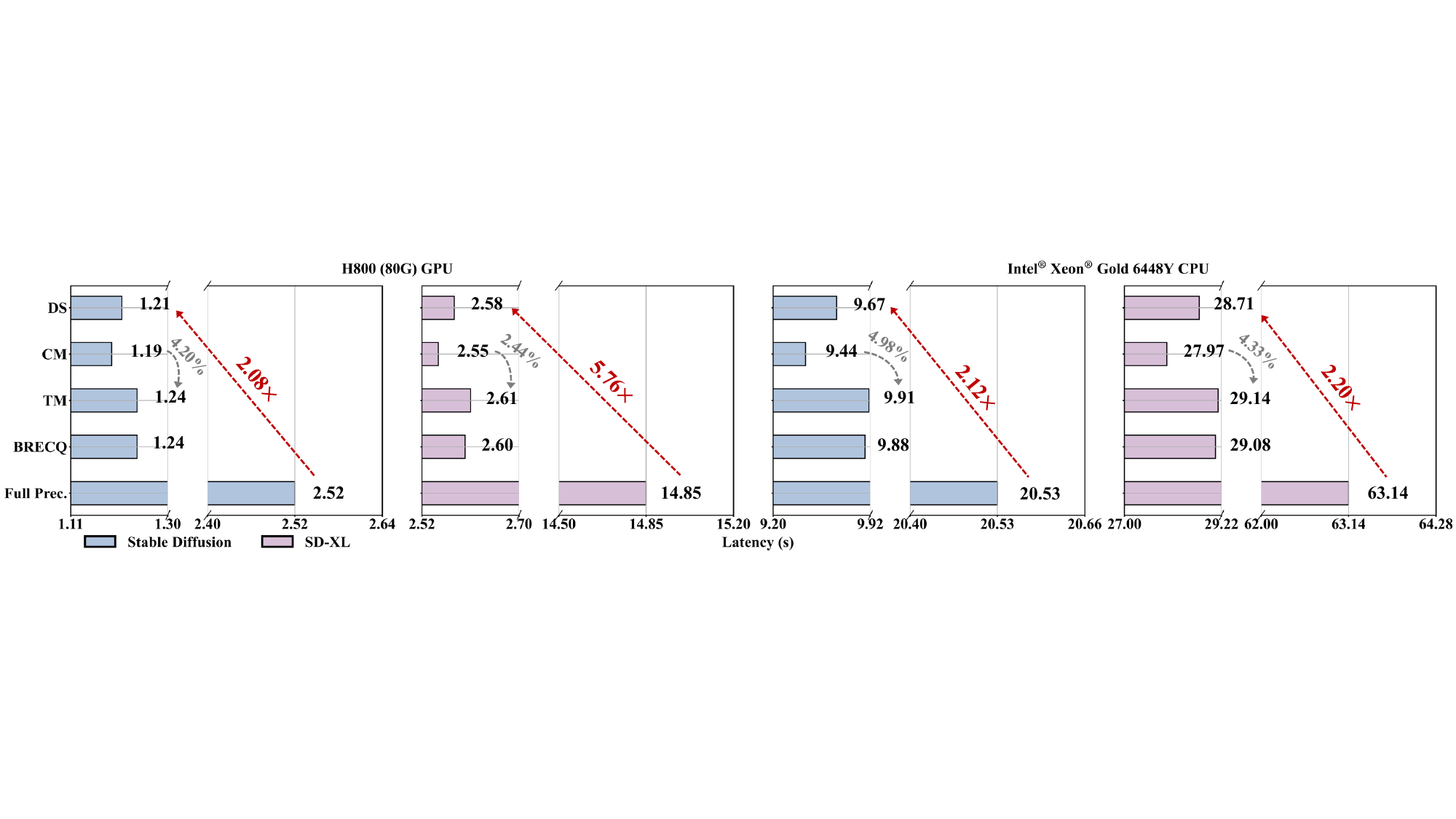}
     \caption{Latency for Stable Diffusion and SD-XL. BRECQ~\cite{li2021brecq} is selected as our baseline. We set the batch size to $1$ here. Full Prec. is implemented in W$32$A$32$ OpenVino and Pytorch.}
    \label{fig:latency}
\end{figure*}
\begin{figure}[!ht]
    \centering
     \setlength{\abovecaptionskip}{0.2cm}
    \includegraphics[width=0.4\textwidth]{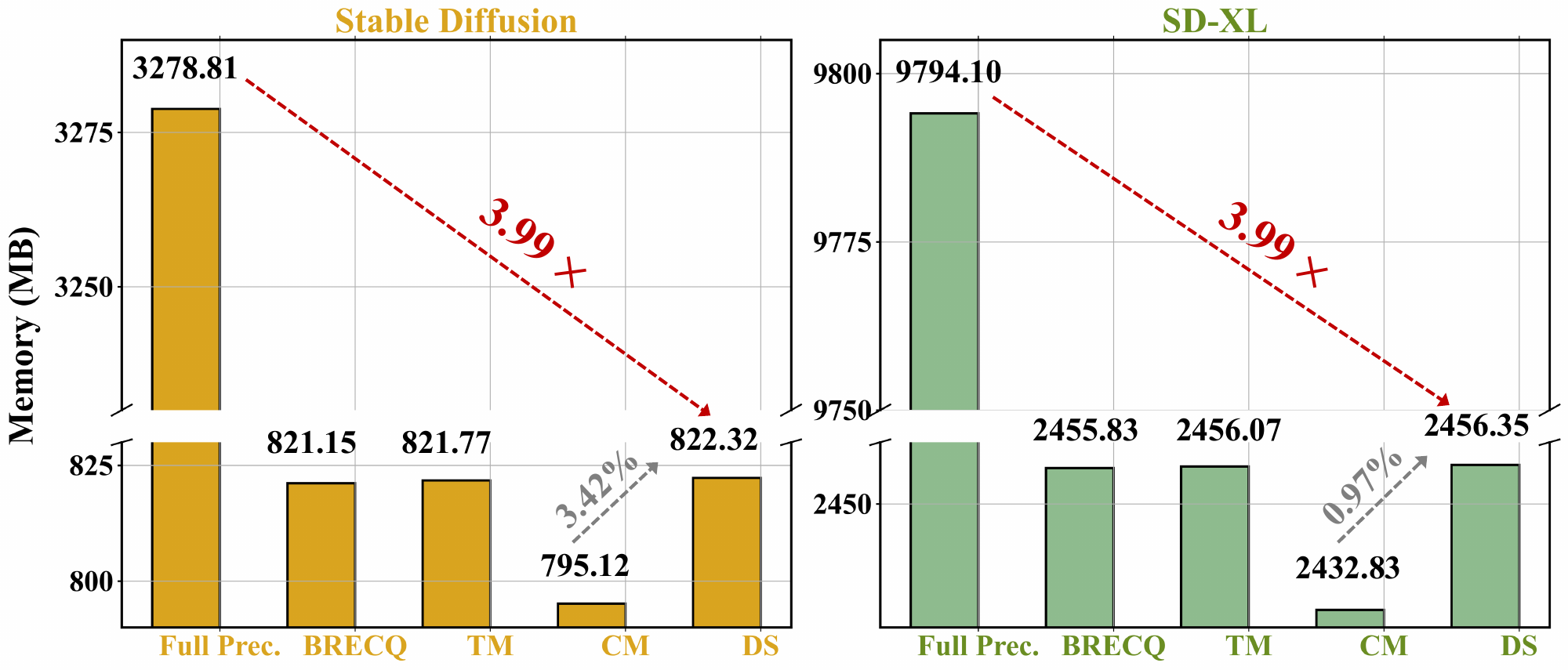}
     \caption{Memory costs for the UNet in Stable Diffusion and SD-XL.}
    \label{fig:memory}
\end{figure}

\vspace{1em}
\noindent\textbf{Text-guided generation.} In this experiment, we generate high-resolution images of $512 \times 512$ pixels using Stable Diffusion-v1-4 with $50$ steps PLMS sampler~\cite{liu2022pseudo}, and SD-XL-turbo with only a single step Euler sampler~\cite{karras2022elucidatingdesignspacediffusionbased}. Moreover, higher resolution images of $1024 \times 1024$ are also generated from SD-XL-base-1.0 with a $50$-step Euler sampler. \texttt{cfg} is fixed to the default $7.5$ for Stable Diffusion and SD-XL as the trade-off between generation quality and diversity. However, we do not use classifier-free guidance~\cite{ho2022classifierfree} to save memory, the same as the original paper~\cite{sauer2023adversarial}. In Tab.~\ref{tab:sota_text}, compared to Q-Diffusion, our approach achieves an FID reduction of $\mathbf{1.20}$, $\mathbf{5.61}$, and $\mathbf{5.50}$ under the W$4$A$8$ configuration for Stable Diffusion, SD-XL, and SD-XL-turbo, respectively. Notably, our FID on W$8$A$8$ and sFID on W$4$A$32$ are even lower than those of the full precision model. The table also shows our significantly lower quantization error than Q-Diffusion. For example, DS obtains an $\mathbf{8.16}$ SQNR upswing for W$4$A$8$ Stable Diffusion. Benefiting from this, the visual effect of our generated images is closer to the FP model with better quality than Q-Diffusion (examples in Fig.~\ref{fig:qualitative-comp} and Sec.~\ref{app:vis}). All of these clues strongly reveal the value of maintaining temporal features. \yushi{More experiments for few-step FLUX.1-Schnell~\cite{flux2024} can also be found in Sec.~\ref{app:flux}.}

\subsection{Efficiency Discussion}\label{sec:deploy}
In this section, we study the inference efficiency of our quantized diffusion models. We discuss all components in our framework in the same order as Sec.~\ref{sec:quantization_framework}. 

Specifically, we focus on popular T2I models, \emph{e.g.}, Stable Diffusion-v1-4~\cite{rombach2022ldm}, SD-XL-base-1.0~\cite{podell2023sdxlimprovinglatentdiffusion} with W$8$A$8$ quantization, and deploy their quantized counterparts on Intel$^\circledR$ Xeon$^\circledR$ Gold $6448Y$ Processor with the OpenVino~\cite{openvino} framework. To demonstrate the acceleration on GPU, we also evaluate the latency on NVIDIA $H800$ GPU with quantized convolutions and multiplications, harnessing CUTLASS~\cite{kerr2017cutlass} with our customized kernels. An overview can be seen from Fig.~\ref{fig:latency} and Fig.~\ref{fig:memory}. \yushi{In Sec.~\ref{app:edge}, we additionally deploy Stable Diffusion on the edge and mobile devices, \emph{e.g.}, NVIDIA Jetson Orin Nano and iPhone $15$ Pro Max.}

\subsubsection{Efficiency of TM}\label{sec:efficiency-tib}
First, compared with BRECQ, our TIB-based Maintenance brings less than $\mathbf{0.08\%}$ extra memory consumption, since we employ per-tensor quantization with our FSC for activations. Besides, our strategy only induces an approximated $\mathbf{0.3\%}$ speed overhead for Stable Diffusion and SD-XL on CPUs, and the latency comparison results on the GPU also show nearly no difference.

\subsubsection{Efficiency of CM}\label{sec:efficiency-cache}
Second, our CM removes the necessity to compute temporal features online, which slightly accelerates inference. Specifically, $\mathbf{4.33\%}$ speed enhancement for SD-XL on CPU is obtained compared with TM. Note that CM also requires less storage here, \emph{e.g.}, $\mathbf{26.65}$MB memory reduction for Stable Diffusion than TM. Even for some models with a large timesteps count, like $500$ steps for LDM on LSUN-Bedroom $256\times256$, an additional $\mathbf{10.4}$MB memory requirement without the TIB brings an extra cost amounting to roughly $\mathbf{3.69\%}$ of the quantized UNet size~($281.8$MB). Moreover, our framework can seamlessly integrate step-reduction techniques like advanced samplers or step distillation to reduce inference costs. We leave this for our future study.

\subsubsection{Efficiency of DS}\label{sec:efficiency-select}
Last, equipped with the DS, our complete framework achieves a pronounced efficient inference with significantly lower memory consumption compared with FP. For example, the quantized SD-XL is $\mathbf{5.76\times}$ faster than the full precision with $\mathbf{3.99\times}$ memory savings. It is obvious that the time consumption of all different selection results is between TM and CM, so less than a $\mathbf{5\%}$ gap between these results is achieved as shown in Fig.~\ref{fig:latency}. Besides, compared to TM in Fig.~\ref{fig:memory}, DS has almost no additional overhead. Thus, we verify the high inference efficiency of our framework.

\subsection{Ablation Studies}\label{sec:ablation}
\begin{table}[!ht]\setlength{\tabcolsep}{5pt}
  \centering
  \caption{The effect of different modules proposed in the paper on LSUN-Bedrooms $256 \times 256$. The subscript numbers represent the improvements compared with our baseline.  The last $3$ rows denote TM, CM, and DS from upper to lower, respectively.} 
  \resizebox{0.95\linewidth}{!}{
      
\begin{tabular}{cccclll}
    \toprule
    \multicolumn{2}{c}{\multirow{3}{*}{\makecell{TIB\\-based\\Maintenance}}} & \multicolumn{1}{c}{\multirow{4}{*}{\makecell{Cache\\-based\\Maintenance}}} &  \multicolumn{1}{c}{\multirow{4}{*}{\makecell{Disturbance\\-aware\\ Selection}}} &  \multicolumn{3}{c}{\multirow{3}{*}{\makecell{\textbf{LSUN-Bedrooms}\\\textbf{$256\times256$}}}} \\ 
    \\ 
    & & & & & &
    \\\cmidrule(r){5-7}
    \cmidrule(r){1-2}
    TIAR & FSC & & & \multicolumn{1}{c}{\multirow{1}{*}{FID$\downarrow$}} & \multicolumn{1}{c}{\multirow{1}{*}{sFID$\downarrow$}} &
    \multicolumn{1}{c}{\multirow{1}{*}{SQNR$\uparrow$}}\\
    \midrule
     \multicolumn{4}{c}{\multirow{1}{*}{Full Prec.}} & 2.98& 7.09& - \\\cmidrule(r){1-7}
     \multicolumn{4}{c}{\multirow{1}{*}{BRECQ~\cite{li2021brecq}~(Baseline)}}& 9.36& 22.73& -1.61 \\
    \cdashline{1-7}
   \ding{52} & & & & 4.84\textsubscript{~\bl{-4.52}}& 9.29\textsubscript{~\bl{-13.44}}&  6.13\textsubscript{~\mg{+7.74}} \\
   & \ding{52}  & & & 6.07\textsubscript{~\bl{-3.29}}& 11.31\textsubscript{~\bl{-11.42}}&  5.21\textsubscript{~\mg{+6.82}} \\
    \rowcolor[gray]{0.92} \ding{52} & \ding{52} & & & 3.68\textsubscript{~\bl{-5.68}}& 7.65\textsubscript{~\bl{-15.08}}&  8.02\textsubscript{~\mg{+9.63}}  \\
      \rowcolor[gray]{0.92}& &\ding{52} & & 3.91\textsubscript{~\bl{-5.45}}& 8.61\textsubscript{~\bl{-14.12}}&  7.30\textsubscript{~\mg{+8.91}}  \\
    \rowcolor[gray]{0.92}\ding{52} & \ding{52} & \ding{52} & \ding{52} & \textbf{3.61\textsubscript{~\bl{-5.75}}}& \textbf{7.49\textsubscript{~\bl{-15.24}}}&  \textbf{8.51\textsubscript{~\mg{+10.12}}}\\
    \bottomrule
\end{tabular}
}
    \label{tab:ablation}
\end{table}
To evaluate the effectiveness of different modules in our proposed method, we perform a thorough ablation study on the LSUN-Bedrooms $256 \times 256$ dataset with W$4$A$8$ quantization, utilizing the LDM-4 model with a DDIM sampler, unless otherwise stated. The basic ablation is shown in Tab.~\ref{tab:ablation}.

The content is organized as follows. We first explore the effect of TIB-based Maintenance in Sec.~\ref{sec:effect-tib}. Then, we present detailed calibration ablation for TIB-based Maintenance and Cache-based Maintenance in Sec.~\ref{sec:cali-tib} and Sec.~\ref{sec:cali-cache}, respectively. Moreover, we evaluate the loss function of Disturbance-aware Selection and conduct detailed analyses of its selection proportion in Sec.~\ref{sec:metric-distur} and Sec.~\ref{sec:ana-distur}. Finally, the performance with advanced samplers is exhibited to further showcase our universality in Sec.~\ref{sec:sampler}.

\subsubsection{Effect of TM}\label{sec:effect-tib}
As shown in Tab.~\ref{tab:ablation}, compared to the baseline method, our TIAR reduces FID and sFID by $\mathbf{4.52}$ and $\mathbf{13.44}$, respectively. Additionally, our FSC also achieves promising results. Equipping both approaches can further mitigate performance degradation. 

\begin{figure}[!ht]
    \centering
     \setlength{\abovecaptionskip}{0.2cm}
    \includegraphics[width=0.45\textwidth]{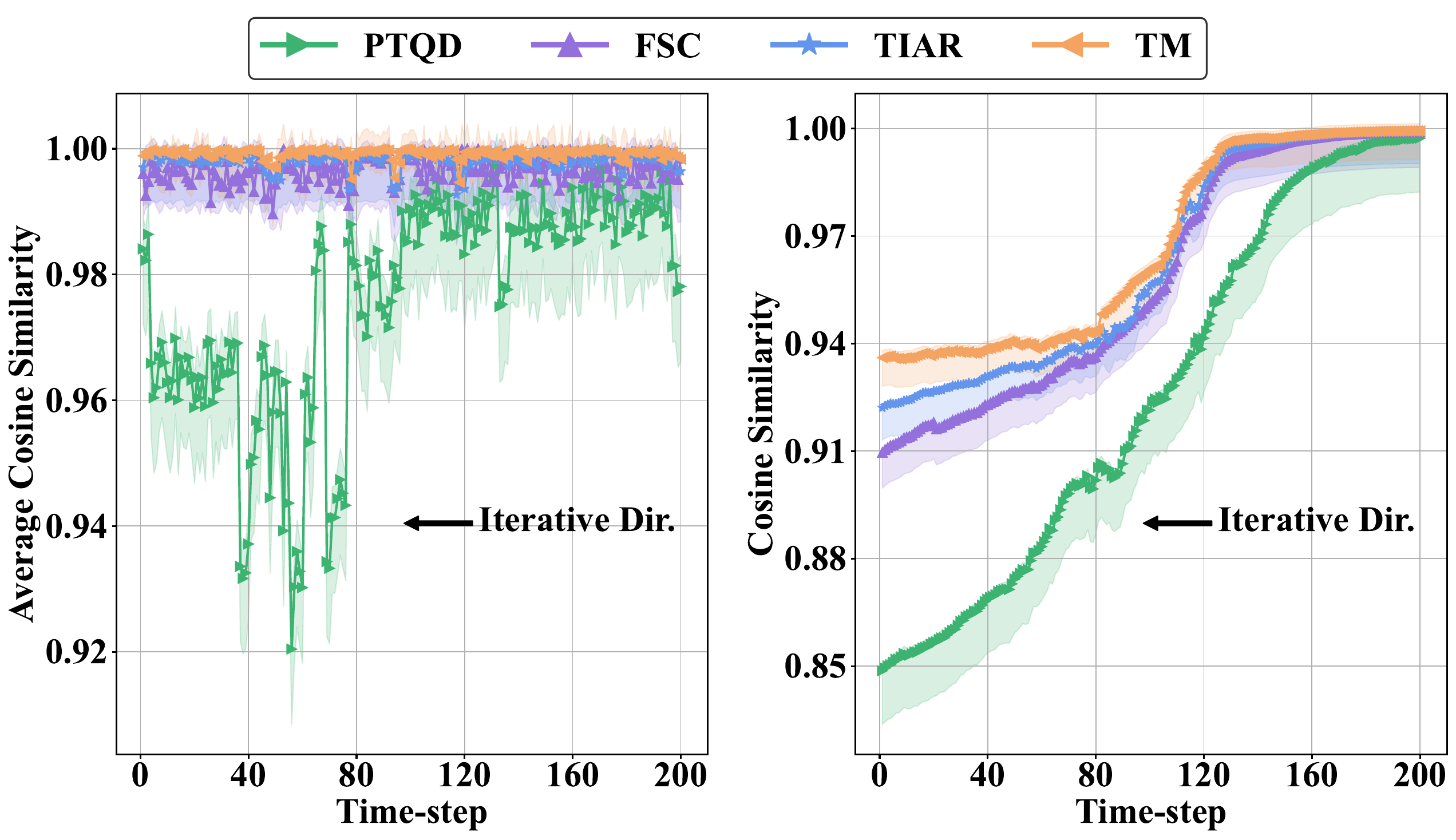}
     \caption{(Left)~Temporal feature errors employing Eq.~(\ref{eq:temporal_feature_error}) across different methods. (Right) Cosine similarity of the Residual Bottleneck's outputs across different PTQ Methods~($i=8$).}
    \label{fig:ablation}
\end{figure}
For a more detailed analysis, Fig.~\ref{fig:ablation}~(Left) reveals that our methods induce significantly less temporal feature disturbance compared to PTQD. This underscores the efficacy of our motivation and contribution to maintaining temporal features. Further insights are gained by examining the cosine similarity between outputs from the \textit{$i$-th} Residual Bottleneck Blocks pre- and post-quantization. Comparing our methods with PTQD, Fig.~\ref{fig:ablation}~(Right) illustrates that our approaches result in significantly lower output errors in the Residual Bottleneck Block. It is also essential to note that these points involve the accumulated errors from multiple denoising iterations in diffusion models. Therefore, our methods are able to significantly eliminate the cumulative error and quantization error by mitigating temporal feature disturbance within the corresponding block for performance enhancements.

\subsubsection{Effect of Calibration Methods for TM}\label{sec:cali-tib}
Given the variety of available methods to determine the optimal step size for activations during calibration, we explore several approaches and evaluate both their performance and the GPU time consumed. As detailed in Tab.~\ref{tab:cali_method}, the performance differences between Min-Max and the best-performing ones are minimal, \emph{e.g.}, a mere $\mathbf{0.06}$ FID gap under W$4$A$8$. Therefore, we opt for the simplest and most efficient Min-Max~\cite{nagel2021white} as our specific calibration strategy, striking a balance between calibration time and effectiveness. Notably, we have found that PTQD and Q-Diffusion cost $4.68$ and $5.29$ GPU hours in their PTQ methods under W$4$A$8$ quantization, respectively. However, our TIB-based maintenance only spends $\mathbf{2.32}$ GPU hours~($\mathbf{2\times}$ speedup) due to our efficient calibration and reconstruction without extra time consumption compared with previous approaches.
\begin{table}[!ht]\setlength{\tabcolsep}{5pt}
  \centering
  \caption{Different range estimation methods for calibration of TM. We report the GPU hours consumed during calibration in the table. Weight reconstruction with our TIAR~($\mathbf{2.20}$ GPU hours) is used before calibration with FSC. The indices represent the improvements between the best-performing methods and Min-Max~\cite{nagel2021white}.} 
  \resizebox{0.95\linewidth}{!}{
  \begin{tabular}{lcllll}
    \toprule
    \textbf{Methods} & \textbf{\#Bits (W/A)} & FID$\downarrow$ & sFID$\downarrow$ & SQNR$\uparrow$& Time \\
    \midrule
    LSQ~\cite{esser2020lsq} & 8/8 & 3.17 & 7.18 & \textbf{9.23\textsubscript{~\mg{+0.11}}} & 2.48 \\
    KL-divergence~\cite{kl} & 8/8 & 3.27 & 7.32 & 9.04 & 19.67 \\
    Percentile~\cite{wu2020integer} & 8/8 & 3.34 & 7.41 & 9.09 & 12.00 \\
    MSE~\cite{choukroun2019low} & 8/8 & \textbf{3.12\textsubscript{~\bl{-0.02}}} & \textbf{7.12\textsubscript{~\bl{-0.14}}} & 9.09 & 8.89 \\
    \rowcolor[gray]{0.92}Min-Max~\cite{nagel2021white} & 8/8 & 3.14 & 7.26 & 9.12 & \textbf{0.12\textsubscript{~\bl{-2.36}}} \\
    \midrule
    LSQ~\cite{esser2020lsq} & 4/8 & 3.69 & 7.48 & \textbf{8.15\textsubscript{~\mg{+0.13}}} & 2.57 \\
    KL-divergence~\cite{kl} & 4/8 & 3.94 & \textbf{7.42\textsubscript{~\bl{-0.23}}} & 8.01 & 19.65 \\
    Percentile~\cite{wu2020integer} & 4/8 & 3.74 & 8.02 & 7.98 & 12.04 \\
    MSE~\cite{choukroun2019low} & 4/8 & \textbf{3.62\textsubscript{~\bl{-0.06}}} & 7.48 & 7.99 & 8.89 \\
    \rowcolor[gray]{0.92}Min-Max~\cite{nagel2021white} & 4/8 & 3.68 & 7.65 & 8.02 & \textbf{0.12\textsubscript{~\bl{-2.45}}} \\
    \bottomrule
\end{tabular}
}
    \label{tab:cali_method}
\end{table}

\subsubsection{Effect of Calibration Methods for CM}\label{sec:cali-cache}
We also investigate different calibration methods for our Cache-based Maintenance. From Tab.~\ref{tab:cali_method_cache}, we observe that LSQ outperforms the others, achieving a $\mathbf{13.3\%}$ increase in SQNR compared to MSE under W$8$A$8$ settings. Hence, we choose LSQ for calibration. Note that each pre-computed temporal feature can be processed in parallel, enabling the algorithm to achieve rapid runtimes of $\mathbf{<0.3}$ hours on a single $H800$ $80$G GPU.
\begin{table}[!ht]\setlength{\tabcolsep}{6pt}
  \centering
  \caption{Different range estimation methods for calibration of Cache-based Maintenance. The subscript numbers represent the improvements between the first and second.} 
  \resizebox{0.9\linewidth}{!}{
  \begin{tabular}{lclll}
    \toprule
    \textbf{Methods} & \textbf{\#Bits (W/A)} & FID$\downarrow$ & sFID$\downarrow$ & SQNR$\uparrow$ \\
    \midrule
    \rowcolor[gray]{0.92}LSQ~\cite{esser2020lsq} & 8/8 & \textbf{3.11\textsubscript{~\bl{-0.57}}} & \textbf{7.12\textsubscript{~\bl{-0.26}}} & \textbf{9.43\textsubscript{~\mg{+1.11}}} \\
    KL-divergence~\cite{kl} & 8/8 & 3.76 & 7.61 & 7.58  \\
    Percentile~\cite{wu2020integer} & 8/8 & 3.68 & 7.63 & 7.89 \\
    MSE~\cite{choukroun2019low} & 8/8 & 3.62 & 7.54 &  8.32 \\
    Min-Max~\cite{nagel2021white} & 8/8 & 4.41 & 7.38 & 6.91 \\
    \midrule
    \rowcolor[gray]{0.92}LSQ~\cite{esser2020lsq} & 4/8 & \textbf{3.91\textsubscript{~\bl{-0.35}}} & \textbf{8.61\textsubscript{~\bl{-0.41}}} & \textbf{7.30\textsubscript{~\mg{+0.18}}} \\
    KL-divergence~\cite{kl} & 4/8 & 4.35 & 9.02 & 6.89 \\
    Percentile~\cite{wu2020integer} & 4/8 & 4.68 & 9.42 & 7.01 \\
    MSE~\cite{choukroun2019low} & 4/8 & 4.26 & 9.08 & 7.12 \\
    Min-Max~\cite{nagel2021white} & 4/8 & 5.41 & 10.38 & 5.02 \\
    \bottomrule
\end{tabular}
}
    \label{tab:cali_method_cache}
\end{table}

\subsubsection{Effect of Loss Functions for DS}\label{sec:metric-distur}
\jing{Since the loss function $\mathcal{L}_{t,i}^{temporal}$ in Eq.~(\ref{eq:select}) measures the error of the temporal feature, we assess various loss functions to enhance performance.}
\jing{As shown in Tab.~\ref{tab:metric-select}, different loss functions produce comparable results, with less than $\mathbf{1\%}$ FID and sFID discrepancies among them.}
Moreover, all results in the table outperform either TIB-based or Cache-based Maintenance, 
\jing{demonstrating the robustness and effectiveness of the selection strategy.}
\begin{table}[!ht]\setlength{\tabcolsep}{10pt}
  \centering
  \caption{Effect of different loss functions in Eq.~(\ref{eq:select}). We employ MSE~\cite{pmlr-v162-zhou22c}, aligning with Eq.~(\ref{eq:cache}) in our framework.} 
  \resizebox{0.9\linewidth}{!}{
  \begin{tabular}{lclll}
    \toprule
    \textbf{Loss Func.} & \textbf{\#Bits (W/A)} & FID$\downarrow$ & sFID$\downarrow$ & SQNR$\uparrow$ \\
    \midrule
    \rowcolor[gray]{0.92}MSE~\cite{pmlr-v162-zhou22c} & 8/8 & \textbf{3.08\textsubscript{~\bl{-0.01}}} & 7.10 & 9.70 \\
    KL~\cite{kim2021comparing} & 8/8 & 3.09 & 7.11 & \textbf{9.85\textsubscript{~\mg{+0.15}}}  \\
    CE~\cite{mao2023cross} & 8/8 & 3.09 & \textbf{7.07\textsubscript{~\bl{-0.03}}} & 9.59 \\
    \midrule
    \rowcolor[gray]{0.92}MSE~\cite{pmlr-v162-zhou22c} & 4/8 & \textbf{3.61\textsubscript{~\bl{-0.02}}} & \textbf{7.49\textsubscript{~\bl{-0.01}}} & 8.51 \\
    KL~\cite{kim2021comparing} & 4/8 & 3.63 & 7.56 & 8.49  \\
    CE~\cite{mao2023cross} & 4/8 & 3.64 & 7.50 & \textbf{8.90\textsubscript{~\mg{+0.39}}} \\
    \bottomrule
\end{tabular}
}
    \label{tab:metric-select}
\end{table}

\subsubsection{Effect of Selection Proportion}\label{sec:ana-distur}
\begin{figure}[!ht]
    \centering
    \setlength{\abovecaptionskip}{0.2cm}
     \includegraphics[width=0.48\textwidth]{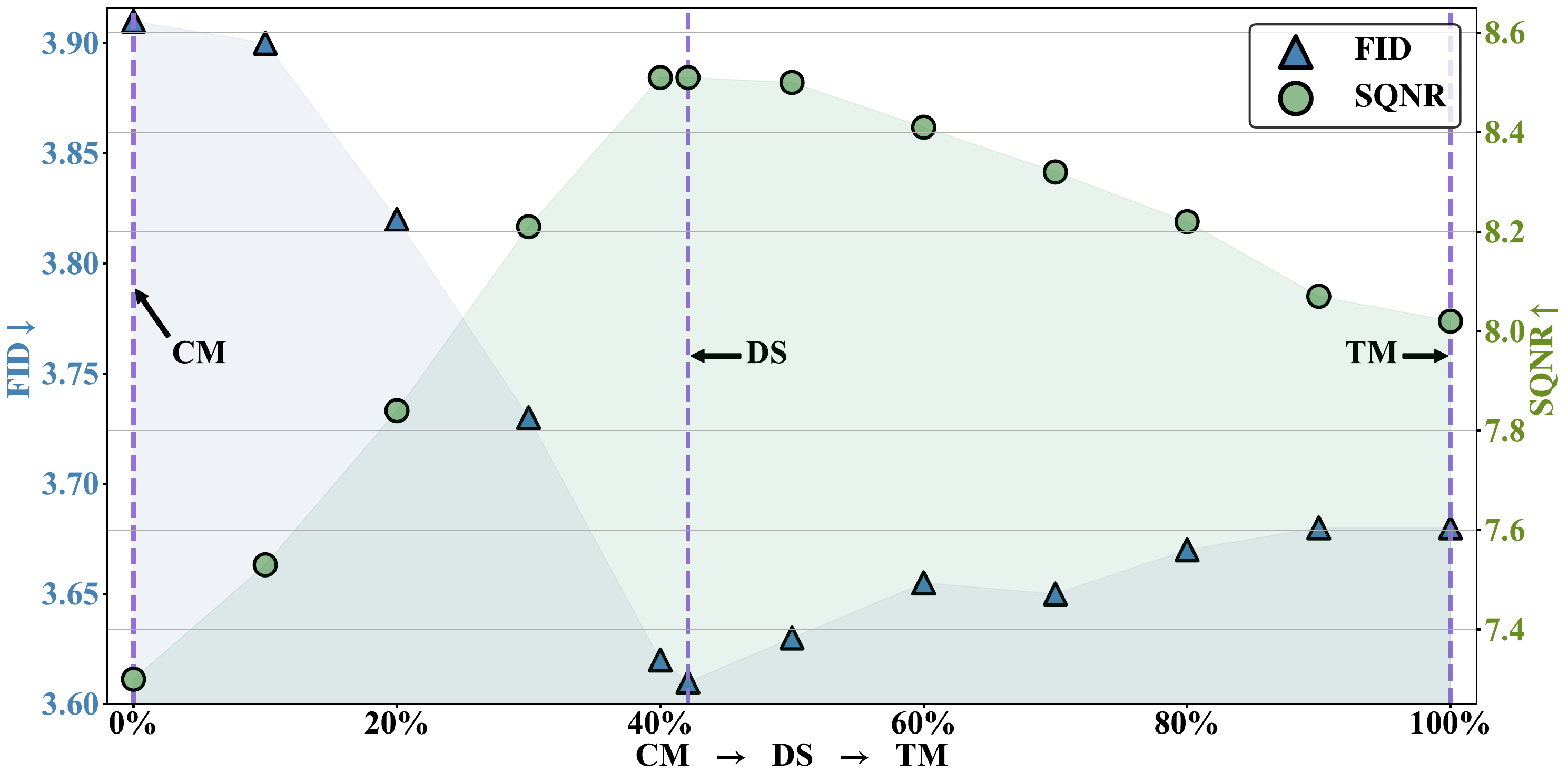}
     \caption{Performance from CM $\rightarrow$ DS $\rightarrow$ TM. The x-axis represents the proportion of temporal features applied using TM.}
    \label{fig:proportion}
\end{figure}

Beginning with Cache-based Maintenance, we select a proportion of the temporal feature with the lowest $\tau_{t,i}$ for TIB-based Maintenance. Then we increase the proportion progressively until the selection pattern is equivalent to TIB-based Maintenance. As shown in Fig.~\ref{fig:proportion}, the lowest FID and highest SQNR were achieved by DS compared with other points. Furthermore, selecting maintenance strategies that minimize temporal feature errors can help enhance performance. This investigation also validates that temporal feature maintenance is crucial.

\subsubsection{Generation with Advanced Samplers}\label{sec:sampler}
Apart from using the DDIM sampler~\cite{songddim}, we also utilize a variant of DDPM~\cite{ho2020denoising} called PLMS~\cite{liu2022pseudo} on the CelebA-HQ $256\times 256$ dataset~\cite{karras2018progressive}. This better demonstrates the superiority of our framework compared to previous works. From Tab.~\ref{tab:sota_plms}, the introduced DS substantially reduces FID and sFID, surpassing PTQD by margins of $\mathbf{12.61}$ and $\mathbf{7.08}$, respectively. Furthermore, we present experiments performed with the DPM++ solver~\cite{lu2023dpmsolver}. As shown in Tab.~\ref{tab:sota_dpm}, our framework consistently outperforms existing methods.
\begin{table}[!ht]\setlength{\tabcolsep}{12pt}
  \centering
  \caption{Quantization results for unconditional image generation with PLMS on CelebA-HQ $256 \times 256$.} 
  \resizebox{0.85\linewidth}{!}{
  \begin{tabular}{lcll}
    \toprule
    \multicolumn{1}{c}{\multirow{2}{*}{\textbf{Methods}}} & \multicolumn{1}{c}{\multirow{2}{*}{\textbf{\#Bits (W/A)}}} & \multicolumn{2}{c}{\textbf{CelebA-HQ $256 \times 256$}} \\ \cmidrule(r){3-4}
    & & FID$\downarrow$ & sFID$\downarrow$ \\
    \midrule
    Full Prec. & 32/32 & 8.92 & 10.42 \\
    \midrule
    Q-Diffusion~\cite{li2023qdiffusion} & 4/8 & 24.31 & 22.11 \\
    PTQD~\cite{he2023ptqd} & 4/8 & 21.08 & 17.38 \\
    \rowcolor[gray]{0.9}DS & 4/8 & \textbf{8.47\textsubscript{~\bl{-12.61}}} & \textbf{10.30\textsubscript{~\bl{-7.08}}} \\
    \bottomrule
\end{tabular}
}
    \label{tab:sota_plms}
\end{table}
\begin{table}[!ht]\setlength{\tabcolsep}{12pt}
  \centering
  \caption{Quantization results for unconditional image generation with DPM++ on LSUN-Churches $256 \times 256$.} 
  \resizebox{0.85\linewidth}{!}{
  \begin{tabular}{lcll}
    \toprule
    \multicolumn{1}{c}{\multirow{2}{*}{\textbf{Methods}}} & \multicolumn{1}{c}{\multirow{2}{*}{\textbf{\#Bits (W/A)}}} & \multicolumn{2}{c}{\textbf{LSUN-Churches $256 \times 256$}} \\ \cmidrule(r){3-4}
    & & FID$\downarrow$ & sFID$\downarrow$ \\
    \midrule
    Full Prec. & 32/32 & 4.12 & 10.55 \\
    \midrule
    Q-Diffusion~\cite{li2023qdiffusion} & 4/8 & 7.80 & 23.24 \\
    PTQD~\cite{he2023ptqd} & 4/8 & 7.45 & 22.74 \\
    \rowcolor[gray]{0.9}DS & 4/8 & \textbf{4.78\textsubscript{~\bl{-2.67}}} & \textbf{11.97\textsubscript{~\bl{-10.77}}} \\
    \bottomrule
\end{tabular}
}
    \label{tab:sota_dpm}
\end{table}

\section{Conclusions \& Limitations} 
In this paper, we have explored the application of quantization to accelerate diffusion models. We have identified a significant and previously unrecognized issue—temporal feature disturbance—in the quantization of diffusion models. Through a detailed analysis, we have pinpointed the root causes of this disturbance and introduced a novel quantization framework to address them. By integrating two maintenance strategies (\emph{i.e.}, TIB-based and Cache-based Maintenance) with our Disturbance-aware Selection, we are able to significantly reduce temporal feature errors. In $4$-bit quantization across various datasets and diffusion models, our framework has exhibited minimal performance degradation compared to full-precision models. Moreover, the enhanced inference speed on both GPU and CPU has validated the considerable hardware efficiency of our quantized diffusion model. 

In terms of limitations, we have discovered that other semantic features introduced in conditional diffusion models possess physical significance and impact the generation effect. Nevertheless, these features are often overlooked in current methodologies. We plan to address this in future research. Furthermore, while temporal feature maintenance is effective in both PTQ and QAT scenarios, our current study focuses primarily on the PTQ setting. Future efforts will extend to the QAT setting to pursue lower-bit quantization and further performance improvements.

\bibliographystyle{abbrv}
{
	\bibliography{reference}

\begin{thebibliography}{100}

\bibitem{bao2022estimating}
F.~Bao, C.~Li, J.~Sun, J.~Zhu, and B.~Zhang.
\newblock Estimating the optimal covariance with imperfect mean in diffusion probabilistic models.
\newblock In {\em ICML}, volume 162, pages 1555--1584, 2022.

\bibitem{bao2022analytic}
F.~Bao, C.~Li, J.~Zhu, and B.~Zhang.
\newblock Analytic-dpm: an analytic estimate of the optimal reverse variance in diffusion probabilistic models.
\newblock In {\em ICLR}, 2022.

\bibitem{bhalgat2020lsq}
Y.~Bhalgat, J.~Lee, M.~Nagel, T.~Blankevoort, and N.~Kwak.
\newblock Lsq+: Improving low-bit quantization through learnable offsets and better initialization, 2020.

\bibitem{brown2020language}
T.~B. Brown, B.~Mann, N.~Ryder, M.~Subbiah, J.~Kaplan, P.~Dhariwal, A.~Neelakantan, P.~Shyam, G.~Sastry, A.~Askell, S.~Agarwal, A.~Herbert-Voss, G.~Krueger, T.~Henighan, R.~Child, A.~Ramesh, D.~M. Ziegler, J.~Wu, C.~Winter, C.~Hesse, M.~Chen, E.~Sigler, M.~Litwin, S.~Gray, B.~Chess, J.~Clark, C.~Berner, S.~McCandlish, A.~Radford, I.~Sutskever, and D.~Amodei.
\newblock Language models are few-shot learners, 2020.

\bibitem{chen2024qditaccurateposttrainingquantization}
L.~Chen, Y.~Meng, C.~Tang, X.~Ma, J.~Jiang, X.~Wang, Z.~Wang, and W.~Zhu.
\newblock Q-dit: Accurate post-training quantization for diffusion transformers, 2024.

\bibitem{chen2023speed}
Y.-H. Chen, R.~Sarokin, J.~Lee, J.~Tang, C.-L. Chang, A.~Kulik, and M.~Grundmann.
\newblock Speed is all you need: On-device acceleration of large diffusion models via gpu-aware optimizations.
\newblock In {\em CVPR}, pages 4651--4655, 2023.

\bibitem{choukroun2019low}
Y.~Choukroun, E.~Kravchik, F.~Yang, and P.~Kisilev.
\newblock Low-bit quantization of neural networks for efficient inference.
\newblock In {\em ICCVW}, pages 3009--3018. IEEE, 2019.

\bibitem{chung2022come}
H.~Chung, B.~Sim, and J.~C. Ye.
\newblock Come-closer-diffuse-faster: Accelerating conditional diffusion models for inverse problems through stochastic contraction.
\newblock In {\em CVPR}, pages 12413--12422, 2022.

\bibitem{deng2009imagenet}
J.~Deng, W.~Dong, R.~Socher, L.-J. Li, K.~Li, and L.~Fei-Fei.
\newblock Imagenet: A large-scale hierarchical image database.
\newblock In {\em CVPR}, pages 248--255. Ieee, 2009.

\bibitem{dosovitskiy2020image}
A.~Dosovitskiy, L.~Beyer, A.~Kolesnikov, D.~Weissenborn, X.~Zhai, T.~Unterthiner, M.~Dehghani, M.~Minderer, G.~Heigold, S.~Gelly, et~al.
\newblock An image is worth 16x16 words: Transformers for image recognition at scale.
\newblock {\em arXiv preprint arXiv:2010.11929}, 2020.

\bibitem{esser2020lsq}
S.~K. Esser, J.~L. McKinstry, D.~Bablani, R.~Appuswamy, and D.~S. Modha.
\newblock Learned step size quantization.
\newblock In {\em ICLR}, 2020.

\bibitem{frantar2022optimal}
E.~Frantar and D.~Alistarh.
\newblock Optimal brain compression: A framework for accurate post-training quantization and pruning.
\newblock {\em NeurIPS}, 35:4475--4488, 2022.

\bibitem{franzese2022much}
G.~Franzese, S.~Rossi, L.~Yang, A.~Finamore, D.~Rossi, M.~Filippone, and P.~Michiardi.
\newblock How much is enough? a study on diffusion times in score-based generative models.
\newblock {\em arXiv preprint arXiv:2206.05173}, 2022.

\bibitem{gong2019dsq}
R.~Gong, X.~Liu, S.~Jiang, T.~Li, P.~Hu, J.~Lin, F.~Yu, and J.~Yan.
\newblock Differentiable soft quantization: Bridging full-precision and low-bit neural networks.
\newblock In {\em ICCV}, pages 4852--4861, 2019.

\bibitem{gong2024llmcbenchmarkinglargelanguage}
R.~Gong, Y.~Yong, S.~Gu, Y.~Huang, C.~Lv, Y.~Zhang, D.~Tao, and X.~Liu.
\newblock {LLMC}: Benchmarking large language model quantization with a versatile compression toolkit.
\newblock In F.~Dernoncourt, D.~Preo{\c{t}}iuc-Pietro, and A.~Shimorina, editors, {\em Proceedings of the 2024 Conference on Empirical Methods in Natural Language Processing: Industry Track}, pages 132--152, Miami, Florida, US, Nov. 2024. Association for Computational Linguistics.

\bibitem{goodfellow2020generative}
I.~Goodfellow, J.~Pouget-Abadie, M.~Mirza, B.~Xu, D.~Warde-Farley, S.~Ozair, A.~Courville, and Y.~Bengio.
\newblock Generative adversarial networks.
\newblock {\em Communications of the ACM}, 63(11):139--144, 2020.

\bibitem{NIPS2015_ae0eb3ee}
S.~Han, J.~Pool, J.~Tran, and W.~Dally.
\newblock Learning both weights and connections for efficient neural network.
\newblock In C.~Cortes, N.~Lawrence, D.~Lee, M.~Sugiyama, and R.~Garnett, editors, {\em NeurIPS}, volume~28. Curran Associates, Inc., 2015.

\bibitem{he2024efficientdm}
Y.~He, J.~Liu, W.~Wu, H.~Zhou, and B.~Zhuang.
\newblock Efficientdm: Efficient quantization-aware fine-tuning of low-bit diffusion models.
\newblock In {\em ICLR}, 2024.

\bibitem{he2023ptqd}
Y.~He, L.~Liu, J.~Liu, W.~Wu, H.~Zhou, and B.~Zhuang.
\newblock Ptqd: Accurate post-training quantization for diffusion models.
\newblock In {\em NeurIPS}, 2023.

\bibitem{hessel2022clipscore}
J.~Hessel, A.~Holtzman, M.~Forbes, R.~L. Bras, and Y.~Choi.
\newblock Clipscore: A reference-free evaluation metric for image captioning, 2022.

\bibitem{heusel2018gans}
M.~Heusel, H.~Ramsauer, T.~Unterthiner, B.~Nessler, and S.~Hochreiter.
\newblock Gans trained by a two time-scale update rule converge to a local nash equilibrium, 2018.

\bibitem{ho2020denoising}
J.~Ho, A.~Jain, and P.~Abbeel.
\newblock Denoising diffusion probabilistic models.
\newblock In {\em NeurIPS}, 2020.

\bibitem{ho2022classifierfree}
J.~Ho and T.~Salimans.
\newblock Classifier-free diffusion guidance, 2022.

\bibitem{Huang_2024_CVPR}
Y.~Huang, R.~Gong, J.~Liu, T.~Chen, and X.~Liu.
\newblock { TFMQ-DM: Temporal Feature Maintenance Quantization for Diffusion Models }.
\newblock In {\em 2024 IEEE/CVF Conference on Computer Vision and Pattern Recognition (CVPR)}, pages 7362--7371, Los Alamitos, CA, USA, June 2024. IEEE Computer Society.

\bibitem{huang2025qvgenpushinglimitquantized}
Y.~Huang, R.~Gong, J.~Liu, Y.~Ding, C.~Lv, H.~Qin, and J.~Zhang.
\newblock Qvgen: Pushing the limit of quantized video generative models, 2025.

\bibitem{huang2025harmonicaharmonizingtraininginference}
Y.~Huang, Z.~Wang, R.~Gong, J.~Liu, X.~Zhang, J.~Guo, X.~Liu, and J.~Zhang.
\newblock Harmonica: Harmonizing training and inference for better feature caching in diffusion transformer acceleration.
\newblock In {\em Forty-second International Conference on Machine Learning}, 2025.

\bibitem{hubara2020improvingadaquant}
I.~Hubara, Y.~Nahshan, Y.~Hanani, R.~Banner, and D.~Soudry.
\newblock Improving post training neural quantization: Layer-wise calibration and integer programming.
\newblock {\em arXiv preprint arXiv:2006.10518}, 2020.

\bibitem{openvino}
Intel.
\newblock Openvino.
\newblock \url{https://github.com/openvinotoolkit/openvino}, 2018.

\bibitem{jacob2018quantization}
B.~Jacob, S.~Kligys, B.~Chen, M.~Zhu, M.~Tang, A.~Howard, H.~Adam, and D.~Kalenichenko.
\newblock Quantization and training of neural networks for efficient integer-arithmetic-only inference.
\newblock In {\em CVPR}, pages 2704--2713, 2018.

\bibitem{jolicoeur2021gotta}
A.~Jolicoeur-Martineau, K.~Li, R.~Pich{\'e}-Taillefer, T.~Kachman, and I.~Mitliagkas.
\newblock Gotta go fast when generating data with score-based models.
\newblock {\em arXiv preprint arXiv:2105.14080}, 2021.

\bibitem{kang2023scaling}
M.~Kang, J.-Y. Zhu, R.~Zhang, J.~Park, E.~Shechtman, S.~Paris, and T.~Park.
\newblock Scaling up gans for text-to-image synthesis, 2023.

\bibitem{karras2018progressive}
T.~Karras, T.~Aila, S.~Laine, and J.~Lehtinen.
\newblock Progressive growing of gans for improved quality, stability, and variation, 2018.

\bibitem{karras2022elucidatingdesignspacediffusionbased}
T.~Karras, M.~Aittala, T.~Aila, and S.~Laine.
\newblock Elucidating the design space of diffusion-based generative models, 2022.

\bibitem{karras2019stylebased}
T.~Karras, S.~Laine, and T.~Aila.
\newblock A style-based generator architecture for generative adversarial networks, 2019.

\bibitem{kerr2017cutlass}
A.~Kerr, D.~Merrill, J.~Demouth, and J.~Tran.
\newblock Cutlass: Fast linear algebra in cuda c++.
\newblock {\em NVIDIA Developer Blog}, 2017.

\bibitem{kim2022denoisingMCMC}
B.~Kim and J.~C. Ye.
\newblock Denoising mcmc for accelerating diffusion-based generative models.
\newblock {\em arXiv preprint arXiv:2209.14593}, 2022.

\bibitem{kim2023bk}
B.-K. Kim, H.-K. Song, T.~Castells, and S.~Choi.
\newblock Bk-sdm: A lightweight, fast, and cheap version of stable diffusion.
\newblock {\em arXiv e-prints}, pages arXiv--2305, 2023.

\bibitem{kim2021comparing}
T.~Kim, J.~Oh, N.~Kim, S.~Cho, and S.-Y. Yun.
\newblock Comparing kullback-leibler divergence and mean squared error loss in knowledge distillation.
\newblock {\em arXiv preprint arXiv:2105.08919}, 2021.

\bibitem{kingma2021variational}
D.~Kingma, T.~Salimans, B.~Poole, and J.~Ho.
\newblock Variational diffusion models.
\newblock In {\em NeurIPS}, pages 21696--21707, 2021.

\bibitem{kingma2022autoencoding}
D.~P. Kingma and M.~Welling.
\newblock Auto-encoding variational bayes, 2022.

\bibitem{kong2021fastdpm}
Z.~Kong and W.~Ping.
\newblock On fast sampling of diffusion probabilistic models.
\newblock {\em arXiv preprint arXiv:2106.00132}, 2021.

\bibitem{krizhevsky2009learning}
A.~Krizhevsky.
\newblock Learning multiple layers of features from tiny images.
\newblock {\em University of Toronto}, 05 2012.

\bibitem{flux2024}
B.~F. Labs.
\newblock Flux.
\newblock \url{https://github.com/black-forest-labs/flux}, 2024.

\bibitem{lam2021bilateral}
M.~W.~Y. Lam, J.~Wang, D.~Su, and D.~Yu.
\newblock {BDDM:} bilateral denoising diffusion models for fast and high-quality speech synthesis.
\newblock In {\em ICLR}, 2022.

\bibitem{li2023qdiffusion}
X.~Li, L.~Lian, Y.~Liu, H.~Yang, Z.~Dong, D.~Kang, S.~Zhang, and K.~Keutzer.
\newblock Q-diffusion: Quantizing diffusion models.
\newblock In {\em ICCV}, 2023.

\bibitem{li2021brecq}
Y.~Li, R.~Gong, X.~Tan, Y.~Yang, P.~Hu, Q.~Zhang, F.~Yu, W.~Wang, and S.~Gu.
\newblock {BRECQ:} pushing the limit of post-training quantization by block reconstruction.
\newblock In {\em ICLR}, 2021.

\bibitem{li2023snapfusiontexttoimagediffusionmodel}
Y.~Li, H.~Wang, Q.~Jin, J.~Hu, P.~Chemerys, Y.~Fu, Y.~Wang, S.~Tulyakov, and J.~Ren.
\newblock Snapfusion: Text-to-image diffusion model on mobile devices within two seconds, 2023.

\bibitem{li2022efficient}
Z.~Li, C.~Guo, Z.~Zhu, Y.~Zhou, Y.~Qiu, X.~Gao, J.~Leng, and M.~Guo.
\newblock Efficient adaptive activation rounding for post-training quantization.
\newblock {\em arXiv preprint arXiv:2208.11945}, 2022.

\bibitem{lin2015microsoft}
T.-Y. Lin, M.~Maire, S.~Belongie, L.~Bourdev, R.~Girshick, J.~Hays, P.~Perona, D.~Ramanan, C.~L. Zitnick, and P.~Dollár.
\newblock Microsoft coco: Common objects in context, 2015.

\bibitem{liu2023pd}
J.~Liu, L.~Niu, Z.~Yuan, D.~Yang, X.~Wang, and W.~Liu.
\newblock Pd-quant: Post-training quantization based on prediction difference metric.
\newblock In {\em CVPR}, pages 24427--24437, 2023.

\bibitem{liu2022pseudo}
L.~Liu, Y.~Ren, Z.~Lin, and Z.~Zhao.
\newblock Pseudo numerical methods for diffusion models on manifolds.
\newblock In {\em ICLR}, 2022.

\bibitem{liu2024evalcrafterbenchmarkingevaluatinglarge}
Y.~Liu, X.~Cun, X.~Liu, X.~Wang, Y.~Zhang, H.~Chen, Y.~Liu, T.~Zeng, R.~Chan, and Y.~Shan.
\newblock Evalcrafter: Benchmarking and evaluating large video generation models, 2024.

\bibitem{lu2022dpmsolver}
C.~Lu, Y.~Zhou, F.~Bao, J.~Chen, C.~Li, and J.~Zhu.
\newblock Dpm-solver: {A} fast {ODE} solver for diffusion probabilistic model sampling in around 10 steps.
\newblock In {\em NeurIPS}, 2022.

\bibitem{lu2023dpmsolver}
C.~Lu, Y.~Zhou, F.~Bao, J.~Chen, C.~Li, and J.~Zhu.
\newblock Dpm-solver++: Fast solver for guided sampling of diffusion probabilistic models, 2023.

\bibitem{luhman2021kdingdm}
E.~Luhman and T.~Luhman.
\newblock Knowledge distillation in iterative generative models for improved sampling speed.
\newblock {\em arXiv preprint arXiv:2101.02388}, 2021.

\bibitem{lyu2022accelerating}
Z.~Lyu, X.~Xu, C.~Yang, D.~Lin, and B.~Dai.
\newblock Accelerating diffusion models via early stop of the diffusion process.
\newblock {\em arXiv preprint arXiv:2205.12524}, 2022.

\bibitem{ma2024deepcache}
X.~Ma, G.~Fang, and X.~Wang.
\newblock Deepcache: Accelerating diffusion models for free.
\newblock In {\em CVPR}, pages 15762--15772, 2024.

\bibitem{mao2023cross}
A.~Mao, M.~Mohri, and Y.~Zhong.
\newblock Cross-entropy loss functions: Theoretical analysis and applications.
\newblock In {\em ICLR}, pages 23803--23828. PMLR, 2023.

\bibitem{kl}
S.~Migacz.
\newblock Nvidia 8-bit inference width tensorrt.
\newblock {\em GPU Technology Conference}, 2017.

\bibitem{nagel2020adaround}
M.~Nagel, R.~A. Amjad, M.~van Baalen, C.~Louizos, and T.~Blankevoort.
\newblock Up or down? adaptive rounding for post-training quantization.
\newblock In {\em ICML}, volume 119, pages 7197--7206, 2020.

\bibitem{nagel2021white}
M.~Nagel, M.~Fournarakis, R.~A. Amjad, Y.~Bondarenko, M.~van Baalen, and T.~Blankevoort.
\newblock A white paper on neural network quantization, 2021.

\bibitem{nguyen2024swiftbrush}
T.~H. Nguyen and A.~Tran.
\newblock Swiftbrush: One-step text-to-image diffusion model with variational score distillation.
\newblock In {\em CVPR}, pages 7807--7816, 2024.

\bibitem{nichol2021improvedDDPM}
A.~Q. Nichol and P.~Dhariwal.
\newblock Improved denoising diffusion probabilistic models.
\newblock In {\em ICML}, pages 8162--8171, 2021.

\bibitem{trt}
Nvidia.
\newblock Tensorrt.
\newblock \url{https://github.com/NVIDIA/TensorRT}, 2019.

\bibitem{stable-diffusion-coreml-apple-silicon}
A.~Orhon, M.~Siracusa, and A.~Wadhwa.
\newblock Stable diffusion with core ml on apple silicon, 2022.

\bibitem{pandey2023softmax}
N.~P. Pandey, M.~Fournarakis, C.~Patel, and M.~Nagel.
\newblock Softmax bias correction for quantized generative models, 2023.

\bibitem{paszke2019pytorch}
A.~Paszke, S.~Gross, F.~Massa, A.~Lerer, J.~Bradbury, G.~Chanan, T.~Killeen, Z.~Lin, N.~Gimelshein, L.~Antiga, et~al.
\newblock Pytorch: An imperative style, high-performance deep learning library.
\newblock {\em NeurIPS}, 32, 2019.

\bibitem{Peebles2022DiT}
W.~Peebles and S.~Xie.
\newblock Scalable diffusion models with transformers.
\newblock {\em arXiv preprint arXiv:2212.09748}, 2022.

\bibitem{peebles2023scalable}
W.~Peebles and S.~Xie.
\newblock Scalable diffusion models with transformers.
\newblock In {\em ICCV}, pages 4195--4205, 2023.

\bibitem{podell2023sdxlimprovinglatentdiffusion}
D.~Podell, Z.~English, K.~Lacey, A.~Blattmann, T.~Dockhorn, J.~Müller, J.~Penna, and R.~Rombach.
\newblock Sdxl: Improving latent diffusion models for high-resolution image synthesis, 2023.

\bibitem{10.1007/978-3-319-46493-0_32}
M.~Rastegari, V.~Ordonez, J.~Redmon, and A.~Farhadi.
\newblock Xnor-net: Imagenet classification using binary convolutional neural networks.
\newblock In {\em ECCV}, pages 525--542, 2016.

\bibitem{ren2022fastspeech}
Y.~Ren, C.~Hu, X.~Tan, T.~Qin, S.~Zhao, Z.~Zhao, and T.-Y. Liu.
\newblock Fastspeech 2: Fast and high-quality end-to-end text to speech, 2022.

\bibitem{rombach2022ldm}
R.~Rombach, A.~Blattmann, D.~Lorenz, P.~Esser, and B.~Ommer.
\newblock High-resolution image synthesis with latent diffusion models.
\newblock In {\em CVPR}, 2022.

\bibitem{ronneberger2015unet}
O.~Ronneberger, P.~Fischer, and T.~Brox.
\newblock U-net: Convolutional networks for biomedical image segmentation, 2015.

\bibitem{salimans2016improved}
T.~Salimans, I.~Goodfellow, W.~Zaremba, V.~Cheung, A.~Radford, and X.~Chen.
\newblock Improved techniques for training gans, 2016.

\bibitem{salimans2022progressivedistillation}
T.~Salimans and J.~Ho.
\newblock Progressive distillation for fast sampling of diffusion models.
\newblock In {\em ICLR}, 2022.

\bibitem{sauer2023adversarial}
A.~Sauer, D.~Lorenz, A.~Blattmann, and R.~Rombach.
\newblock Adversarial diffusion distillation.
\newblock {\em arXiv preprint arXiv:2311.17042}, 2023.

\bibitem{shang2022ptq4dm}
Y.~Shang, Z.~Yuan, B.~Xie, B.~Wu, and Y.~Yan.
\newblock Post-training quantization on diffusion models.
\newblock In {\em CVPR}, 2023.

\bibitem{shen2018natural}
J.~Shen, R.~Pang, R.~J. Weiss, M.~Schuster, N.~Jaitly, Z.~Yang, Z.~Chen, Y.~Zhang, Y.~Wang, R.~Skerry-Ryan, R.~A. Saurous, Y.~Agiomyrgiannakis, and Y.~Wu.
\newblock Natural tts synthesis by conditioning wavenet on mel spectrogram predictions, 2018.

\bibitem{so2023temporal}
J.~So, J.~Lee, D.~Ahn, H.~Kim, and E.~Park.
\newblock Temporal dynamic quantization for diffusion models.
\newblock In {\em NeurIPS}, 2023.

\bibitem{songddim}
J.~Song, C.~Meng, and S.~Ermon.
\newblock Denoising diffusion implicit models.
\newblock In {\em ICLR}, 2021.

\bibitem{song2020score}
Y.~Song, J.~Sohl{-}Dickstein, D.~P. Kingma, A.~Kumar, S.~Ermon, and B.~Poole.
\newblock Score-based generative modeling through stochastic differential equations.
\newblock In {\em ICLR}, 2021.

\bibitem{sui2024bitsfusion199bitsweight}
Y.~Sui, Y.~Li, A.~Kag, Y.~Idelbayev, J.~Cao, J.~Hu, D.~Sagar, B.~Yuan, S.~Tulyakov, and J.~Ren.
\newblock Bitsfusion: 1.99 bits weight quantization of diffusion model, 2024.

\bibitem{wang2023towards}
C.~Wang, Z.~Wang, X.~Xu, Y.~Tang, J.~Zhou, and J.~Lu.
\newblock Towards accurate data-free quantization for diffusion models.
\newblock {\em arXiv preprint arXiv:2305.18723}, 2023.

\bibitem{wang2024quest}
H.~Wang, Y.~Shang, Z.~Yuan, J.~Wu, and Y.~Yan.
\newblock Quest: Low-bit diffusion model quantization via efficient selective finetuning, 2024.

\bibitem{wang2020towards}
P.~Wang, Q.~Chen, X.~He, and J.~Cheng.
\newblock Towards accurate post-training network quantization via bit-split and stitching.
\newblock In {\em ICML}, pages 9847--9856, 2020.

\bibitem{wnag2024ptsbenchcomprehensiveposttrainingsparsity}
Z.~Wang, J.~Guo, R.~Gong, Y.~Yong, A.~Liu, Y.~Huang, J.~Liu, and X.~Liu.
\newblock Ptsbench: A comprehensive post-training sparsity benchmark towards algorithms and models.
\newblock In {\em Proceedings of the 32nd ACM International Conference on Multimedia}, MM '24, page 5742–5751, New York, NY, USA, 2024. Association for Computing Machinery.

\bibitem{watson2022learning}
D.~Watson, W.~Chan, J.~Ho, and M.~Norouzi.
\newblock Learning fast samplers for diffusion models by differentiating through sample quality.
\newblock In {\em ICLR}, 2022.

\bibitem{wei2022qdrop}
X.~Wei, R.~Gong, Y.~Li, X.~Liu, and F.~Yu.
\newblock Qdrop: Randomly dropping quantization for extremely low-bit post-training quantization.
\newblock In {\em ICLR}, 2022.

\bibitem{wu2020integer}
H.~Wu, P.~Judd, X.~Zhang, M.~Isaev, and P.~Micikevicius.
\newblock Integer quantization for deep learning inference: Principles and empirical evaluation.
\newblock {\em arXiv preprint arXiv:2004.09602}, 2020.

\bibitem{wu2023exploringvideoqualityassessment}
H.~Wu, E.~Zhang, L.~Liao, C.~Chen, J.~Hou, A.~Wang, W.~Sun, Q.~Yan, and W.~Lin.
\newblock Exploring video quality assessment on user generated contents from aesthetic and technical perspectives, 2023.

\bibitem{wu2024ptq4ditposttrainingquantizationdiffusion}
J.~Wu, H.~Wang, Y.~Shang, M.~Shah, and Y.~Yan.
\newblock Ptq4dit: Post-training quantization for diffusion transformers, 2024.

\bibitem{yang2023diffusion}
X.~Yang, D.~Zhou, J.~Feng, and X.~Wang.
\newblock Diffusion probabilistic model made slim.
\newblock In {\em CVPR}, pages 22552--22562, 2023.

\bibitem{yu2016lsun}
F.~Yu, A.~Seff, Y.~Zhang, S.~Song, T.~Funkhouser, and J.~Xiao.
\newblock Lsun: Construction of a large-scale image dataset using deep learning with humans in the loop, 2016.

\bibitem{zhang2023root}
L.~Zhang, Y.~He, Z.~Lou, X.~Ye, Y.~Wang, and H.~Zhou.
\newblock Root quantization: a self-adaptive supplement ste.
\newblock {\em Applied Intelligence}, 53(6):6266--6275, 2023.

\bibitem{zhang2022fast}
Q.~Zhang and Y.~Chen.
\newblock Fast sampling of diffusion models with exponential integrator.
\newblock {\em arXiv preprint arXiv:2204.13902}, 2022.

\bibitem{zhang2022gddim}
Q.~Zhang, M.~Tao, and Y.~Chen.
\newblock gddim: Generalized denoising diffusion implicit models.
\newblock {\em arXiv preprint arXiv:2206.05564}, 2022.

\bibitem{zhang2022opt}
S.~Zhang, S.~Roller, N.~Goyal, M.~Artetxe, M.~Chen, S.~Chen, C.~Dewan, M.~Diab, X.~Li, X.~V. Lin, T.~Mihaylov, M.~Ott, S.~Shleifer, K.~Shuster, D.~Simig, P.~S. Koura, A.~Sridhar, T.~Wang, and L.~Zettlemoyer.
\newblock Opt: Open pre-trained transformer language models, 2022.

\bibitem{zhao2025viditqefficientaccuratequantization}
T.~Zhao, T.~Fang, H.~Huang, E.~Liu, R.~Wan, W.~Soedarmadji, S.~Li, Z.~Lin, G.~Dai, S.~Yan, H.~Yang, X.~Ning, and Y.~Wang.
\newblock Vidit-q: Efficient and accurate quantization of diffusion transformers for image and video generation, 2025.

\bibitem{zhao2023mobilediffusion}
Y.~Zhao, Y.~Xu, Z.~Xiao, and T.~Hou.
\newblock Mobilediffusion: Subsecond text-to-image generation on mobile devices.
\newblock {\em arXiv preprint arXiv:2311.16567}, 2023.

\bibitem{zheng2022truncated}
H.~Zheng, P.~He, W.~Chen, and M.~Zhou.
\newblock Truncated diffusion probabilistic models.
\newblock {\em stat}, 1050:7, 2022.

\bibitem{opensora}
Z.~Zheng, X.~Peng, T.~Yang, C.~Shen, S.~Li, H.~Liu, Y.~Zhou, T.~Li, and Y.~You.
\newblock Open-sora: Democratizing efficient video production for all, 2024.

\bibitem{pmlr-v162-zhou22c}
J.~Zhou, X.~Li, T.~Ding, C.~You, Q.~Qu, and Z.~Zhu.
\newblock On the optimization landscape of neural collapse under {MSE} loss: Global optimality with unconstrained features.
\newblock In K.~Chaudhuri, S.~Jegelka, L.~Song, C.~Szepesvari, G.~Niu, and S.~Sabato, editors, {\em ICML}, volume 162 of {\em Proceedings of Machine Learning Research}, pages 27179--27202. PMLR, 17--23 Jul 2022.

\bibitem{zhuang2018towards}
B.~Zhuang, C.~Shen, M.~Tan, L.~Liu, and I.~Reid.
\newblock Towards effective low-bitwidth convolutional neural networks.
\newblock In {\em CVPR}, pages 7920--7928, 2018.

\end{thebibliography}
}

\begin{IEEEbiography}[{\includegraphics[width=1in,height=1.25in,clip,keepaspectratio]{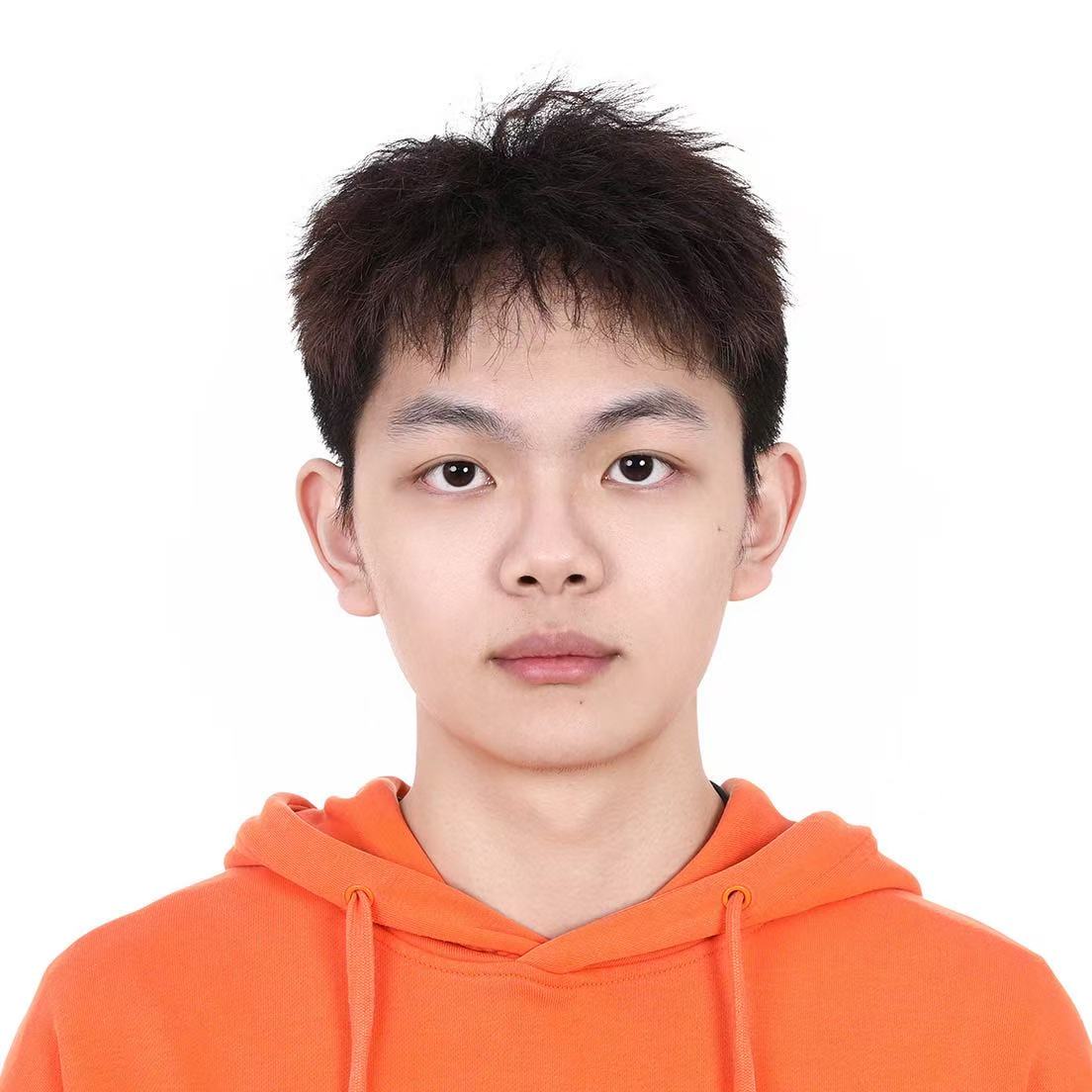}}]
{Yushi Huang} received the B.E. in Shenyuan Honors College, Beihang University, in 2024. He is pursuing his Ph.D. at the Hong Kong University of Science and Technology, supervised by Prof. Jun Zhang. He is interested in efficient AI (\emph{e.g.}, model compression and acceleration) and generative modeling (\emph{e.g.}, diffusion models and large language models).
\end{IEEEbiography}

\begin{IEEEbiography}[{\includegraphics[width=1in,height=1.25in,clip,keepaspectratio]{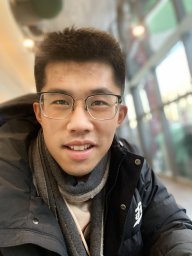}}]
{Ruihao Gong} received B.E. and M.E. degrees in computer science from Beihang University in 2018 and 2021, respectively. He is currently a Ph.D. candidate student at Beihang University, Beijing, China, and a vice director at SenseTime Research, Beijing, China. His research interests include efficient deep learning, model compression, and machine learning systems for large-scale, high-performance, low-power, and low-cost deployment.
\end{IEEEbiography}

\begin{IEEEbiography}[{\includegraphics[width=1in,height=1.25in,clip,keepaspectratio]{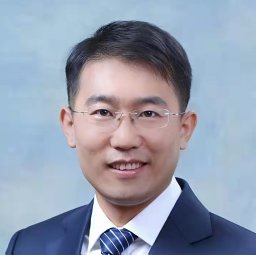}}]
{Xianglong Liu} (Member, IEEE) received the BS and
PhD degrees under the supervision of Prof.Wei Li and
visited DVMM Lab, Columbia University as a joint
PhD student supervised by Prof. Shih-Fu Chang. He is
a full professor with the School of Computer Science
and Engineering, Beihang University. His research interests include fast visual computing (e.g., large-scale
search/understanding) and robust deep learning (e.g.,
network quantization, adversarial attack/defense, and
few-shot learning).
\end{IEEEbiography}

\begin{IEEEbiography}[{\includegraphics[width=1in,height=1.25in,clip,keepaspectratio]{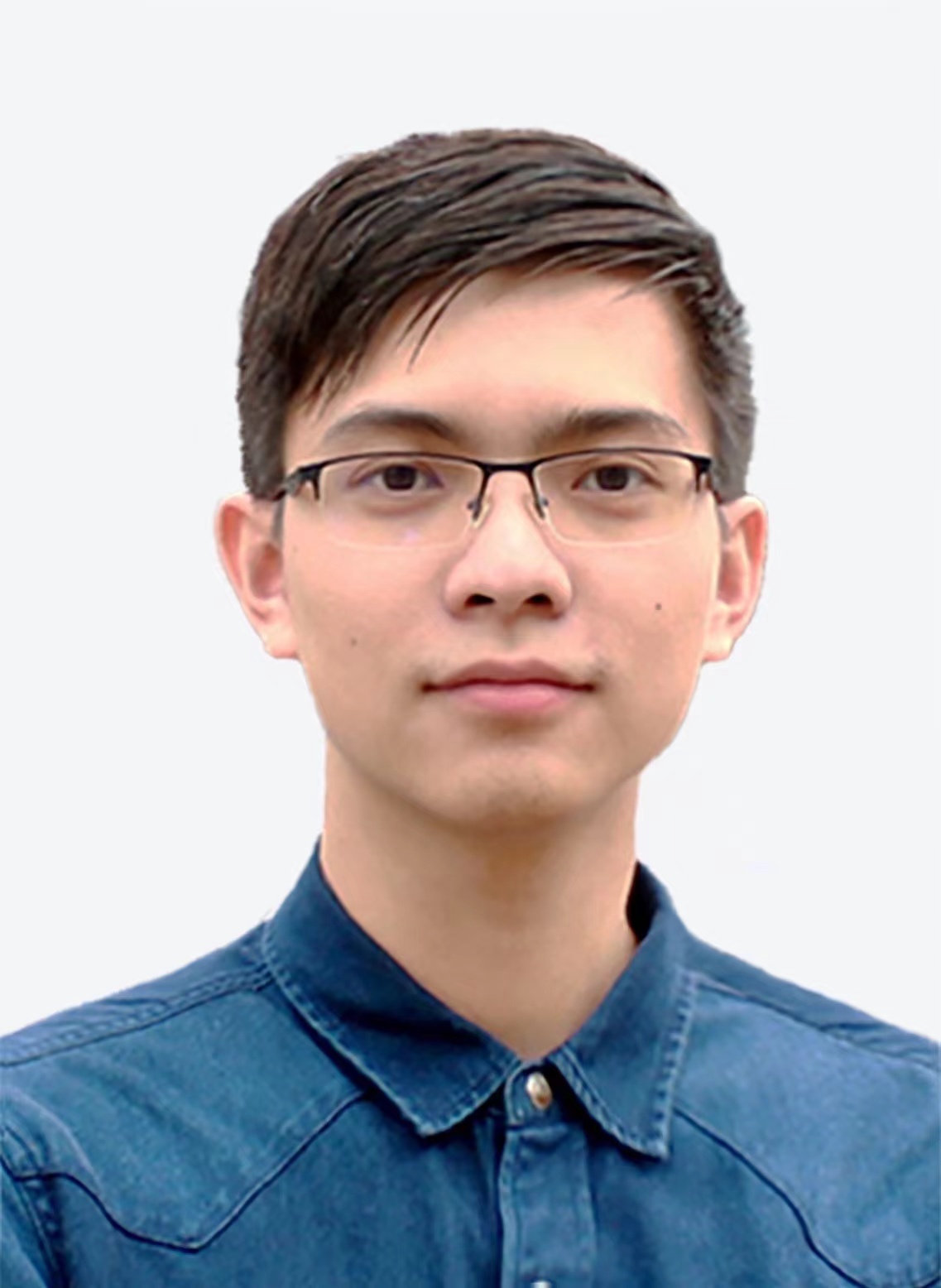}}]
{Jing Liu} is a Ph.D. student in the Faculty of Information Technology, Monash University Clayton Campus, Australia. He received his Bachelor Degree in 2017 and Master Degree in 2020, both from the School of Software Engineering at South China University of Technology, China.
His research interests include computer vision, model compression and acceleration.
\end{IEEEbiography}

\begin{IEEEbiography}[{\includegraphics[width=1in,height=1.25in,clip,keepaspectratio]{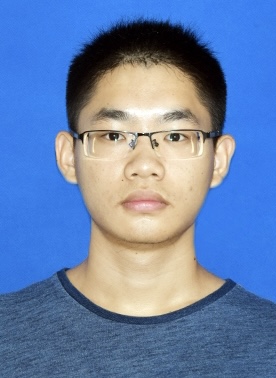}}]
{Yuhang Li} received the B.E. in Department of Computer Science and Technology, University of Electronic Science and Technology of China in 2020. He was a research assistant in National University of Singapore and University of Electronic Science and Technology of China in 2019 and 2021, respectively. Now he is pursuing his Ph.D. degree in Yale University, supervised by Prof. Priyadarshini Panda. His research interests include Efficient Deep Learning, Brain-inspired Computing, Model Compression. 
\end{IEEEbiography}

\begin{IEEEbiography}[{\includegraphics[width=1in,height=1.25in,clip,keepaspectratio]{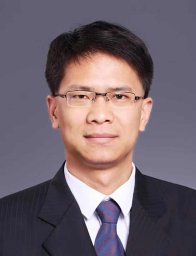}}]
{Jiwen Lu} (Fellow, IEEE) received the PhD
degree in electrical and electronic engineering from
Nanyang Technological University, Singapore. He
is an Associate Professor with the Department of
Automation, Tsinghua University, China. His research interests include computer vision (object detection, scene understanding, visual forensics), machine learning (foundation models, representation
learning, similarity learning), and unmanned systems
(autonomous driving, robotic grasping, visual navigation).
\end{IEEEbiography}

\begin{IEEEbiography}[{\includegraphics[width=1in,height=1.25in,clip,keepaspectratio]{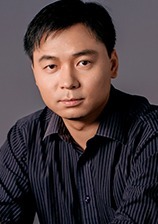}}]
{Dacheng Tao} (Fellow, IEEE) is currently a Distinguished University Professor in the School of Computer Science and Engineering at Nanyang Technological University. He mainly applies statistics and mathematics to artificial intelligence and data science, and his research is detailed in one monograph and over 200 publications in prestigious journals and proceedings at leading conferences, with best paper awards, best student paper awards, and test-of-time awards. His publications have been cited over 112K times and he has an h-index 160+ in Google Scholar. He received the 2015 and 2020 Australian Eureka Prize, the 2018 IEEE ICDM Research Contributions Award, and the 2021 IEEE Computer Society McCluskey Technical Achievement Award. He is a Fellow of the Australian Academy of Science, AAAS, ACM, and IEEE. \end{IEEEbiography}

\newpage
\onecolumn

\begin{LARGE}
~~~\vspace{1pt}
\begin{center}
    \bf Appendix
\end{center}
\end{LARGE}
\vspace{2pt}

\renewcommand{\thefigure}{\Roman{figure}}
\renewcommand{\theequation}{\Alph{equation}}
\renewcommand{\thetable}{\Roman{table}}
\renewcommand\thesection{\Alph{section}}
\setcounter{equation}{0}
\setcounter{section}{0}
\setcounter{figure}{0}
\setcounter{table}{0}

We organize the supplementary material as follows:
\begin{itemize}[leftmargin=*]
    \item In Sec.~\ref{app:relation}, we provide a detailed relationship between temporal feature errors and final generation quality;
    \item In Sec.~\ref{app:induce-vali}, we conduct more experiments to validate the inappropriate reconstruction target of previous methods;
    \item In Sec.~\ref{app:video}, we explore the effectiveness of our framework on video generation;
    \item In Sec.~\ref{app:guide}, we provide a guideline on when to use which proposed strategies;
    \item In Sec.~\ref{app:sub-4}, we conduct experiments to show the superiority of our framework compared to previous PTQ methods under sub-$4$-bit quantization;
    \item In Sec.~\ref{app:flux}, we apply our framework to the advanced step-distilled model (\emph{i.e.}, FLUX.1-Schnell);
    \item In Sec.~\ref{app:edge}, we deploy our quantized models on edge and mobile devices;
    \item In Sec.~\ref{app:vis}, we offer comprehensive visualization comparison across different methods.
\end{itemize}

\section{Relationship between Temporal Feature Errors and Final Generation Quality}\label{app:relation}
\yushi{In this section, we sort out the relationship between temporal feature error and final generation quality. First, we define temporal feature disturbance as a significant temporal feature error (Sec.~\ref{sec:sensitivity_disturbance}). Then we show that temporal feature disturbance (\emph{i.e.}, significant temporal feature error) accounts for temporal information mismatch (Sec.~\ref{sec:impact}), which leads to trajectory deviation (Sec.~\ref{sec:impact}). In Fig.~\ref{denoising_compare}, the deviated trajectory can result in noisy images gradually shifting from a normal denoising trajectory. Thus, the image quality degrades severely. Moreover, as reflected in Fig.~\ref{fig:sensitivity}, with temporal feature error increasing\footnote{Obviously, temporal feature error which can also be defined by $\mathcal{L}^{temporal}_{t,i}$ in Eq.~(\ref{eq:cache}) increases with the increase of noise level $\lambda$.}, generation quality measured by FID score deteriorates more sharply.}

\section{Inappropriate Reconstruction Target Validation}\label{app:induce-vali}
\begin{table}[!ht]\setlength{\tabcolsep}{10pt}
  \centering
  \caption{FID, sFID and SQNR on LSUN-Bedrooms $256\times256$~\cite{yu2016lsun} for LDM-4. ``Freeze'' denotes the trial the same as that in Tab.~\ref{tab:recon_comp}.} 
  \resizebox{0.48\linewidth}{!}{
  \begin{tabular}{lclll}
    \toprule
    \textbf{Methods} & \textbf{\#Bits (W/A)} & FID$\downarrow$ & sFID$\downarrow$ & SQNR$\uparrow$\\
    \midrule
    Full Prec. & 32/32 & 2.98 & 7.09 & -\\
    \midrule
    PTQ4DM~\cite{shang2022ptq4dm} & 8/8 & 4.75 & 9.59 & 5.16\\
    \rowcolor[gray]{0.92} +Freeze & 8/8 & \textbf{4.21\textsubscript{\bl{-0.54}}} & \textbf{8.36\textsubscript{\bl{-1.23}}} & \textbf{6.01\textsubscript{\mg{+0.85}}}\\

    Q-Diffusion~\cite{li2023qdiffusion} & 8/8 & 4.51 & 8.17 & 5.21\\
    \rowcolor[gray]{0.92} +Freeze & 8/8 & \textbf{4.12\textsubscript{\bl{-0.39}}} & \textbf{7.96\textsubscript{\bl{-0.21}}} & \textbf{6.38\textsubscript{\mg{+1.17}}}\\

    PTQD~\cite{he2023ptqd} & 8/8 & 3.75 & 9.89 & 6.60\\
    \rowcolor[gray]{0.92} +Freeze & 8/8 & \textbf{3.52\textsubscript{\bl{-0.23}}} & \textbf{8.68\textsubscript{\bl{-1.21}}} & \textbf{7.22\textsubscript{\mg{+0.62}}}\\

    \midrule
    PTQ4DM~\cite{shang2022ptq4dm} & 4/8 & 20.72 & 54.30 & 1.42\\
    \rowcolor[gray]{0.92} +Freeze & 4/8 & \textbf{15.38\textsubscript{\bl{-5.34}}} & \textbf{32.21\textsubscript{\bl{-22.09}}} & \textbf{3.02\textsubscript{\mg{+1.60}}}\\

    Q-Diffusion~\cite{li2023qdiffusion} & 4/8 & 6.40 & 17.93 & 3.89\\
    \rowcolor[gray]{0.92} +Freeze & 4/8 & \textbf{5.22\textsubscript{\bl{-1.18}}} & \textbf{12.01\textsubscript{\bl{-5.92}}} & \textbf{5.94\textsubscript{\mg{+2.05}}}\\

    PTQD~\cite{he2023ptqd} & 4/8 & 5.94 & 15.16 & 4.42\\
    \rowcolor[gray]{0.92} +Freeze & 4/8 & \textbf{4.73\textsubscript{\bl{-1.21}}} & \textbf{10.08\textsubscript{\bl{-5.08}}} & \textbf{6.24\textsubscript{\mg{+1.82}}}\\
    \bottomrule
\end{tabular}
}
    \label{tab:induce-vali}
\end{table}
\yushi{We further conduct the validation experiments for the baselines (\emph{i.e.}, PTQ4DM~\cite{shang2022ptq4dm}, Q-Diffusion~\cite{li2023qdiffusion}, and PTQD~\cite{he2023ptqd}) in Tab.~\ref{tab:sota_ldm}. As shown in Tab.~\ref{tab:induce-vali}, the Freeze strategy still significantly improves the performance, akin to that in Tab.~\ref{tab:recon_comp}. For example, it decreases FID and sFID by $\mathbf{1.18}$ and $\mathbf{5.92}$ for W$4$A$8$ Q-Diffusion. Hence, the results further validate our analysis and make it more convincing.}

\section{Exploration for Video Generation}\label{app:video}
\begin{table}[!ht]\setlength{\tabcolsep}{8pt}
  \centering
  \caption{Quantization results for text-guided video generation with OpenSora~\cite{opensora} on OpenSora prompt set~\cite{opensora}. We implement our method based on PTQ4DiT.} 
  \resizebox{0.6\linewidth}{!}{
  \begin{tabular}[t!]{lclllll}
\toprule
\multirow{2}{*}{\textbf{Method}} & \multirow{2}{*}{\textbf{\#Bits (W/A)}} & \multirow{2}{*}{CLIPSIM$\uparrow$} &  \multirow{2}{*}{\makecell{CLIP-\\Temp}$\uparrow$} & \multirow{2}{*}{\makecell{VQA-\\Aesthetic}$\uparrow$} & \multirow{2}{*}{\makecell{VQA-\\Technical}$\uparrow$} & \multirow{2}{*}{\makecell{$\Delta$Flow\\Score}$\downarrow$}  \\
 & & & & & & \\
  \midrule
 Full Prec. & 16/16 &  0.1797 & 0.9988 & 63.40 & 50.46 & - \\

 \midrule
  Q-DiT~\cite{chen2024qditaccurateposttrainingquantization} & 8/8 & 0.1788 & 0.9977 & 61.03 & 34.97 & 0.473  \\
  PTQ4DiT~\cite{wu2024ptq4ditposttrainingquantizationdiffusion} & 8/8 & 0.1836 & 0.9991 & 54.56 & 53.33 & 0.440 \\
  
    \rowcolor[gray]{0.9}TM & 8/8 & 0.1932\textsubscript{~\mg{+0.0096}} & 0.9992\textsubscript{~\mg{+0.0001}} & 57.87\textsubscript{~\mg{+3.31}} & 53.38\textsubscript{~\mg{+0.05}} & 0.301\textsubscript{~\bl{-0.139}}\\

    \rowcolor[gray]{0.9}CM & 8/8 & 0.1928\textsubscript{~\mg{+0.0092}} & \textbf{0.9993\textsubscript{~\mg{+0.0002}}} & 58.62\textsubscript{~\mg{+4.06}} & 53.41\textsubscript{~\mg{+0.08}} & 0.312\textsubscript{~\bl{-0.128}}\\

    \rowcolor[gray]{0.9}DS & 8/8 & \textbf{0.1944\textsubscript{~\mg{+0.0108}}} & \textbf{0.9993\textsubscript{~\mg{+0.0003}}} & \textbf{58.94\textsubscript{~\mg{+4.38}}} & \textbf{53.87\textsubscript{~\mg{+0.54}}} & \textbf{0.298\textsubscript{~\bl{-0.142}}}\\

 \midrule
   Q-DiT~\cite{chen2024qditaccurateposttrainingquantization} & 8/8 & 0.1687 & 0.9833 & 0.007 & 0.018 & 3.013  \\
  PTQ4DiT~\cite{wu2024ptq4ditposttrainingquantizationdiffusion} & 8/8 &  0.1735 & 0.9973 & 2.210 & 0.318 & 0.108 \\
  
   \rowcolor[gray]{0.9}TM & 4/8 & 0.1802\textsubscript{~\mg{+0.0067}} & 0.9982\textsubscript{~\mg{+0.0009}} &9.171\textsubscript{~\mg{+6.26}} & 4.713\textsubscript{~\mg{+4.40}}&0.253\textsubscript{~\bl{-0.145}}\\

    \rowcolor[gray]{0.9}CM & 4/8 & 0.1807\textsubscript{~\mg{+0.0072}} & 0.9979\textsubscript{~\mg{+0.0006}} &10.22\textsubscript{~\mg{+8.01}} & 2.056\textsubscript{~\mg{+1.74}}&0.232\textsubscript{~\bl{-0.124}}\\

    \rowcolor[gray]{0.9}DS & 4/8 & \textbf{0.1809\textsubscript{~\mg{+0.0074}}} & \textbf{0.9986\textsubscript{~\mg{+0.0013}}} &\textbf{11.23\textsubscript{~\mg{+9.02}}} & \textbf{5.122\textsubscript{~\mg{+4.81}}}&\textbf{0.261\textsubscript{~\bl{-0.153}}}\\
\bottomrule
\end{tabular}
}
    \label{tab:open-sora}
\end{table}
\yushi{Advanced Video diffusion models employ DiT~\cite{Peebles2022DiT} as their backbones. We have found that this type of backbone has temporal features that can be formulated by Eq. (\ref{emb_rel}), where $g_i(\cdot)$ is a simple multi-layer perception (MLP) network. Moreover, temporal features and modules generating them are also independent of any $\bx_t$. Thus, we can directly use our framework for these video diffusion models. Due to similar architectures and
algorithms, we believe our approach can also achieve superior performance. As shown in Tab.~\ref{tab:open-sora}, we conduct experiments on OpenSora~\cite{opensora} with $100$-steps DDIM~\cite{songddim} and \texttt{cfg} scale of $4.0$. We select representative metrics and measure them on $10$ examples in the OpenSora prompts set~\cite{opensora}, the same as the previous study~\cite{zhao2025viditqefficientaccuratequantization}. Specifically, following EvalCrafter~\cite{liu2024evalcrafterbenchmarkingevaluatinglarge}, we select CLIPSIM and CLIP-Temp to measure the text-video alignment and temporal semantic consistency. DOVER~\cite{wu2023exploringvideoqualityassessment}’s video quality
assessment (VQA) metrics to evaluate the generation quality from aesthetic and technical aspects are also involved. $\Delta$Flow Score~\cite{liu2024evalcrafterbenchmarkingevaluatinglarge} is used for evaluating the temporal
consistency. We employ $8$ prompts from the OpenSora prompt set to quantize the model. Other quantization settings are the same as those of PTQ4DiT~\cite{wu2024ptq4ditposttrainingquantizationdiffusion}. Compared with methods tailored for DiT architecture in Tab.~\ref{tab:open-sora}, our framework outperforms them across all metrics. We will explore more about video generation in the future.}

\section{Guidelines on When to Use Which Proposed Strategies}\label{app:guide}
\yushi{We provide a guideline here on when to use which maintenance strategy. First, we clarify that the time overhead~\footnote{The additional time overhead only related to the process to obtain quantized temporal features is unrelated to $T$.} of the two maintenance strategies is all less than $\mathbf{5\%}$ as shown in Fig.~\ref{fig:latency}. Only considering memory overhead, both methods might be suitable for different scenarios. For example, to accommodate a dynamic timestep range of $1\sim1000$, solely storing temporal features without $\{g_i\}_{i=1,\ldots,n}\cup\{h\}$ (\emph{i.e.}, Cache-based Maintenance) for W$4$A$8$ LDM-8 on LSUN-Churches $256\times256$~\cite{yu2016lsun} requires an additional $\mathbf{20.7}$MB. It is a significant extra cost amounting to roughly $\mathbf{14.7\%}$ of the quantized UNet size (\emph{i.e.}, $140.9$MB). In contrast, TIB-based Maintenance barely incurs less than $\mathbf{1\%}$ memory overhead in this scenario. However, Cache-based Maintenance can also require less storage in other cases, \emph{e.g.}, $\mathbf{26.65}$MB memory reduction for Stable Diffusion than TIB-based Maintenance as depicted in Fig.~\ref{fig:memory}. In conclusion, users can choose these two maintenance methods for deployment according to their specific hardware, diffusion backbone, inference arguments, \emph{etc}. Without considering any resource restriction, we suggest users use both methods with our selection approach for higher performance.}

\section{Sub-4-bit Quantization Results}\label{app:sub-4}
\begin{table}[!ht]\setlength{\tabcolsep}{8pt}
  \centering
  \caption{Sub-$4$-bit quantization results for unconditional/class-conditional image generation with LDM. Experimental details are the same as those in the main text.} 
  \resizebox{0.8\linewidth}{!}{
  \begin{tabular}{lclllllll}
    \toprule
    \multicolumn{1}{c}{\multirow{2}{*}{\textbf{Methods}}} & \multicolumn{1}{c}{\multirow{2}{*}{\textbf{Bits (W/A)}}} & \multicolumn{3}{c}{\textbf{ImageNet $256 \times 256$}} & \multicolumn{2}{c}{\textbf{LSUN-Bedrooms $256 \times 256$}} & \multicolumn{2}{c}{\textbf{LSUN-Churches $256 \times 256$}}\\ \cmidrule(r){3-5} \cmidrule(r){6-7} \cmidrule(r){8-9} 
    & & IS$\uparrow$ & FID$\downarrow$ & sFID$\downarrow$ & FID$\downarrow$ & sFID$\downarrow$ & FID$\downarrow$ & sFID$\downarrow$ \\
    \midrule
    Full Prec. & 32/32 & 235.64 & 10.91 &  7.67 & 2.98 & 7.09 & 4.12 & 10.89 \\
    \midrule
     PTQ4DM~\cite{shang2022ptq4dm} & 4/4 & 25.61 &  213.45 & 242.56 & 87.64 & 41.08 & 70.26 & 48.21 \\
    Q-Diffusion~\cite{li2023qdiffusion} & 4/4 & 29.74 & 202.87 &  221.97 & 79.42 & 35.31 & 59.98 & 42.35 \\
    PTQD~\cite{he2023ptqd} & 4/4 & 34.67 & 186.45 &  197.32 & 72.66 & 31.24 & 52.46 & 38.21 \\
    
    \rowcolor[gray]{0.9}TM & 4/4 & 41.68\textsubscript{~\mg{+7.01}} & 157.21\textsubscript{~\bl{-29.24}} &  169.32\textsubscript{~\bl{-28.00}} & 61.02\textsubscript{~\bl{-11.64}} & 27.43\textsubscript{~\bl{-3.81}} & 46.56\textsubscript{~\bl{-5.90}} & 32.84\textsubscript{~\bl{-5.37}}\\

    \rowcolor[gray]{0.9}CM & 4/4 & 44.89\textsubscript{~\mg{+10.22}} & 142.98\textsubscript{~\bl{-43.47}} &  172.31\textsubscript{~\bl{-25.01}} & 56.93\textsubscript{~\bl{-15.73}} & 26.42\textsubscript{~\bl{-4.82}} & 47.31\textsubscript{~\bl{-5.15}} & 30.42\textsubscript{~\bl{-7.79}}\\

    \rowcolor[gray]{0.9} DS & 4/4 & \textbf{49.67\textsubscript{~\mg{+15.00}}} & \textbf{138.56\textsubscript{~\bl{-47.89}}} &  \textbf{162.44\textsubscript{~\bl{-24.88}}} & \textbf{52.42\textsubscript{~\bl{-20.24}}} & \textbf{25.10\textsubscript{~\bl{-6.14}}} & \textbf{44.32\textsubscript{~\bl{-8.14}}} & \textbf{27.43\textsubscript{~\bl{-10.77}}}\\
    \midrule
    
    PTQ4DM~\cite{shang2022ptq4dm} & 2/4 & 22.68 & 231.04 &  251.87 & 92.43 & 52.79 & 76.22 & 53.44 \\
    Q-Diffusion~\cite{li2023qdiffusion} & 2/4 & 27.41 & 210.56 &  238.92 & 83.77 & 41.03 & 62.99 & 49.41 \\
    PTQD~\cite{he2023ptqd} & 2/4 & 30.12 & 205.33 &  217.06 & 78.44 & 37.63 & 55.60 & 44.23\\
    \rowcolor[gray]{0.9}TM & 2/4 & 38.44\textsubscript{~\mg{+8.32}} & 162.33\textsubscript{~\bl{-43.00}} &  174.42\textsubscript{~\bl{-42.64}} & 64.20\textsubscript{~\bl{-14.24}} & 29.05\textsubscript{~\bl{-8.58}} & 50.01\textsubscript{~\bl{-5.59}} & 35.44\textsubscript{~\bl{-8.79}} \\

    \rowcolor[gray]{0.9}CM & 2/4 & 42.37\textsubscript{~\mg{+12.25}} & 150.21\textsubscript{~\bl{-55.12}} &  179.04\textsubscript{~\bl{-38.02}} & 60.24\textsubscript{~\bl{-18.20}} & 28.41\textsubscript{~\bl{-9.22}} & 49.46\textsubscript{~\bl{-6.14}} & 33.90\textsubscript{~\bl{-10.33}}\\

    \rowcolor[gray]{0.9} DS & 2/4 & \textbf{46.02\textsubscript{~\mg{+15.90}}} & \textbf{147.32\textsubscript{~\bl{-58.01}}} &  \textbf{170.11\textsubscript{~\bl{-46.95}}} & \textbf{56.38\textsubscript{~\bl{-22.06}}} & \textbf{27.24\textsubscript{~\bl{-10.39}}} & \textbf{47.42\textsubscript{~\bl{-8.18}}} & \textbf{31.04\textsubscript{~\bl{-13.19}}}\\
    \bottomrule
\end{tabular}
}
    \label{tab:sub-4-bit}
\end{table}
\yushi{For sub-$4$-bit settings, we employ W$4$A$4$ and W$2$A$4$ quantization with LDM on ImageNet $256\times 256$~\cite{deng2009imagenet}, LSUN-Bedrooms $256\times256$~\cite{deng2009imagenet}, and LSUN-Churches $256\times256$~\cite{yu2016lsun}. The results in Tab.~\ref{tab:sub-4-bit} show the significant performance improvement achieved by our method compared with other PTQ baselines. Specifically, the superiority of our method to other baselines widens when the bit-width decreases. However, we have to clarify that most current hardware does not support sub-$4$-bit quantization. Moreover, in these extremely low-bit settings, QAT, which also re-trains or fine-tunes model weights, is a more common choice. As discussed in the conclusion section, we believe the insight of temporal feature maintenance can also be generalized to QAT settings for further improving sub-$4$-bit quantization performance. This can be left as our future work.}

\section{Experments for FLUX.1-Schnell}\label{app:flux}
\begin{table}[!ht]\setlength{\tabcolsep}{8pt}
  \centering
  \caption{Quantization results for text-guided image generation with FLUX.1-Schnell on MS-COCO captions.} 
  \resizebox{0.8\linewidth}{!}{
  \begin{tabular}[t!]{lclllllllll}
    \toprule
    \multicolumn{1}{c}{\multirow{2}{*}{\textbf{Methods}}} & \multicolumn{1}{c}{\multirow{2}{*}{\textbf{\#Bits (W/A)}}} & \multicolumn{4}{c}{\textbf{FLUX.1-Schnell ($\mathbf{T=1}$)}} &  \multicolumn{4}{c}{\textbf{FLUX.1-Schnell ($\mathbf{T=4}$)}} \\ \cmidrule(r){3-6} \cmidrule(r){7-10} 
    & & FID$\downarrow$ & sFID$\downarrow$ & CLIP$\uparrow$ & SQNR$\uparrow$ & FID$\downarrow$ & sFID$\downarrow$ & CLIP$\uparrow$ & SQNR$\uparrow$ \\
    \midrule
    Full Prec. & 32/32 & 25.64 & 31.33 & 32.42 & - & 22.51 & 35.02 & 33.31 & - \\
    \midrule
    
     Q-Diffusion~\cite{li2023qdiffusion} & 8/8 & 27.44 & 31.87 & 30.45 & 1.07 & 24.92  & 37.52 & 31.48 & 2.14 \\
    \rowcolor[gray]{0.92}TM & 8/8 & 25.14\textsubscript{~\bl{-2.30}} & 30.62\textsubscript{~\bl{-1.25}} &
    32.28\textsubscript{~\mg{+1.83}}&
    3.05\textsubscript{~\mg{+1.98}} & 22.53\textsubscript{~\bl{-2.39}} & 34.27\textsubscript{~\bl{-3.25}} &
   32.99\textsubscript{\mg{+1.51}}&
    4.45\textsubscript{~\mg{+2.31}}\\
    
    \rowcolor[gray]{0.92}CM & 8/8 & 25.48\textsubscript{~\bl{-1.96}} & 31.02\textsubscript{~\bl{-0.85}} &
    31.98\textsubscript{~\mg{+1.53}}&
    3.42\textsubscript{~\mg{+2.35}} & 23.25\textsubscript{~\bl{-1.67}} & 34.08\textsubscript{~\bl{-3.44}} &
   32.87\textsubscript{\mg{+1.39}}&
    5.84\textsubscript{~\mg{+3.70}} \\
    
    \rowcolor[gray]{0.92}DS & 8/8 & \textbf{25.12\textsubscript{~\bl{-2.32}}} & \textbf{30.58\textsubscript{~\bl{-1.29}}} &
    \textbf{32.30\textsubscript{~\mg{+1.85}}}&
    \textbf{4.01\textsubscript{~\mg{+2.94}}} & \textbf{22.51\textsubscript{~\bl{-2.41}}} & \textbf{33.97\textsubscript{~\bl{-3.55}}} &
   \textbf{33.12\textsubscript{\mg{+1.64}}}&
    \textbf{6.10\textsubscript{~\mg{+3.96}}} \\
    \midrule
    
     Q-Diffusion~\cite{li2023qdiffusion} & 4/8 & 28.83 & 32.56 & 30.23 & 0.54 & 25.82  & 38.27 & 31.17 & 0.92 \\
    \rowcolor[gray]{0.92}TM & 4/8 & 25.98\textsubscript{~\bl{-2.85}} & 31.01\textsubscript{~\bl{-1.55}} &
   32.15\textsubscript{\mg{+1.92}}&
    2.73\textsubscript{~\mg{+2.19}} & 23.06\textsubscript{~\bl{-2.76}} & 35.87\textsubscript{~\bl{-2.40}} &
   32.14\textsubscript{\mg{+1.03}} & 3.97\textsubscript{\mg{+3.05}} \\
    
    \rowcolor[gray]{0.92}CM & 4/8 & 26.01\textsubscript{~\bl{-2.82}} & 31.77\textsubscript{~\bl{-0.97}} &
   31.94\textsubscript{\mg{+1.71}}&
    2.49\textsubscript{~\mg{+1.95}} & 23.43\textsubscript{~\bl{-2.39}} & 35.78\textsubscript{~\bl{-2.49}} &
   32.25\textsubscript{\mg{+1.08}}& 4.01\textsubscript{\mg{+3.09}} \\
    
    \rowcolor[gray]{0.92}DS & 4/8 & \textbf{25.81\textsubscript{~\bl{-3.02}}} & \textbf{30.97\textsubscript{~\bl{-1.59}}} &
   \textbf{32.24\textsubscript{\mg{+2.01}}}&
    \textbf{2.94\textsubscript{~\mg{+2.40}}} & \textbf{22.87\textsubscript{~\bl{-2.95}}} & \textbf{35.49\textsubscript{~\bl{-2.78}}} &
   \textbf{32.45\textsubscript{\mg{+1.28}}}&
    \textbf{4.38\textsubscript{~\mg{+3.46}}} \\
    \bottomrule
\end{tabular}
}
    \label{tab:flux}
\end{table}
\yushi{As shown in Tab.~\ref{tab:sota_text}, our method for SD-XL-turbo~\cite{sauer2023adversarial} with a single sampling step distilled from SD-XL-base~\cite{podell2023sdxlimprovinglatentdiffusion} shows satisfactory boosts. The improvement is even $\mathbf{>2}$ points larger than that for the $50$-step SD-XL-base under the W$4$A$8$ setting. Here, we further conduct experiments for FLUX.1-Schnell~\cite{flux2024} in this section. Specifically, we employ $512$ COCO prompts for quantization. Other settings are the same as those of SD-XL-turbo~\cite{sauer2023adversarial}. The results in Tab.~\ref{tab:flux} further validate the essential role of temporal features in this kind of scenario.}

\section{Deployment on Edge and Mobile Devices}\label{app:edge}
 \begin{table}[!ht]\setlength{\tabcolsep}{10pt}
  \centering
  \caption{Inference analysis of our quantized Stable Diffusion employing Distrubance-aware Selection with $50$ denoising timesteps and batch size of $1$. We employ TensorRT~\cite{trt} and Apple Core ML toolkit~\cite{stable-diffusion-coreml-apple-silicon} for deployment on Jetson Orin Nano and iPhone $15$ Pro Max, respectively. Each hardware has $8$GB of GPU memory. The speedup ratio depends on the utilized hardware and acceleration toolkits.} 
  \resizebox{0.45\linewidth}{!}{
  \begin{tabular}{lcclc}
    \toprule
    \textbf{Device} & \textbf{\#Bits (W/A)} & \textbf{\#UNet Size (MB)} & \textbf{Latency (s)}\\
    \midrule
    \multirow{2}{*}{Jetson Orin Nano} & 32/32 & 3278.81 & 138.74 \\
        & \cellcolor[gray]{0.92}8/8  & \cellcolor[gray]{0.92}821.15 & \cellcolor[gray]{0.92}48.68$_{(\mathbf{2.85\times})}$\\
    \midrule
     \multirow{2}{*}{iPhone $15$ Pro Max}& 32/32 & 3278.81 & 19.67\\
       &  \cellcolor[gray]{0.92}8/8  & \cellcolor[gray]{0.92}821.15 & \cellcolor[gray]{0.92}9.93$_{(\mathbf{1.98\times})}$\\
    \bottomrule
\end{tabular}
}
    \label{tab:infer}
\end{table}
\yushi{More than focusing on very high-calculus devices in Sec.~\ref{sec:deploy}, we also deploy quantized models on the edge and mobile devices. Since our methods do not induce any operators that require specific hardware support, there is no constraint for deploying our models on devices that support normal uniform quantization. As illustrated in Tab.~\ref{tab:infer}, we deploy Stable Diffusion on NVIDIA Jetson Orin Nano and iPhone $15$ Pro Max. The quantized model achieves $\mathbf{2.85\times}$ and $\mathbf{1.98\times}$ speedup, respectively.}

\section{Visualization Results}\label{app:vis}
\begin{figure*}[!ht]
   \centering
   \setlength{\abovecaptionskip}{0.2cm}
   \begin{minipage}[t]{0.48\textwidth}
      \centering
      \begin{subfigure}[t]{\textwidth}
          \centering
      \includegraphics[width=\textwidth]{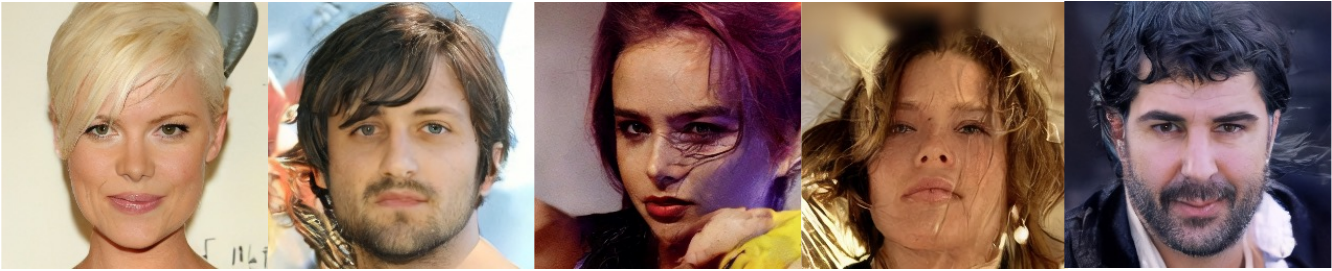}
      \subcaption{FP}
      \end{subfigure}
      \begin{subfigure}[t]{\textwidth}
          \centering
      \includegraphics[width=\textwidth]{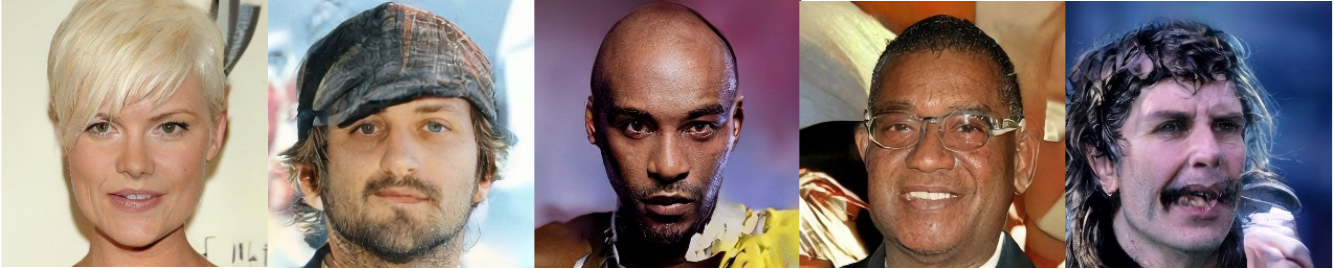}
      \subcaption{Q-Diffusion (W$4$A$8$)}
      \end{subfigure}
      \begin{subfigure}[t]{\textwidth}
          \centering
      \includegraphics[width=\textwidth]{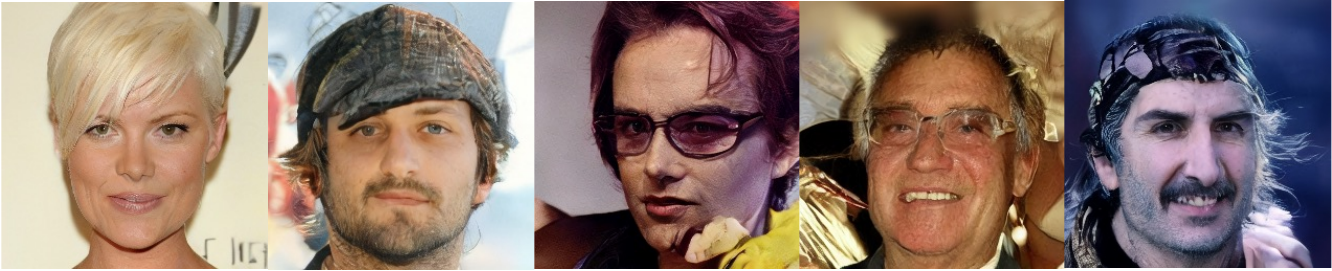}
      \subcaption{PTQD (W$4$A$8$)}
      \end{subfigure}
      \begin{subfigure}[t]{\textwidth}
          \centering
      \includegraphics[width=\textwidth]{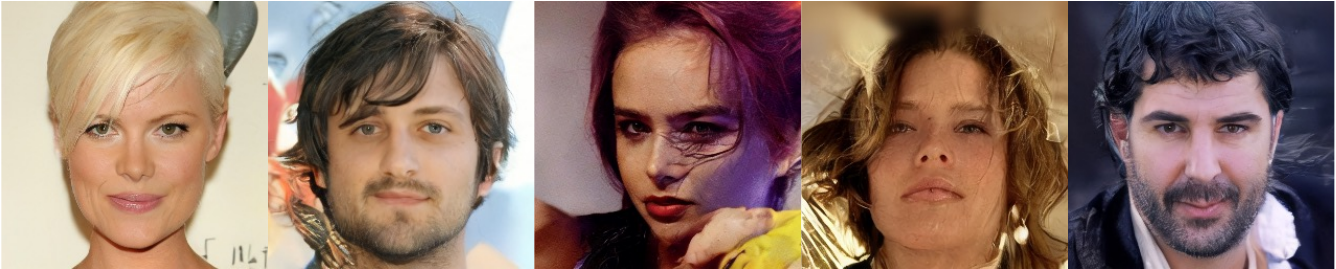}
      \subcaption{Ours (W$4$A$8$)}
      \end{subfigure}
      \caption{Random samples from W$4$A$8$ quantized and FP LDM-4 on CelebA-HQ $256 \times 256$. The resolution of each sample is $256\times256$.}
      \label{first_celeba-hq}
   \end{minipage}\hfill
   \begin{minipage}[t]{0.48\textwidth}
      \centering
      \begin{subfigure}[t]{\textwidth}
          \centering
      \includegraphics[width=\textwidth]{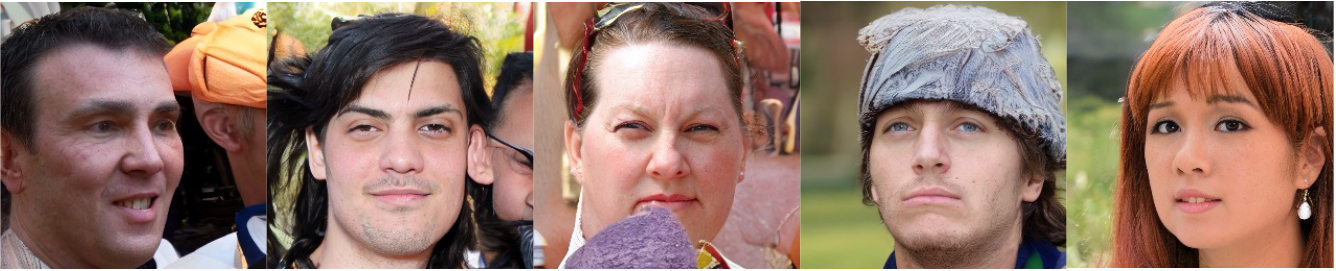}
      \subcaption{FP}
      \end{subfigure}
      \begin{subfigure}[t]{\textwidth}
          \centering
      \includegraphics[width=\textwidth]{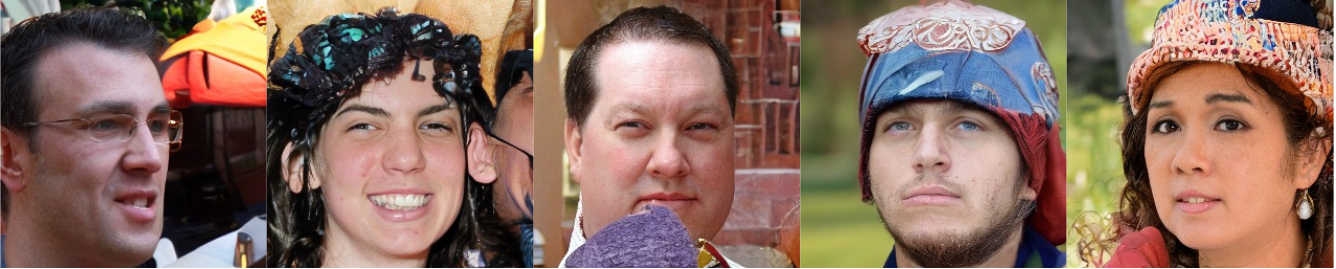}
      \subcaption{Q-Diffusion (W$4$A$8$)}
      \end{subfigure}
      \begin{subfigure}[t]{\textwidth}
          \centering
      \includegraphics[width=\textwidth]{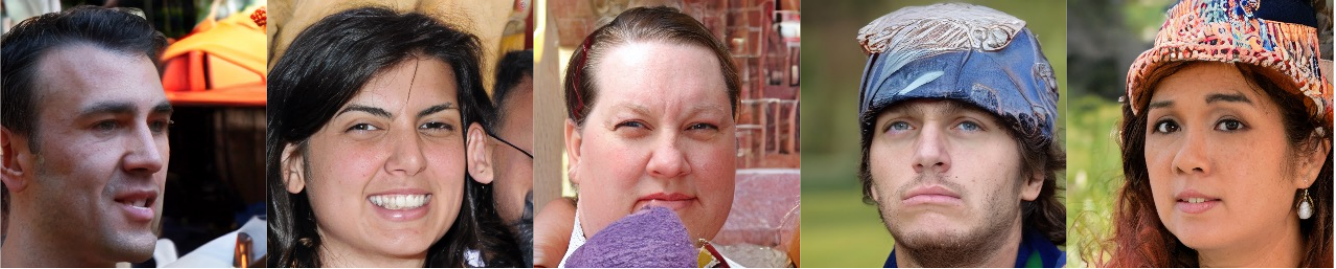}
      \subcaption{PTQD (W$4$A$8$)}
      \end{subfigure}
      \begin{subfigure}[t]{\textwidth}
          \centering
      \includegraphics[width=\textwidth]{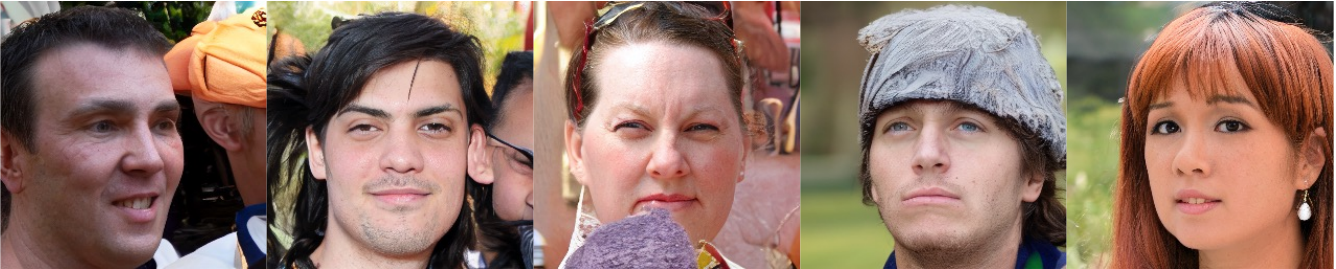}
      \subcaption{Ours (W$4$A$8$)}
      \end{subfigure}
      \caption{Random samples from W$4$A$8$ quantized and FP LDM-4 on FFHQ $256 \times 256$. The resolution of each sample is $256\times256$.}
   \end{minipage}
\end{figure*}

\begin{figure*}[!ht]
\centering
\setlength{\abovecaptionskip}{0.2cm}
\renewcommand{\arraystretch}{0.5}
\begin{tabularx}{\linewidth}{X@{\hspace{5pt}}X}
\toprule
\footnotesize{\shortstack{\textit{``A surreal landscape featuring a giant clock melting over a cliff, under a sky }\\\textit{filled with swirling clouds and two suns, reminiscent of Salvador Dali's}\\\textit{ style.
''}}} & \footnotesize{\shortstack{\textit{``A bustling medieval market scene with vendors selling colorful spices and}\\\textit{ textiles, horses and carts in the background, and a castle on a hill in the }\\\textit{ distance.
''}}}\\
\midrule
\includegraphics[width=0.488\textwidth]{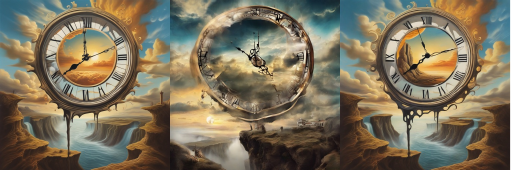} & \includegraphics[width=0.48\textwidth]{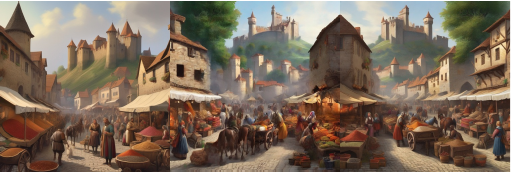}\\
\midrule
\footnotesize{\shortstack{\textit{``A futuristic cityscape at night, illuminated by neon lights, with flying cars }\\\textit{ zooming between high-rise buildings that have gardens growing on their }\\\textit{roofs.
''}}} & \footnotesize{\shortstack{\textit{``An underwater scene showing a coral reef with a variety of colorful fish, a }\\\textit{sunken pirate ship in the background, and light filtering through the water }\\\textit{from above.
''}}}\\
\midrule
\includegraphics[width=0.488\textwidth]{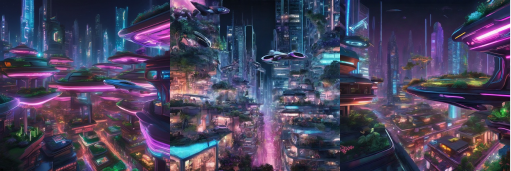} & \includegraphics[width=0.488\textwidth]{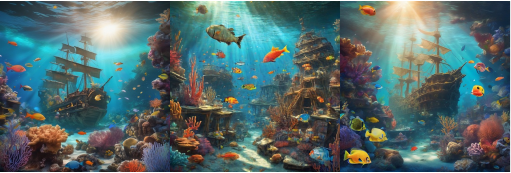}\\
\midrule
\footnotesize{\shortstack{\textit{``A scene from a fantasy novel showing a wizard casting a spell in a library }\\\textit{full of ancient books, with magical symbols glowing in the air around him.
''}\\\textit{}}} & \footnotesize{\shortstack{\textit{``A serene autumn landscape in a Japanese garden with a red bridge over a }\\\textit{pond filled with koi fish, surrounded by maple trees with leaves changing }\\\textit{color.
''}}}\\
\midrule
\includegraphics[width=0.488\textwidth]{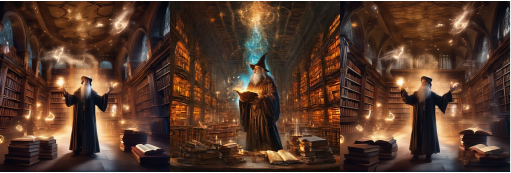} & \includegraphics[width=0.488\textwidth]{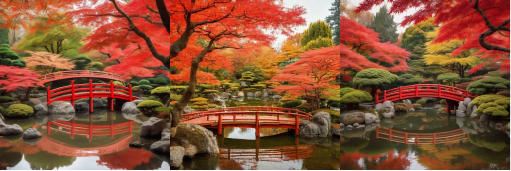}\\
\midrule
\footnotesize{\shortstack{\textit{``An action-packed scene from a superhero comic book showing a hero in a }\\\textit{bright costume flying above a city, chasing a villain who is escaping on a }\\\textit{high-tech motorcycle.
''}}} & \footnotesize{\shortstack{\textit{``A traditional African village at sunset, with round mud huts, people dressed }\\\textit{in colorful clothing, and a large baobab tree in the center of the village.
''}\\\textit{}}}\\
\midrule
\includegraphics[width=0.488\textwidth]{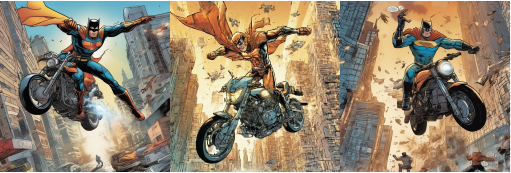} & \includegraphics[width=0.488\textwidth]{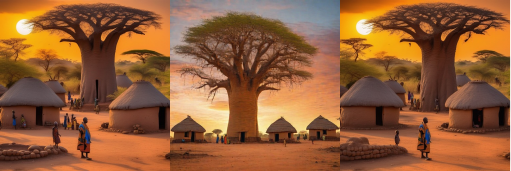}\\
\bottomrule
\end{tabularx}
\caption{\label{last_sd}Random samples from W$4$A$8$ quantized and FP SD-XL. The resolution of each sample is $1024\times 1024$. The images below the corresponding prompts are generated from FP (Left), Q-Diffusion (Middle), and our framework (Right).}
\end{figure*}

We present random samples derived from FP and W$4$A$8$ quantized diffusion models with a fixed random seed. These quantized models were created through our complete framework or previous methods. As shown from Fig.~\ref{first_celeba-hq} to Fig.~\ref{last_sd}, our framework yields results that closely resemble those of the FP model, showcasing higher fidelity. Moreover, it excels in finer details, producing superior outcomes in some intricate aspects (zoom in to closely examine the relevant images).

\begin{figure*}[!ht]
   \centering
    \setlength{\abovecaptionskip}{0.2cm}
\begin{subfigure}[t]{\textwidth}
          \centering
          \includegraphics[width=0.35\textwidth]{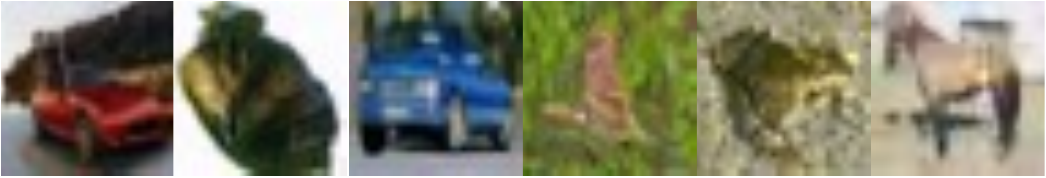}
          \subcaption{FP}
      \end{subfigure}
      \begin{subfigure}[t]{\textwidth}
          \centering
          \includegraphics[width=0.35\textwidth]{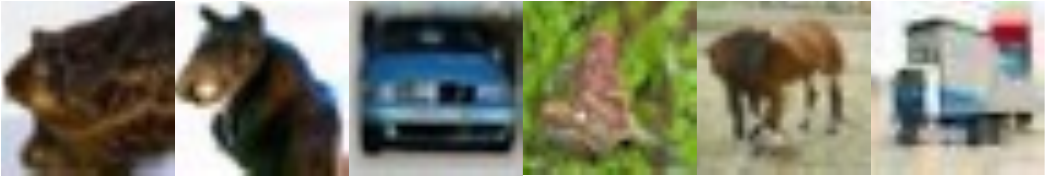}
          \subcaption{Q-Diffusion (W$4$A$8$)}
      \end{subfigure}
      \begin{subfigure}[t]{\textwidth}
          \centering
          \includegraphics[width=0.35\textwidth]{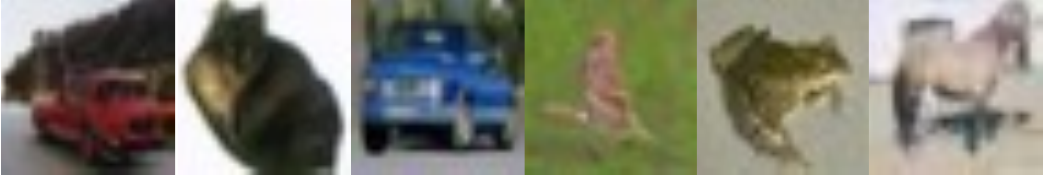}
          \subcaption{Ours (W$4$A$8$)}
      \end{subfigure}
      \caption{Random samples from W$4$A$8$ quantized and FP DDIM on CIFAR-10 $32 \times 32$. The resolution of each sample is $32\times 32$.}
\end{figure*}

\begin{figure*}[!ht]
   \centering
   \setlength{\abovecaptionskip}{0.2cm}
   \begin{minipage}[t]{0.48\textwidth}  %
      \centering
      \begin{subfigure}[t]{\textwidth}
          \centering
          \includegraphics[width=\textwidth]{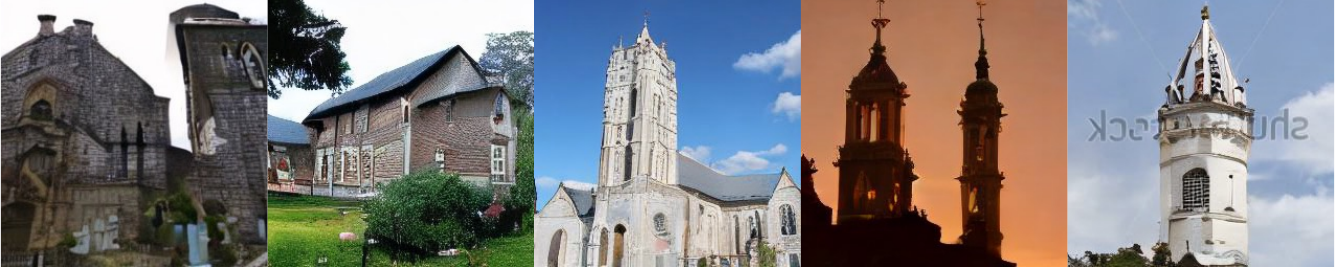}
          \subcaption{FP}
      \end{subfigure}
      \begin{subfigure}[t]{\textwidth}
          \centering
          \includegraphics[width=\textwidth]{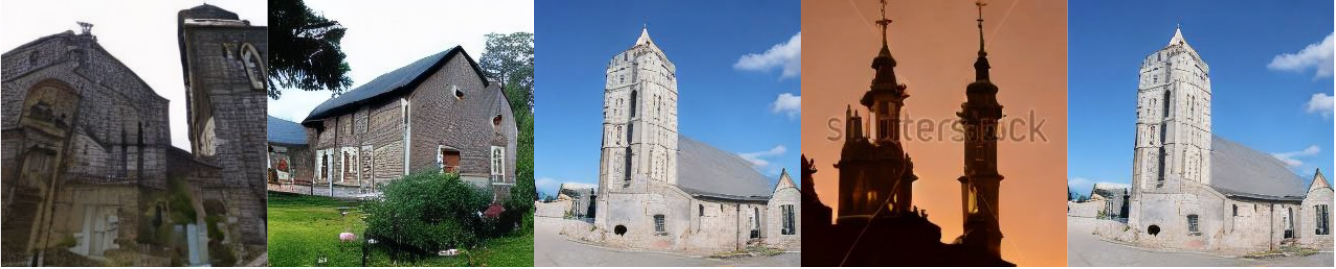}
          \subcaption{Q-Diffusion (W$4$A$8$)}
      \end{subfigure}
      \begin{subfigure}[t]{\textwidth}
          \centering
          \includegraphics[width=\textwidth]{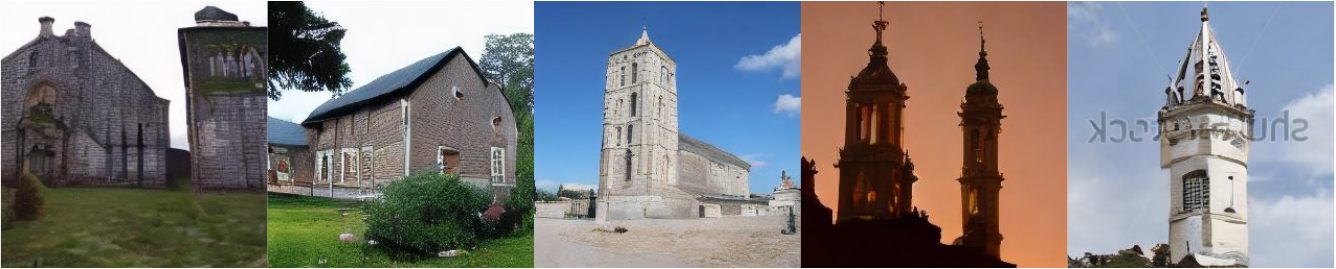}
          \subcaption{PTQD (W$4$A$8$)}
      \end{subfigure}
      \begin{subfigure}[t]{\textwidth}
          \centering
          \includegraphics[width=\textwidth]{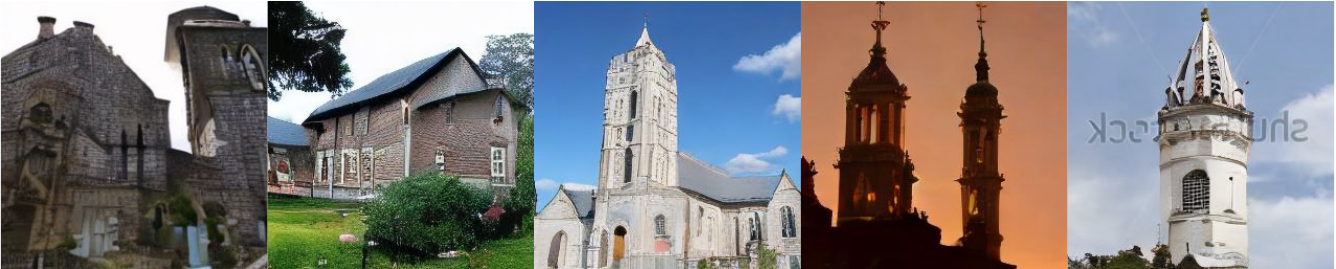}
          \subcaption{Ours (W$4$A$8$)}
      \end{subfigure}
      \caption{Random samples from W$4$A$8$ quantized and FP LDM-8 on LSUN-Churches $256 \times 256$. The resolution of each sample is $256\times256$.}
   \end{minipage}\hfill
   \begin{minipage}[t]{0.48\textwidth}  %
      \centering
      \vspace*{\fill}
      \begin{subfigure}[t]{\textwidth}
          \centering
          \includegraphics[width=\textwidth]{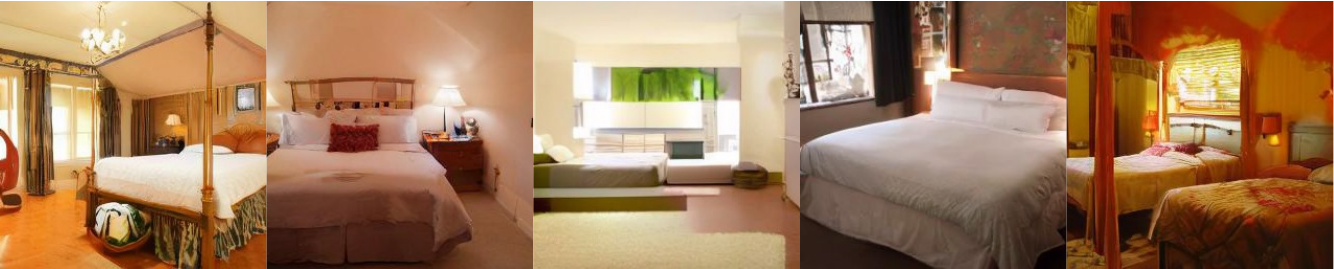}
          \subcaption{FP}
      \end{subfigure}
      \begin{subfigure}[t]{\textwidth}
          \centering
          \includegraphics[width=\textwidth]{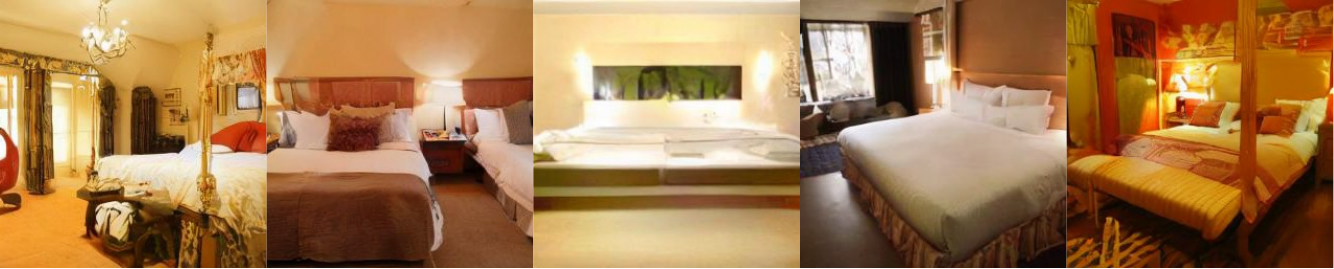}
          \subcaption{PTQD (W$4$A$8$)}
      \end{subfigure}
      \begin{subfigure}[t]{\textwidth}
          \centering
          \includegraphics[width=\textwidth]{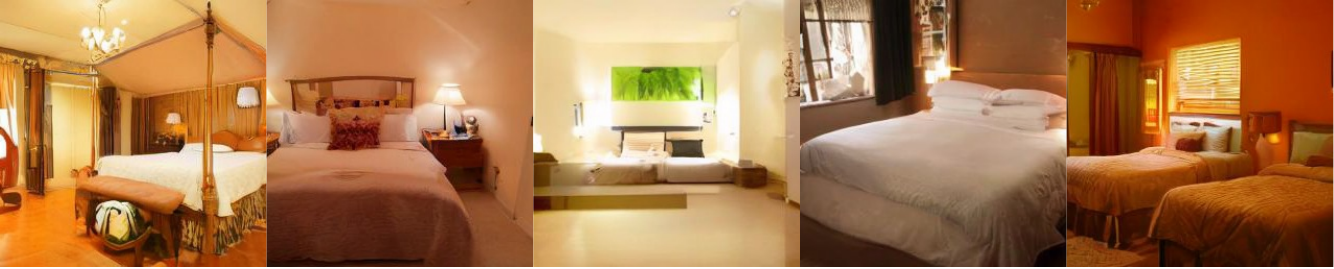}
          \subcaption{Ours (W$4$A$8$)}
      \end{subfigure}
      \caption{Random samples from W$4$A$8$ quantized and FP LDM-4 on LSUN-Bedrooms $256 \times 256$. The resolution of each sample is $256\times256$.}
   \end{minipage}
\end{figure*}

\begin{figure*}[!ht]
   \centering
   \setlength{\abovecaptionskip}{0.2cm}
   
   \begin{minipage}[b]{0.48\textwidth}
        \centering
        \begin{subfigure}[tp!]{\textwidth}
        \centering
        \includegraphics[width=\textwidth]{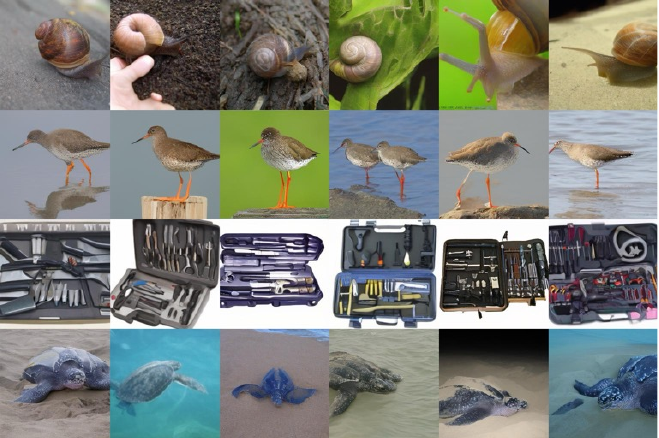}
        \subcaption{FP}
        \end{subfigure}
   \end{minipage}\hfill
   \begin{minipage}[b]{0.48\textwidth}
        \centering
        \begin{subfigure}[tp!]{\textwidth}
        \centering
        \includegraphics[width=\textwidth]{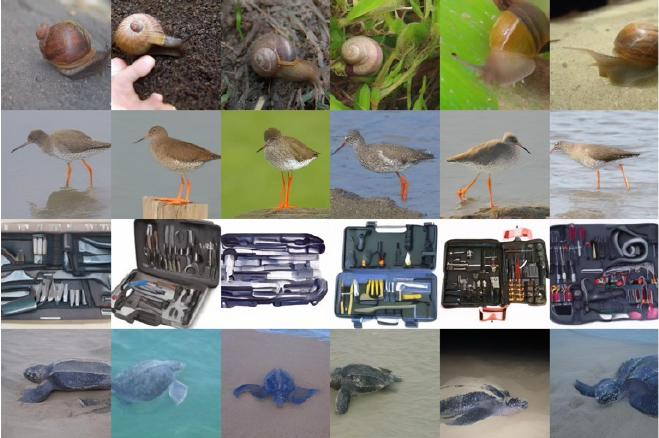}
        \subcaption{PTQD (W$4$A$8$)}
        \end{subfigure}
   \end{minipage}
   
   \begin{minipage}[b]{\textwidth}
        \centering
        \begin{subfigure}[tp!]{0.48\textwidth}
        \centering
        \includegraphics[width=\textwidth]{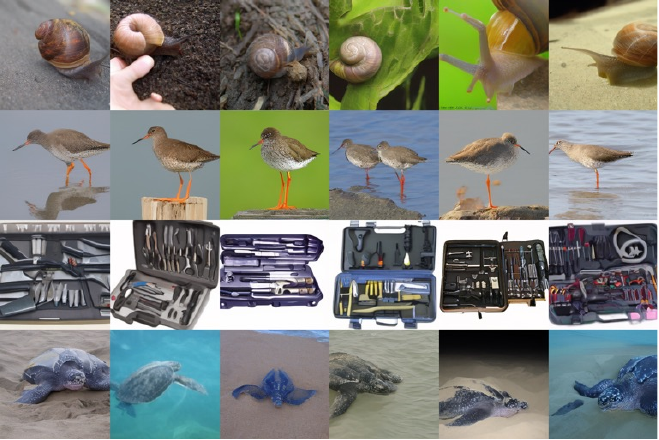}
        \subcaption{Ours (W$4$A$8$)}
        \end{subfigure}
   \end{minipage}
   \caption{Random samples from W$4$A$8$ quantized and FP LDM-4 on ImageNet $256 \times 256$. The resolution of each sample is $256 \times 256$.}
\end{figure*}

\begin{figure*}[!ht]
\centering
\setlength{\abovecaptionskip}{0.2cm}
\renewcommand{\arraystretch}{0.5}
\begin{tabularx}{\linewidth}{X@{\hspace{1.5pt}}X@{\hspace{1.5pt}}X}
\toprule
\footnotesize{\shortstack{\textit{``A dog lies in the grass with a frisbee under its} \\\textit{paw.
''}}} & \footnotesize{\shortstack{\textit{``An eagle is in full wing span while flying in a}\\\textit{ sunset sky.
''}}} & \footnotesize{\shortstack{\textit{``Two people out in the snow doing cross country} \\\textit{skiing.
''}}}\\
\midrule
\includegraphics[width=0.323\textwidth]{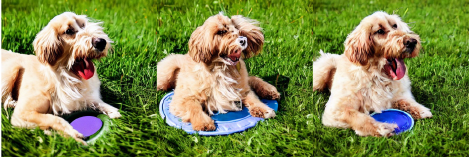} & \includegraphics[width=0.323\textwidth]{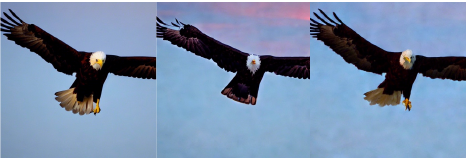}&
\includegraphics[width=0.323\textwidth]{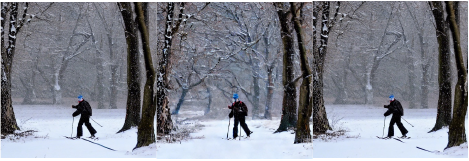}\\
\midrule
\footnotesize{\shortstack{\textit{``A large display of many different types of}\\\textit{doughnuts.
''}}} & \footnotesize{\shortstack{\textit{``A thick crusted pizza just taken out of the
}\\\textit{oven.
''}}} & \footnotesize{\shortstack{\textit{``A thick man in a white suit and tie wearing a
}\\\textit{ name badge.
''}}}\\
\midrule
\includegraphics[width=0.323\textwidth]{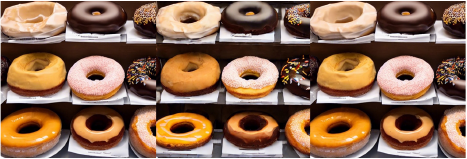} & \includegraphics[width=0.323\textwidth]{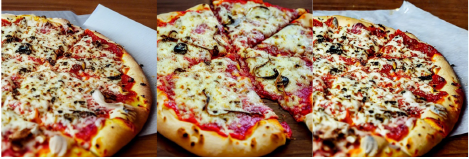} &\includegraphics[width=0.323\textwidth]{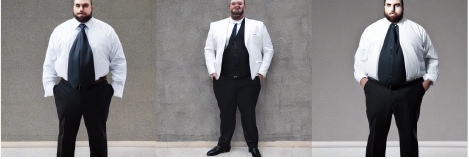}\\
\bottomrule
\end{tabularx}
\caption{Random samples from W$4$A$8$ quantized and FP Stable Diffusion-v1-4 on MS-COCO captions. The resolution of each sample is $512\times 512$. The images below the corresponding prompts are generated from FP (Left), Q-Diffusion (Middle), and our framework (Right).
}
\end{figure*}

\begin{figure*}[!ht]
\centering
\setlength{\abovecaptionskip}{0.2cm}
\renewcommand{\arraystretch}{0.5}
\begin{tabularx}{\linewidth}{X@{\hspace{1.5pt}}X@{\hspace{1.5pt}}X}
\toprule
\footnotesize{\shortstack{\textit{``Early morning on a Parisian street, newspapers}\\\textit{ scattered on small tables outside a café, with the }\\\textit{Eiffel Tower visible in the soft dawn light.
''}\\\textit{}}} & \footnotesize{\shortstack{\textit{``A digital artist's studio cluttered with multiple }\\\textit{screens displaying vibrant digital art, sketches on  }\\\textit{the walls, and a high-tech drawing tablet on the  }\\\textit{desk.
''}}} & \footnotesize{\shortstack{\textit{``A vibrant carnival scene in Rio de Janeiro, with  }\\\textit{dancers in feathered costumes, crowded streets  }\\\textit{filled with revelers, and colorful floats.
''}\\\textit{}}}\\
\midrule
\includegraphics[width=0.323\textwidth]{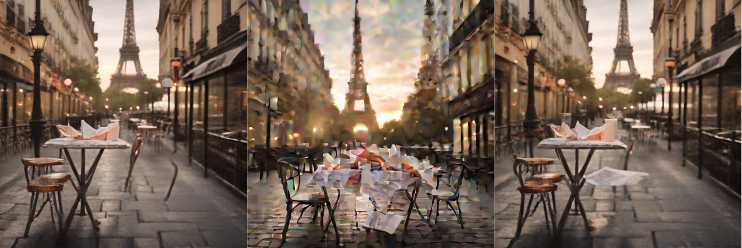} & \includegraphics[width=0.323\textwidth]{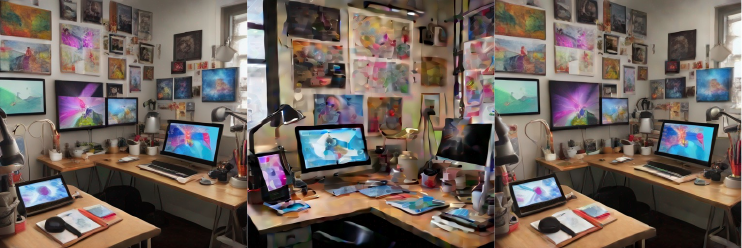}&
\includegraphics[width=0.323\textwidth]{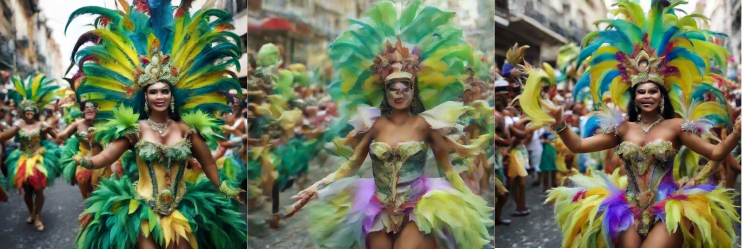}\\
\midrule
\footnotesize{\shortstack{\textit{``A serene mountain landscape in the Himalayas, }\\\textit{with a monk meditating by a crystal-clear lake}\\\textit{surrounded by snow-capped peaks.
''}\\\textit{}}} & \footnotesize{\shortstack{\textit{``A bustling spaceport on Mars, with spacecrafts of }\\\textit{various designs docked, astronauts in futuristic }\\\textit{gear, and a red Martian landscape in the }\\\textit{background.
''}}} & \footnotesize{\shortstack{\textit{``A dense rainforest canopy viewed from above, }\\\textit{with sunlight filtering through the leaves and}\\\textit{ a glimpse of a river winding through the forest.
''}\\\textit{}}}\\
\midrule
\includegraphics[width=0.323\textwidth]{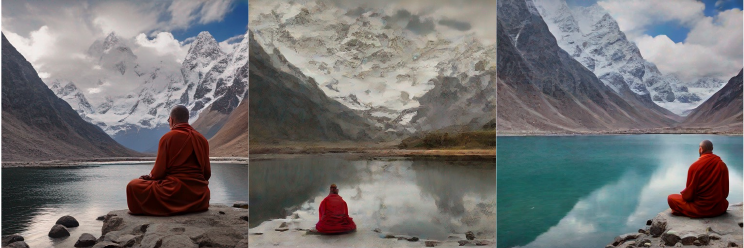} & \includegraphics[width=0.323\textwidth]{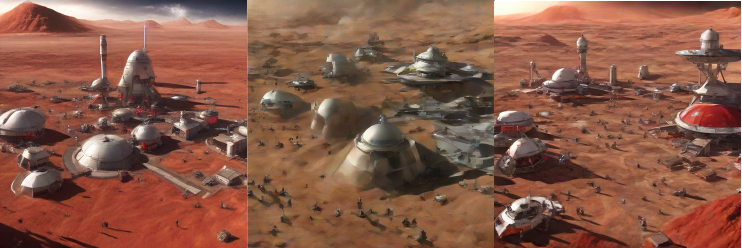} &\includegraphics[width=0.323\textwidth]{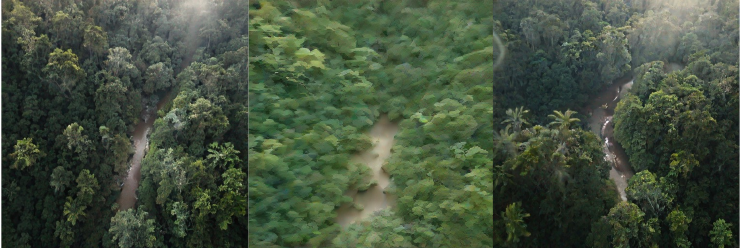}\\
\midrule
\footnotesize{\shortstack{\textit{``A traditional Arabian night scene, with a market}\\\textit{ lit by lanterns, camels resting, and traders}\\\textit{ selling spices and silks under starry skies.
''}\\\textit{}}} & \footnotesize{\shortstack{\textit{``An old Victorian library with tall, wooden }\\\textit{bookshelves filled to the brim, a ladder to reach}\\\textit{ the higher shelves, and a large globe next to a }\\\textit{classic reading nook.
''}}} & \footnotesize{\shortstack{\textit{``A high-speed chase in a futuristic city, with }\\\textit{hovercars zooming past neon billboards and }\\\textit{through towering skyscrapers.
''}\\\textit{}}}\\
\midrule
\includegraphics[width=0.323\textwidth]{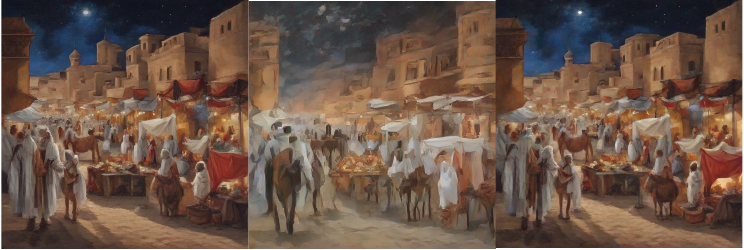} & \includegraphics[width=0.323\textwidth]{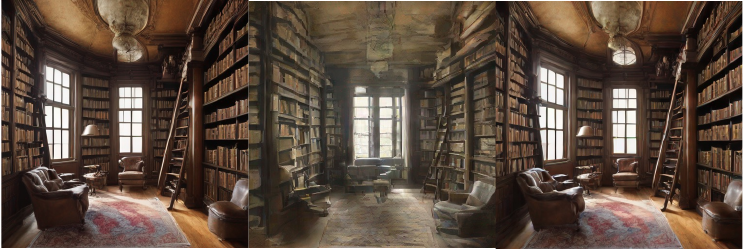} &\includegraphics[width=0.323\textwidth]{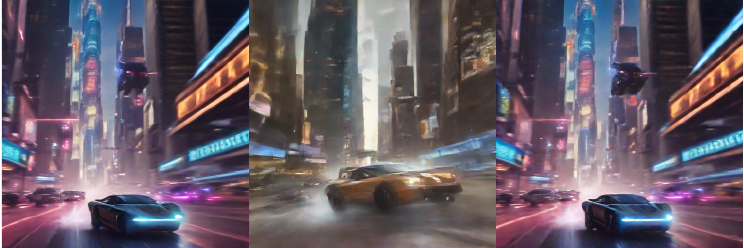}\\
\midrule
\footnotesize{\shortstack{\textit{``A mystical fairy garden at twilight, with glowing }\\\textit{flowers, tiny fairies with delicate wings, and an }\\\textit{ancient oak tree at the center.
''}\\\textit{}}} & \footnotesize{\shortstack{\textit{``A noir-themed detective's office from the 1940s, }\\\textit{with a desk lamp casting shadows, rain against }\\\textit{the window, and a mystery novel left open.
''}\\\textit{}}} & \footnotesize{\shortstack{\textit{``A bustling medieval blacksmith shop with sparks }\\\textit{flying as a blacksmith hammers a sword on an anvil, }\\\textit{armor pieces hanging on the wall, and a fire blazing }\\\textit{in the forge.
''}}}\\
\midrule
\includegraphics[width=0.323\textwidth]{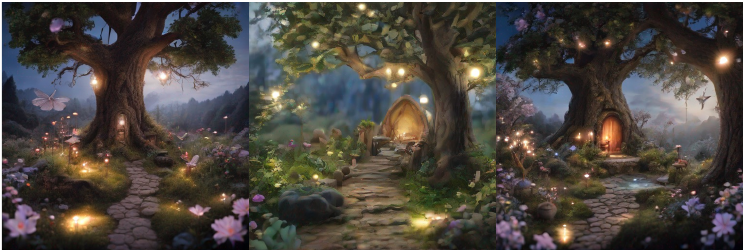} & \includegraphics[width=0.323\textwidth]{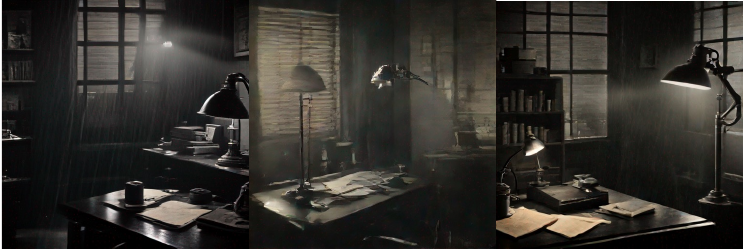} &\includegraphics[width=0.323\textwidth]{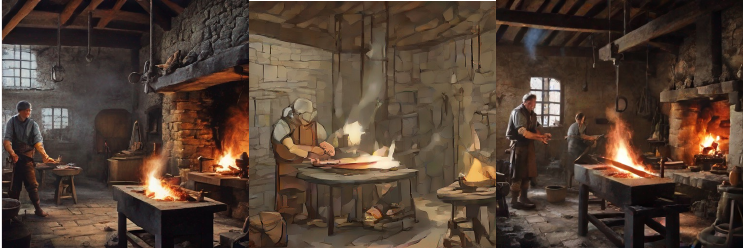}\\
\bottomrule
\end{tabularx}
\caption{Random samples from W$4$A$8$ quantized and FP SD-XL-turbo. The resolution of each sample is $512\times 512$. The images below the corresponding prompts are generated from FP (Left), Q-Diffusion (Middle), and our framework (Right).}
\end{figure*}

\end{document}